\documentclass[11pt]{article}

\usepackage{amsmath,amsfonts,bm}

\usepackage[preprint]{acl}

\usepackage{times}

\usepackage[T1]{fontenc}

\usepackage[utf8]{inputenc}

\usepackage{microtype}

\usepackage{inconsolata}

\usepackage{graphicx}

\usepackage{hyperref}
\usepackage{url}

\usepackage{booktabs}
\usepackage{caption}
\usepackage{array}
\usepackage{fvextra}
\usepackage{xspace}
\usepackage{xcolor}
\usepackage{subcaption}

\usepackage{float}

\usepackage{enumitem}

\DefineVerbatimEnvironment{Prompt}{Verbatim}{
  breaklines=true,
  breakanywhere=true,
  formatcom=\ttfamily
}

\usepackage{tikz}
\usetikzlibrary{shapes,calc,positioning}

\global

\usepackage[most]{tcolorbox}
\tcbset{
  aibox/.style={
    width=\textwidth,
    top=10pt,
    colback=white,
    colframe=black,
    colbacktitle=black,
    enhanced,
    center,
    attach boxed title to top left={yshift=-0.1in,xshift=0.15in},
    boxed title style={boxrule=0pt,colframe=white,},
  }
}
\newtcolorbox{AIbox}[2][]{aibox,title=#2,#1}

\newcommand{\rev}[1]{#1}

\definecolor{goldorange}{RGB}{204,132,0}

\title{\fwname: Agent Meets Checklist for Evaluating LLMs on Long-Context Legal Summarization}

\author{
Yao~Dou \quad Benjamin~Mamut \quad Wei~Xu \\
Georgia Institute of Technology \\
\texttt{douy@gatech.edu, bmamut3@gatech.edu, wei.xu@cc.gatech.edu} \\[0.6em]
\href{https://yao-dou.github.io/gavel/}{\textcolor{goldorange}{\texttt{yao-dou.github.io/gavel/}}}
}

\newcommand{\fwname}{\textsc{Gavel}\xspace}
\newcommand{\fwnameRef}{\textsc{Gavel-Ref}\xspace}
\newcommand{\fwnameAgent}{\textsc{Gavel-Agent}\xspace}

\newcommand{\code}[1]{\texttt{\detokenize{#1}}}

\begin{document}

\maketitle

\begin{abstract}
Large language models (LLMs) now support contexts of up to 1M tokens, but their strengths and weaknesses on complex long-context tasks remain unclear.
To study this, we focus on multi-document legal case summarization, where a single case often spans many documents exceeding 100K tokens.
We systematically evaluate 12 frontier LLMs with \fwname, which consists of \fwnameRef, a reference-based evaluation framework with checklist, residual-fact, and writing-style evaluations, and \fwnameAgent, a reference-free agent for evaluating factual coverage directly from source documents.
Our results show that current models are more prone to omitting key information than hallucinating. 
They all perform well on simple checklist items, such as \texttt{filing date}, but struggle with rare and complex items, such as \texttt{settlements}.
Performance also declines as case length increases.
To meta-evaluate \fwname, we collect 160 hours of human annotations. \fwnameAgent reduces token usage by at least 36\% compared to end-to-end and chunk-by-chunk methods while achieving competitive performance.
\fwnameAgent also generalizes to the medical domain, performing the best with at least 77\% fewer tokens.
\end{abstract}

\section{Introduction}

Modern LLMs advertise very long context windows, with models such as Gemini \cite{comanici2025gemini} and GPT \cite{achiam2023gpt}  supporting one million tokens since 2025. However, \textit{do LLMs actually utilize this entire long context effectively?}

\begin{figure}[t]
  \centering
  \includegraphics[width=\linewidth]{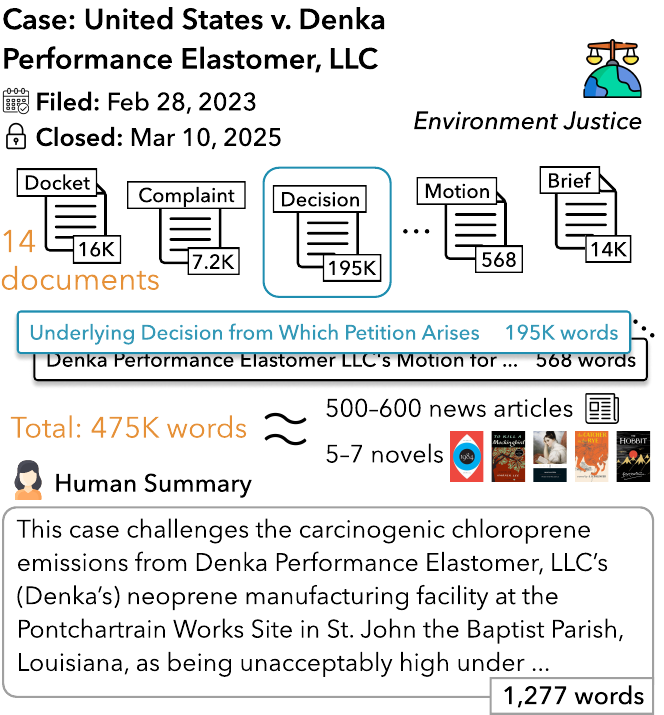}
  \vspace{-20pt}
  \caption{Example of a closed 2025 legal case. The case contains 14 documents totaling roughly the length of five novels, requiring a summarization model to integrate information across case records rather than summarize a single linear document.}
  \label{fig:figure1}
  \vspace{-14pt}
\end{figure}

To answer this question, we develop, \fwname, a comprehensive evaluation framework and present an in-depth study using multi-document legal summarization for civil rights lawsuits as a representative long-context task. We focus on this setting because a single litigation case can contain dozens of court documents, easily exceeding 100K tokens and sometimes surpassing 1M tokens. Unlike summarization of news, where salient information often appears near the beginning of articles \citep{narayan-etal-2018-dont,liu2019text}, or novels, where events are typically presented sequentially \citep{chang2024booookscore}, legal summarization requires models to integrate interconnected arguments across multiple documents while maintaining chronology, preserving relationships among parties, claims, and rulings, and ensuring accurate cross-references. Importantly, the availability of high-quality summaries not only enables us to identify weaknesses in current LLMs relative to human experts, but also  to conduct a meta-evaluation of this task to assess the reliability of the automatic evaluation framework itself.\footnote{Civil Rights Litigation Clearinghouse, a website with over 250,000 yearly visitors, provides case summaries written by legal experts under strict content and style guidelines. Each summary requires 1$\sim$4 hours of expert effort, along with additional time for updates as cases evolve.}

\begin{figure*}[!t]
  \centering
  \includegraphics[width=0.99\textwidth]{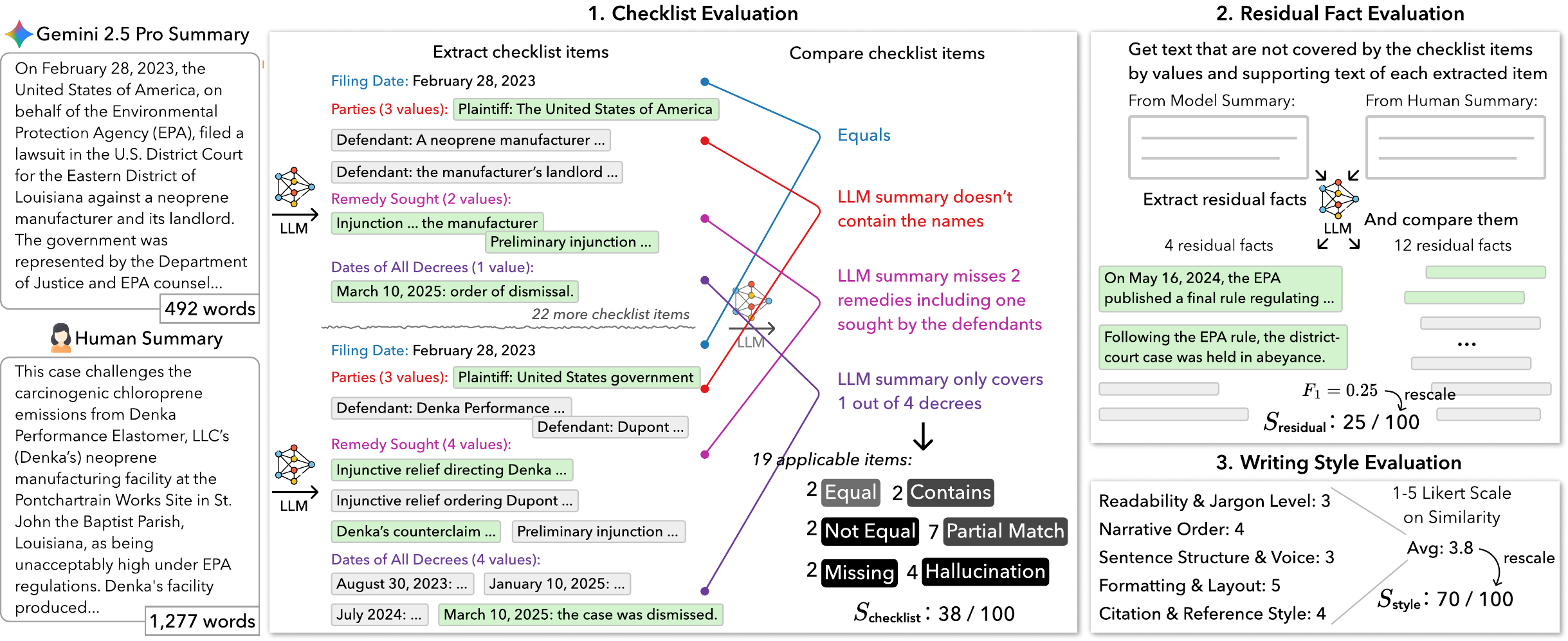}
  \vspace{-7pt}
  \caption{Example of evaluating a Gemini 2.5 Pro summary with \fwnameRef, which contains: checklist evaluation supporting both string-wise and list-wise comparisons, residual fact evaluation, and writing-style evaluation.}
  \label{fig:gavel-ref}
  \vspace{-15pt}
\end{figure*}

\fwname consists of two complementary components. First, \fwnameRef (Fig. \ref{fig:gavel-ref}) is a reference-based framework grounded in expert-written summaries. It augments checklist evaluation with residual-fact evaluation to capture important information beyond predefined categories (78\% of expert summaries contain residual facts, averaging 6.8 such facts). 
Second, \fwnameAgent is a reference-free agent scaffold for assessing factual consistency and coverage directly from source documents.
It equips LLMs with six tools to navigate long document collections and locate checklist items, emulating how humans process case records.

Using \fwnameRef, we evaluate 12 proprietary and open-source LLMs on litigation cases covering 18 diverse subject areas, including Immigration and Border Policy, Education, Free Speech and Religious Freedom, Environmental Justice, and Equal Employment. We deliberately select recent cases from 2025 to minimize the risk of data contamination, with many cases substantially exceeding the 128K-token context limit examined in prior work \cite{yen2024helmet,ruan2025expertlongbench}. 
Our results show that current LLMs remain far from reliable for this task. Even the best model we test, Gemini 2.5 Pro, achieves only 59.1 F$_1$ on checklist evaluation, with 66.6 precision and 55.3 recall, despite receiving the checklist definitions in the prompt. The main failure mode is omission rather than hallucination. Omissions are especially severe for rare items such as \texttt{related cases} and \texttt{settlements}. Processing the case documents chunk by chunk and iteratively extracting checklist items substantially improves coverage, increasing recall by 28 points on average compared to top LLMs' summaries. Summary performance also declines with case length: Gemini 2.5 Pro drops by about 10\% in checklist F$_1$ from 32K-token to 512K-token cases.
Beyond checklist items, GPT-4.1 and GPT-5 include additional facts that better overlap with human summaries. Over 90\% of their summaries contain more than 8 such facts, compared with fewer than 4 per case for other models.

Although prior work often evaluates automatic evaluators by manually verifying their outputs, systematic human annotations for meta-evaluation remain limited. We thus spend 160 hours of human effort collecting 5,442 checklist item-level annotations, 450 checklist comparison judgments, 281 residual-fact extractions, and 375 style-similarity ratings. 
We find that \fwnameRef with open-source GPT-oss \citep{agarwal2025gpt} and Qwen3 \citep{yang2025qwen3} achieves performance comparable to GPT-5, showing that reliable large-scale automatic evaluation can be both accurate and cost-effective. For reference-free document-based checklist extraction, \fwnameAgent achieves competitive performance while using 36--54\% fewer tokens than end-to-end and chunk-by-chunk methods, though it still trails the oracle setting where the model extracts from reference summaries.
Since \fwnameAgent is not inherently constrained to the legal domain, adapting it to a new domain only requires redefining the checklist items. We test this by applying it to medical literature reviews. In this setting, \fwnameAgent achieves the best performance while using at least 77\% fewer tokens than end-to-end and chunk-by-chunk approaches, demonstrating its generalizability across domains.

We release our data and code at \url{https://yao-dou.github.io/gavel/}. In summary, our contributions are as follows:
\begin{enumerate}[leftmargin=*,itemsep=0em,topsep=-0.3em]

\item We introduce \fwname, a comprehensive framework for evaluating long-context multi-document summarization, with \fwnameRef for reference-based evaluation and \fwnameAgent for reference-free document-based evaluation.

\item We conduct a large-scale human meta-evaluation with 160 hours of annotation effort, showing that \fwnameRef can provide accurate and cost-effective automatic evaluation using strong open-source models.

\item Using \fwnameRef, we evaluate 12 frontier LLMs on recent long-context litigation cases and show that current models remain far from reliable: they often omit rare legal information and degrade as case length increases.

\item We develop \fwnameAgent, a generalizable agent scaffold that extracts checklist information directly from source documents with competitive performance and substantially lower token usage, and demonstrate its transferability beyond law on medical literature reviews.

\end{enumerate}

\section{Related Work}
Due to space constraints, we focus on the most relevant work on summarization evaluation. Additional related work on legal summarization datasets and LLM agent scaffolds is discussed in App. \ref{app:additional_related_work}.

As modern LLMs can generate fluent narrative text, evaluation becomes focusing on whether model summaries capture the same important facts as human references. Accordingly, the field has shifted from n-gram metrics \cite{lin-2004-rouge} toward factual evaluation \cite{min-etal-2023-factscore,chen2023menli,kamoi2023wice}, where summaries and references are decomposed into atomic facts and compared against each other. 
To improve the interpretability and objectivity of atomic-fact evaluation, \citet{ruan2025expertlongbench} introduce expert-designed checklists for 11 long-context tasks, including a 26-item checklist for legal summarization, inspired by rubric- or checklist-based evaluation \citep{lee2024checkeval,qin-etal-2024-infobench,lin2025wildbench,furuhashi2025checklists,arora2025healthbench}. These checklist items capture key information that should appear in a summary when applicable, such as \texttt{filing date} and \texttt{parties}. However, the checklist does not cover all content. We find that 78\% of human expert summaries contain residual facts not covered by the checklist, averaging 6.8 atomic facts and spanning $\sim$10\% of the summary text. We therefore augment checklist evaluation with residual-fact evaluation to capture this overlooked content.

Besides evaluating summaries, it is also important to evaluate the reliability of automatic evaluators (i.e., meta-evaluation). One way is to ask humans to verify the outputs of the evaluator \citep{yen2024helmet,ruan2025expertlongbench}, but this is difficult to reuse for new evaluators. 
A more systematic way is to collect human annotations for the same tasks performed by automatic evaluators, enabling repeatable meta-evaluation across models.
We therefore collect 160 hours of human annotations.
Since human summaries are costly and often unavailable, we further take an initial step toward reference-free evaluation with \fwnameAgent, an agent scaffold for extracting checklist items directly from case documents. This provides the first study of whether tool-equipped LLMs can perform document-based checklist extraction for long legal cases.

\section{\fwname---Evaluating Long-Context Legal Summarization}
\label{sec:gavel-ref}

This section describes \fwnameRef, our reference-based evaluation framework, and \fwnameAgent, our agent scaffold for reference-free checklist extraction directly from source documents.

\subsection{\fwnameRef{}: Reference-Based Evaluation}
\label{subsec:gavel-ref-description}

\fwnameRef (Figure \ref{fig:gavel-ref}) consists of three components: checklist, residual-fact, and writing-style evaluation. We describe each component below. Prompts for the LLM judges are in Appendix \ref{app:prompts}.

\noindent \textbf{Checklist Evaluation.} Checklist evaluation has become a standard for domain-specific long-form generation tasks \citep{arora2025healthbench,ruan2025expertlongbench}, where outputs are expected to cover a well-defined set of facts. For each checklist item $c_i$, an LLM extracts the corresponding information from the model summary, $H(c_i)$, and from the human reference, $R(c_i)$, and then compares the two.
For our task, we use the 26 expert-designed legal checklist items from \cite{ruan2025expertlongbench}, covering information such as \texttt{filing date}, \texttt{parties}, and \texttt{remedies sought}. In Section \ref{sec:agent-in-medical}, we further design a 29-item checklist for medical review summarization to test whether \fwnameAgent generalizes beyond the legal domain.

We improve checklist extraction by representing each item as a list of values with supporting text rather than as a single text block, as we find 76\% of checklist items contain multiple values. Formally, we extract: $H(c_i) = \{(v_{i,1}, s_{i,1}), \ldots, (v_{i,n}, s_{i,n})\}$, where $v_{i,j}$ is the $j$-th extracted value and $s_{i,j}$ is its supporting text. This enables partial credit when only some values are captured and helps identify residual facts. We use an LLM to compare extracted items. Single-value items are classified into one of four relationships: equal, A contains B, B contains A, or different, with assigned matching scores $m_i$ of 1, 0.5, 0.5, and 0. Multi-value items are matched element-wise to compute $F_1$, which is used as their $m_i$.
To aggregate item-level comparisons into a final score, we pool values across all checklist items and compute value-level precision, recall, and F$_1$, so items with more values receive higher weight. Containment for a single-value item counts as 0.5. We use this F$1$ as $S_\text{checklist}$.

\vspace{2pt}
\noindent \textbf{Residual Facts Evaluation.} While the checklist captures core case information, we find 78\% of human summaries also include outside-checklist facts, averaging 6.8 per summary, such as case status information (e.g., ``The case is closed.'').
However, such facts are ignored by prior evaluations \citep{yen2024helmet,ruan2025expertlongbench}.
To evaluate them, we identify summary text not covered by checklist values or their supporting evidence, then use an LLM to extract atomic \textit{residual facts} from the uncovered text. We compare residual facts from model and human summaries using the same list-wise comparison method as in checklist evaluation.
The resulting $F_1$ score is used as $S_\text{residual}$.

\vspace{2pt}
\noindent \textbf{Writing Style Evaluation.} Beyond factual content, we evaluate how closely model summaries match the writing style of human references. An LLM rates similarity across five style aspects on a 1--5 Likert scale, which we average and rescale to 0--100 as $S_{\text{style}}$. Definitions of the five aspects are provided in Appendix \ref{appendix:writing_style_definitions}.

\vspace{2pt}
\noindent \textbf{Overall \fwnameRef{} Score.}
We combine checklist coverage, residual facts, and writing style into a single score:
\begin{equation}
\begin{aligned}
S_{\fwnameRef}
&= \alpha \bigl[
      (1-r)\, S_{\text{checklist}}
       \\
&\quad+ r\, S_{\text{residual}}
    \bigr] + (1-\alpha)\, S_{\text{style}}
\end{aligned}
\label{eq:gavel-ref-score}
\end{equation}
\noindent where $\alpha$ controls the balance between content and style, and $r$ sets the relative weight of $S_{\text{checklist}}$ and $S_{\text{residual}}$, It is computed as the proportion of residual content in the reference summary (i.e., the total length of residual text spans divided by the summary length) and averages 11\% across cases.
We set $\alpha=0.9$ throughout our paper.

\subsection{Gavel-Agent: Extracting Checklists from Case Documents}

While reference-based evaluation (\S\ref{subsec:gavel-ref-description}) effectively benchmarks LLMs, it relies on expert-written human summaries that are costly to collect. 
This raises a natural question: can we extract checklist items directly from case documents? To this end, we develop \fwnameAgent, an agent scaffold for source-based checklist extraction. Rather than reading every document exhaustively, \fwnameAgent lets LLMs navigate long case records by strategically searching, skimming, and updating checklist items, similar to how humans locate information in legal filings.

\begin{figure*}[!t]
  \centering
  \includegraphics[width=0.99\textwidth]{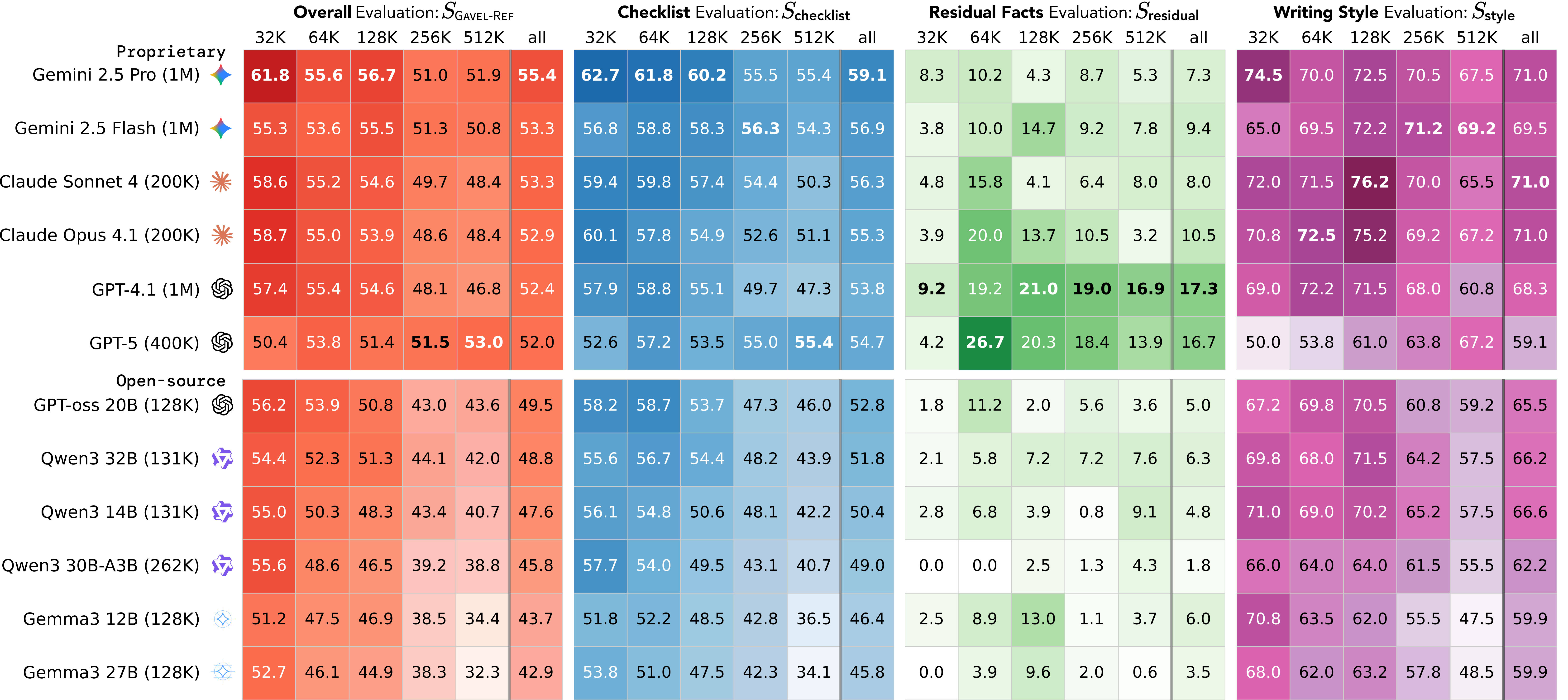}
  \vspace{-5pt}
  \caption{Benchmarking results of 12 LLMs on long-context legal summarization with our \fwnameRef framework across case lengths from 32K to 512K tokens. Models are ordered by $S_\text{\fwnameRef}$.}
  \label{fig:llm_eval_heatmap}
  \vspace{-12pt}
\end{figure*}

\vspace{2pt}
\noindent \textbf{Tools.} \fwnameAgent provides the following six tools. At each step, the LLM makes a tool call or stops based on the current state and history.
\begin{itemize}[topsep=0pt,itemsep=2pt,leftmargin=*,parsep=0pt]
\item \texttt{list\_documents()}: Returns all documents with metadata (e.g., document type, token count), providing an initial case overview.
\item \code{read_document(doc_name, start_token, end_token)}: Reads a specific token range from a document, with a maximum of 10,000 tokens.
\item \code{search_document_regex(pattern, doc_name, top_k, context_tokens)}: Searches one, multiple, or all documents using regex, returning top-k matches with surrounding context (100-1000 tokens).
\item \texttt{get\_checklist(item/items)}: Retrieves extracted values for specified checklist items.
\item \texttt{append\_checklist(patch)}: Adds new values for specific checklist items, supporting multiple values per item with required evidence.
\item \texttt{update\_checklist(patch)}: Replaces all values for specified checklist items, used for corrections or marking items as ``Not Applicable''.
\end{itemize}
Both \texttt{append\_checklist} and \texttt{update\_checklist} use a \texttt{patch} structure for batch updates. Each patch maps checklist items to extracted values, each paired with supporting evidence (verbatim text, source document, and location), ensuring traceability to the source documents.

\vspace{2pt}
\noindent \textbf{Context Management.}
Standard agent scaffolds append every tool call and response to the context, which becomes impractical for long cases with 256K+ tokens and 50+ calls. \fwnameAgent instead refreshes the context after each action.
At each step, the LLM is given a system prompt high-level task instruction and tool descriptions, and a user prompt that contains user instruction (e.g., ``Extract all 26 checklist items''), definitions of the items to extract, document catalog showing explored areas, extracted-item summary, and recent action history.
We keep up to 100 tool calls in history: the 5 most recent include full responses (e.g., full text from \texttt{read\_document}), while the other 95 are compressed to the tool name and brief outcome (e.g., ``read 3,000 tokens'', ``updated filing date''). This gives model enough awareness to avoid repeated actions while keeping the prompt compact.

\section{Evaluation of LLM Legal Summarization with \fwnameRef}
\label{sec:evaluation-of-llm}

Recent LLMs now support contexts of up to 1M tokens and have pretraining cutoffs up to 2025. To evaluate long-context summarization beyond the 128K-token range studied in prior work \citep{yen2024helmet,ruan2025expertlongbench}, we use recent legal cases from the Civil Rights Litigation Clearinghouse, which provides public case documents and expert-written summaries.
Using \fwnameRef, we evaluate 12 proprietary and open-source LLMs across five case-length scales: 32K, 64K, 128K, 256K, and 512K tokens, measured with the GPT-4o tokenizer. For each scale, we select 20 cases whose lengths fall within $\pm$20\% of the target length, yielding 100 cases across 18 case types (see Fig. \ref{fig:case_type_distribution}). 
Most are recent: 83 were filed in 2025, while 17 come from earlier years due to limited availability at longer lengths. Since models have different context limits, we truncate over-limit inputs by proportionally removing tokens from the end of each document, following prior work. We evaluate using the best open-source evaluator from our meta-evaluation (Table \ref{tab:meta-eval} in Section \ref{sec:meta-eval}).

\begin{figure*}[!t]
  \centering
  \includegraphics[width=0.99\textwidth]{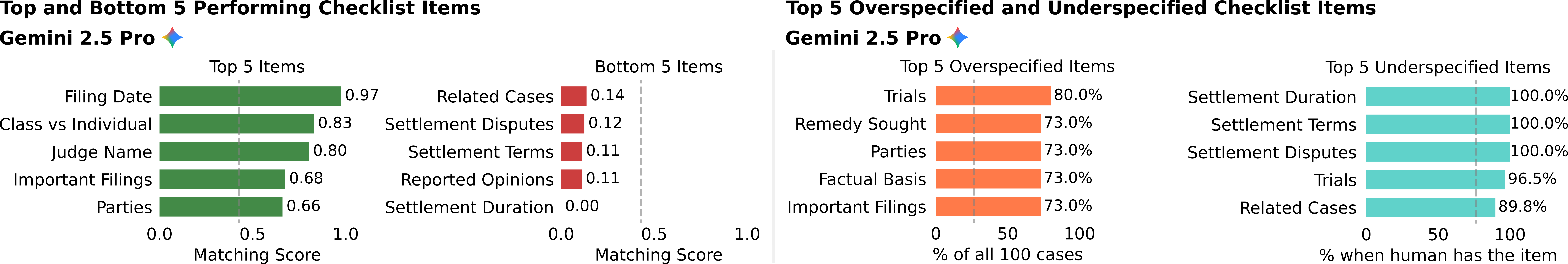}
  \vspace{-6pt}
  \caption{Gemini 2.5 Pro performance breakdown: top/bottom 5 checklist items by matching score and most frequently over/under-specified items. Overspecification measured as frequency across all 100 cases; underspecification as frequency among cases where human summary includes that item. Dashed lines are medians: 0.49 matching score, 59\% overspecification, 70\% underspecification.}
  \label{fig:llm_eval_by_items_gemini_25_pro}
  \vspace{-10pt}
\end{figure*}

\subsection{Benchmarking Results for 12 Models}

Figure \ref{fig:llm_eval_heatmap} shows \fwnameRef evaluation results for 12 models across different case length bins. Figure \ref{fig:precision_recall_length_heatmap} in the Appendix further breaks down checklist precision and recall, and compares each model’s summary length with the corresponding human summary length in each length bin.

\vspace{2pt}
\noindent \rev{\textbf{Gemini 2.5 Pro performs best, and the main failure mode of LLMs is underspecification.} 
Proprietary models consistently outperform open-source models by a clear margin. Overall, Gemini 2.5 Pro achieves the best performance, with an $S_\text{\fwnameRef}$ of 51.0, while the best open-source model, GPT-oss 20B, reaches 45.9. In checklist evaluation, 10 out of 12 models have precision at least 10 points higher than recall. For example, Gemini 2.5 Pro achieves 66.6 precision but only 55.3 recall, suggesting that models more often omit relevant information than hallucinate incorrect content.
Within the Claude family, Sonnet 4 slightly outperforms Opus 4.1. To understand which checklist items drive this gap, we present checklist item–level performance for each LLM in Figures \ref{fig:model_checklist_item_level_1}–\ref{fig:model_checklist_item_level_3} in the Appendix. We find that Sonnet 4 is stronger in identifying items such as Cause of action, Class action vs. individual, and Remedy sought than Opus 4.1.}

\vspace{2pt}
\noindent \rev{\textbf{Models degrade as case length increases, even with 1M-token context windows.}
Among models that support 1M-token contexts, Gemini 2.5 Pro, Gemini 2.5 Flash, and GPT-4.1 drop by 4--11 points in $S_\text{\fwnameRef}$ from the 32K to the 512K setting. GPT-5 is an outlier, but it performs worse overall across all length ranges. Open-source models degrade more sharply on 256K and 512K cases, as expected, since they do not support such long contexts and truncation causes substantial information loss. These results motivate scaffolded agents for long-context legal summarization.}

\vspace{2pt}
\noindent \rev{\textbf{GPT-4.1 performs best on residual facts evaluation, with GPT-5 close behind.} 
Both models tend to capture more non-checklist details than other models. On average, the residual ratio
$r$ (the proportion of residual content in the whole summary, Eq. \ref{eq:gavel-ref-score}) is 18.4\% for GPT-4.1 and 18.6\% for GPT-5 (as shown in Figure \ref{fig:model_residual_proportion}). These are the only two models that exceed the human residual ratio of 11.1\%.
As a result, GPT-4.1 and GPT-5 obtain the highest $S_\text{residual}$ of 17.3 and 16.7, respectively.}

\vspace{2pt}
\noindent \textbf{Surprisingly, GPT-5 has the lowest writing-style rating, while Gemini and Claude models are the most human-like.} As shown in Fig. \ref{fig:summary_example}, GPT-5 often ignores the prompt to write in narrative form, instead producing sectioned, checklist-style summaries. It is also verbose on 32K--128K cases, sometimes approaching 1,000 words when the corresponding human summary is around 700 words (Fig. \ref{fig:precision_recall_length_heatmap}). All models become less human-like on longer cases (256K--512K): human summaries average about 1,200 words, while proprietary models, excluding GPT-5, typically produce only 500--800 words, even with 1M-context LLMs.

\begin{figure}[t]
  \centering
  \includegraphics[width=\linewidth]{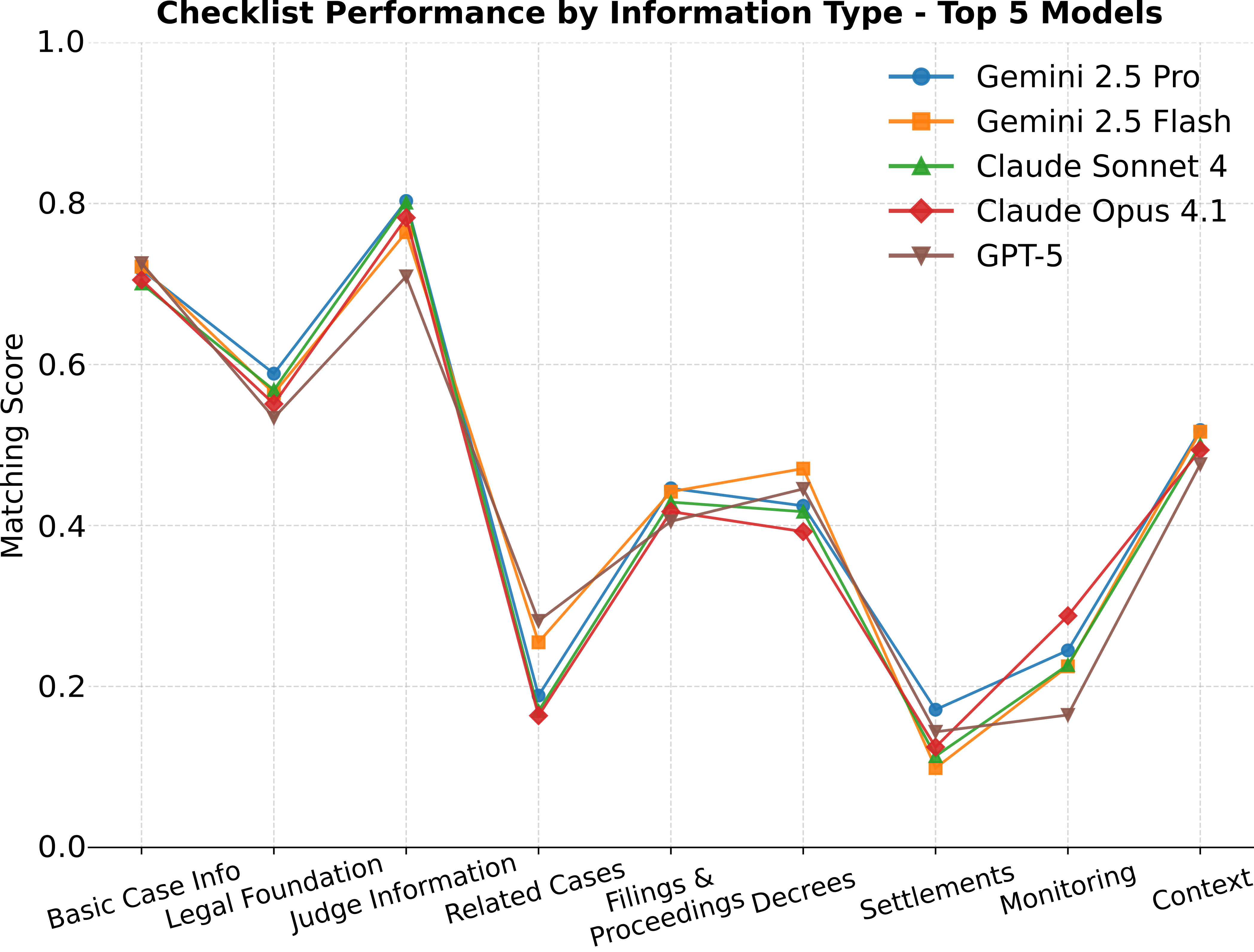}
  \vspace{-20pt}
  \caption{Top-5 LLMs' performance across checklist groups, struggling the most on rare items such as related cases and settlements.}
  \label{fig:checklist-groups}
  \vspace{-12pt}
\end{figure}

\subsection{How Top Models Handle Different Checklist Information}

Figure \ref{fig:checklist-groups} shows the performance of the top five models across nine checklist groups using the matching score $m_i$ (see \S\ref{sec:gavel-ref}). All models follow a similar pattern. \textbf{They are good at extracting basic case information, legal foundations, and judge details,} scoring above 0.6, as these groups contain mostly single-value items like filing date, cause of action, type of counsel, and judge name.
\textbf{Performance drops noticeably for multi-value items.} Court rulings, decrees, and factual basis (context) prove more challenging, with scores around 0.4-0.5. Models must track multiple related pieces of information scattered across lengthy documents and determine which ones are important enough to include.
\textbf{The models struggle most with related cases, settlements and monitoring,} scoring below 0.2. These items are rare in two senses: they appear in fewer cases---for example, settlement terms appear in only 8 out of 100 cases, and monitoring information appears in only 15 cases---and they appear less frequently within the case documents. This is because many selected 2025 cases are still ongoing and have not reached settlement or the monitoring stage. When these items do appear, they usually occur near the end of the case documents, unlike common items such as parties, which recur throughout the record.
We therefore investigate whether processing each case chunk by chunk and iteratively updating the checklist (the same method described later in \S\ref{subsec:gavel_implementation_details}) helps recover this information. It does: recall improves by 28.2 points on average---a relative gain of 60\%---over the average of the top-five proprietary LLMs' summaries, with consistent improvements across all nine information groups, as shown in Table~\ref{tab:recall-improvement}.

\begin{table}[H]
\centering
\setlength{\tabcolsep}{3pt}
\renewcommand{\arraystretch}{1.15}
\resizebox{\columnwidth}{!}{%
\begin{tabular}{@{}ccccc@{}}
\toprule
\shortstack{\rotatebox{35}{Basic Case Info}\\[2pt]+13.9} &
\shortstack{\rotatebox{35}{Legal Foundation}\\[2pt]+25.3} &
\shortstack{\rotatebox{35}{Judge Information}\\[2pt]+20.1} &
\shortstack{\rotatebox{35}{Related Cases}\\[2pt]+45.9} &
\shortstack{\rotatebox{35}{Filings}\\[2pt]+26.4} \\
\midrule
\shortstack{\rotatebox{35}{Decrees}\\[2pt]+30.4} &
\shortstack{\rotatebox{35}{Settlements}\\[2pt]+36.6} &
\shortstack{\rotatebox{35}{Monitoring}\\[2pt]+20.9} &
\shortstack{\rotatebox{35}{Context}\\[2pt]+34.6} &
\shortstack{\rotatebox{35}{Average}\\[2pt]+28.2} \\
\bottomrule
\end{tabular}%
}
\caption{Recall gains, in points, from chunk-by-chunk iterative processing of case documents over the average of the top-five LLM summary-based results, by information group.}
\vspace{-6pt}
\label{tab:recall-improvement}
\end{table}

\subsection{Dissecting the Top Performer}

Figure \ref{fig:llm_eval_by_items_gemini_25_pro} analyzes \rev{Gemini 2.5 Pro}'s item-level performance, showing its top and bottom 5 checklist items plus consistently over- and under-specified items (see Appendix Figure \ref{fig:llm_eval_by_items_full} for top-3 models).

\vspace{2pt}
\noindent \textbf{Single-value items are Gemini's strength, while settlement details are its blind spots.} 
\texttt{Filing date} has a near-perfect matching score of 0.97, followed by \texttt{Class action vs.\ Individual} (0.83) and \texttt{Judge name} (0.80). However, scores drop below 0.7 even for the next-best items, \texttt{Important Filings} and \texttt{Parties}, and the median score across all 26 items is only 0.43. Gemini performs worst on settlement-related items, scoring 0.12, 0.11, and 0.00 on the three settlement categories.

\vspace{2pt}
\noindent \textbf{Gemini 2.5 Flash tends to overspecify and underspecify checklist items with multiple values in its summaries.}
All top over- and under-specified items are multi-value items, with \texttt{Trials} appearing in both lists. \texttt{Settlement Duration}, \texttt{Settlement Terms}, and \texttt{Settlement Disputes} are under-specified 100\% of the time, showing that Gemini often misses settlement details. Overall, Gemini is much more prone to under-specification than over-specification, a trend across models, with median rates of 76.5\% and 26.5\%, respectively.

\section{Meta Evaluation of \fwname}
\label{sec:meta-eval}

To systematically meta-evaluate \fwname, we collect evaluator-independent human judgments for the same tasks performed by our automatic evaluators: \textit{checklist extraction}, \textit{checklist comparison}, \textit{residual-fact extraction}, and \textit{writing-style similarity rating}. This differs from prior work that asks annotators to verify a specific evaluator's outputs \citep{yen2024helmet,ruan2025expertlongbench}.
In total, we spend 160 hours of human effort collecting 5,442 checklist item-level annotations, 450 checklist comparison judgments, 281 extracted residual facts and 375 writing-style ratings. Appendix \ref{app:annotation_details} provides full annotation details, inter-annotator agreement, and interface screenshots.

\subsection{Metrics}
For \textit{checklist comparison}, we use accuracy for single-value items and matching-pairs F$_1$ for multi-value items.
We then use the best comparison model to evaluate \textit{checklist extraction} and \textit{residual-fact extraction}, computing $S_\text{checklist}$ against human-extracted checklist items from the same summary and macro F$_1$ against human-extracted residual texts from residual spans.
We also compute word-level coverage agreement on supporting text: how often model and human agree on whether words are covered by checklist items or are residual.
For \textit{writing style rating}, we report Cohen's Kappa for LLM-human agreement.

\begin{table}[t]
  \centering
  \resizebox{0.49\textwidth}{!}{
  \setlength{\tabcolsep}{2pt}
  \begin{tabular}{lccccccc}
    \toprule
    & \multicolumn{2}{c}{\rev{Checklist Extr.}} 
    & \multicolumn{2}{c}{Checklist Comp.} 
    & Resid. & Writ. \\
    \cmidrule(lr){2-3} \cmidrule(lr){4-5}
    Model & $S_{\text{chk}}$ & Cov. & Single & Multi & Fact & Style \\
    \midrule
    GPT-5          & \textbf{72.2} & \textbf{92.9\%} & 0.567 & 0.847 & 0.663 & 0.115 \\
    GPT-oss 20B    & 69.2 & 83.7\% & 0.567 & 0.801 & 0.653 & \textbf{0.157} \\
    Gemma3 27B     & 61.9 & 75.3\% & \textbf{0.740} & 0.841 & 0.546 & 0.091 \\
    Qwen3 32B      & 68.4 & 66.0\% & 0.600 & 0.820 & \textbf{0.699} & 0.084 \\
    Qwen3 30B-A3B  & 63.6 & 63.0\% & 0.700 & \textbf{0.854} & 0.663 & -0.011 \\
    \bottomrule
  \end{tabular}
  }
  \caption{Meta-evaluation of five LLMs in \fwnameRef. Metrics are $S_{\text{checklist}}$ and coverage for checklist extraction, accuracy and matching F$_1$ for single- and multi-value checklist comparison, F$_1$ for residual-fact extraction, and Cohen's $\kappa$ for writing style rating.}
  \vspace{-12pt}
  \label{tab:meta-eval}
\end{table}

\subsection{Implementation Details}
\label{subsec:gavel_implementation_details}

\noindent \textbf{Backbone LLMs for \fwnameRef.}
We select backbone evaluators based on two criteria: state-of-the-art performance and open-source availability. We evaluate five LLMs: GPT-5 and four open-source models---Qwen3 32B, Qwen3 30B-A3B, GPT-oss 20B, and Gemma3 27B.

\vspace{2pt}
\noindent \textbf{\fwnameAgent Configurations.}
We test three \fwnameAgent configurations to study whether agents should extract checklist items jointly or separately: one agent extracting all 26 items, nine agents extracting grouped items, and 26 agents each extracting a single item. We use Qwen3 30B-A3B and GPT-oss 20B, both of which natively support 128K+ context, sufficient for \fwnameAgent's context management.

\begin{figure}[t]
  \centering
  \includegraphics[width=\linewidth]{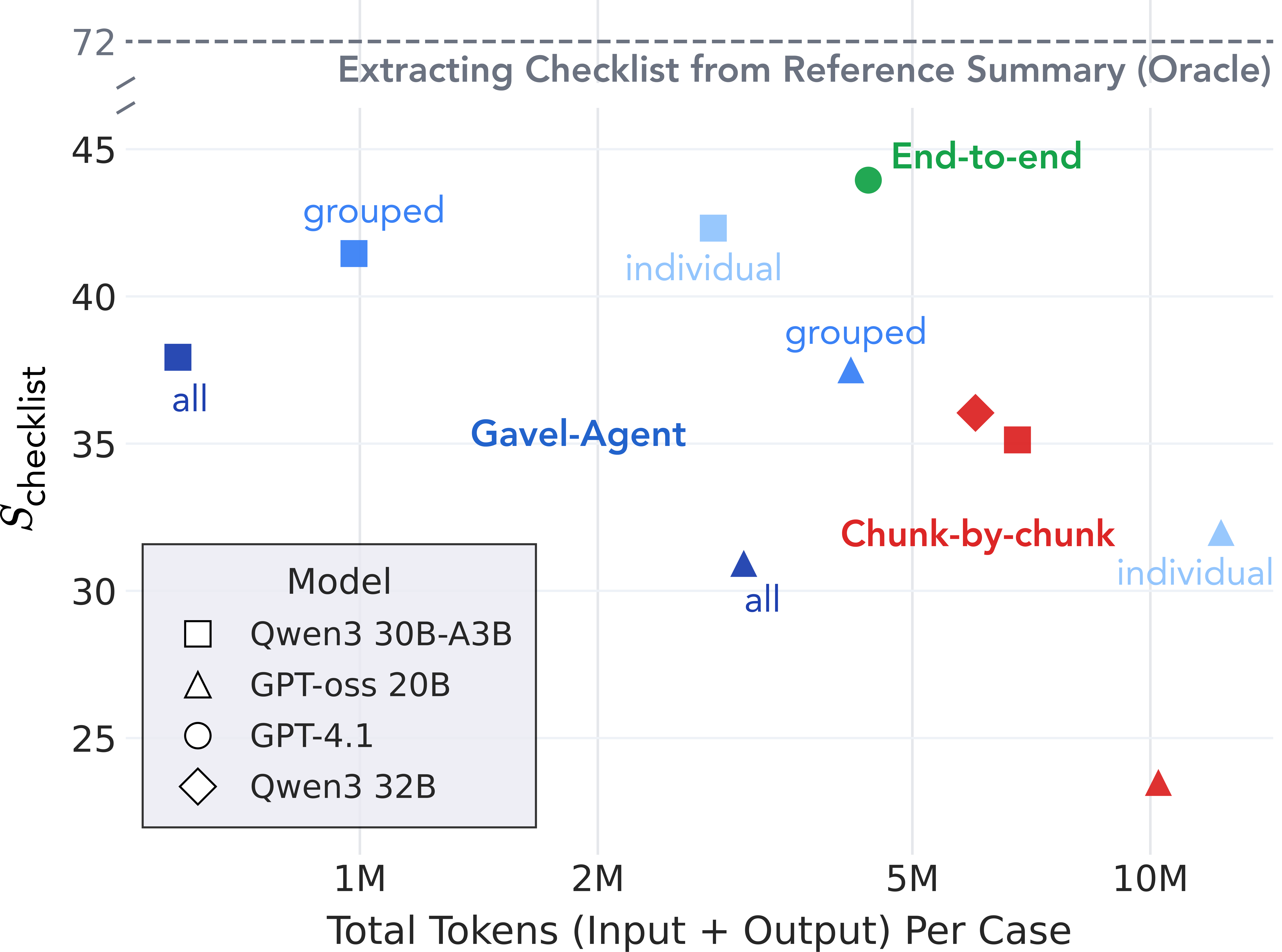}
    \vspace{-20pt}     
  \caption{$S_\text{checklist}$ versus total token usage for different methods extracting from legal case documents. \fwnameAgent achieves competitive performance while using 36\% fewer tokens.}
  \label{fig:meta_eval_from_docs_total_tokens}
  \vspace{-15pt}
\end{figure}

\vspace{2pt}
\noindent \textbf{Document-Based Extraction Baselines.}
We compare \fwnameAgent with the following two non-agentic baselines for extracting checklist items directly from case documents.

\noindent \textbf{1. End-to-end extraction} concatenates all case documents in chronological order and feeds them to a long-context LLM. We use GPT-4.1 because of its 1M-token context window. We query each checklist item separately rather than extracting all 26 items at once, which gives more accurate results.

\noindent \textbf{2. Chunk-by-chunk extraction} splits documents into 16K-token chunks and processes them chronologically. At each step, the model receives the current chunk and checklist state, updating it by retaining existing values or adding new ones. As in end-to-end extraction, we query each checklist item separately. We evaluate GPT-oss 20B, Qwen3 32B, and Qwen3 30B-A3B. This mirrors long-context methods that process segmented text iteratively \citep{zhang2024chain,zhao-etal-2024-longagent}.

\subsection{Results.}

Table \ref{tab:meta-eval} presents the meta-evaluation results for \fwnameRef. Different models perform best on different evaluator components: GPT-5 leads checklist extraction, Gemma3 27B and Qwen3 30B perform best on single- and multi-value checklist comparison, Qwen3 32B leads residual-fact extraction, and GPT-oss 20B aligns best with human writing-style judgments.
Figure \ref{fig:meta_eval_from_docs_total_tokens} compares document-based checklist extraction methods by $S_{\text{checklist}}$ versus total token usage.
Table~\ref{tab:tool-usage} presents the tool usage across all  \fwnameAgent configurations.
The main findings are:

\vspace{2pt}
\noindent \textbf{1. Document-based checklist extraction remains much harder than oracle summary-based extraction.} The best document-based method, end-to-end extraction with GPT-4.1, reaches $S_{\text{checklist}}=46.9$, below GPT-5 oracle extraction from case summaries at 72.2, showing that extracting checklist items directly from long case documents remains substantially more difficult.

\begin{table}[!t]
  \centering
  \resizebox{0.49\textwidth}{!}{
  \setlength{\tabcolsep}{3pt}
  \begin{tabular}{lrrrrrr}
    \toprule
    & \multicolumn{3}{c}{Navigation}
    & \multicolumn{3}{c}{Checklist} \\
    \cmidrule(lr){2-4} \cmidrule(lr){5-7}
    Config & List & Read & Search & Get & Append & Update \\
    \midrule
    \multicolumn{7}{l}{\textbf{GPT-oss 20B}} \\
    \quad All
      & 40 & 1,211 & 5,490 & 17 & 165 & 11 \\
    \quad Grouped
      & 376 & 3,008 & 9,746 & 406 & 363 & 260 \\
    \quad Individual
      & 1,076 & 8,072 & 28,975 & 1,764 & 1,455 & 871 \\
    \addlinespace
    \multicolumn{7}{l}{\textbf{Qwen3 30B-A3B}} \\
    \quad All
      & 40 & 420 & 21 & 45 & 200 & 48 \\
    \quad Grouped
      & 360 & 834 & 123 & 366 & 370 & 163 \\
    \quad Individual
      & 1,040 & 2,551 & 504 & 1,074 & 1,039 & 362 \\
    \bottomrule
  \end{tabular}
  }
  \caption{Number of tool calls across all Gavel-Agent configs. The tools are \texttt{list\_documents} (List), \texttt{read\_document} (Read), \texttt{search\_document\_regex} (Search),  \texttt{get\_checklist} (Get), \texttt{append\_checklist} (Append), \texttt{update\_checklist} (Update).}
  \vspace{-12pt}
  \label{tab:tool-usage}
\end{table}

\vspace{2pt}
\noindent \textbf{2. \fwnameAgent achieves competitive performance with lower token cost.} \fwnameAgent with 26 individual agents using Qwen3 30B-A3B achieves the second-best document-grounded score of 43.5 while using only 2.8M tokens---36\% fewer than end-to-end GPT-4.1 and 59\% fewer than chunk-by-chunk extraction with the same Qwen3 model. Its multi-agent decomposition also outperforms a single agent extracting many items jointly, suggesting that item-level specialization helps agents navigate long case records.

\vspace{2pt}
\noindent \textbf{3. Chunk-by-chunk extraction has high recall but suffers from error accumulation.} The best chunk-by-chunk method, Qwen3 30B-A3B, scores 38.8, substantially below both end-to-end extraction and \fwnameAgent. The gap is mainly due to iterative update errors: incorrect values persist across chunks and cause over-extraction, as reflected in the ``Ref Empty, Model Not'' column in Figure \ref{fig:by-item-from-documents-chunk-by-chunk}.
However, because the method processes every chunk, it achieves very high recall (Fig. \ref{fig:precision_vs_recall_from_docs}).

\vspace{2pt}
\noindent \textbf{4. Different models adopt different document-navigation strategies.}
For the navigation tools (\texttt{list\_documents}, \texttt{read\_document}, and \texttt{search\_document\_regex}), \texttt{list\_documents} is called approximately once per agent run to obtain an initial overview of the case record.
After this initial step, the two models behave very differently. GPT-oss 20B relies heavily on regex search, which accounts for 69--79\% of its tool calls across configurations.
In contrast, Qwen3 30B-A3B primarily reads documents directly and makes 6.6 times fewer tool calls overall.
The checklist-management tools (\texttt{get\_checklist}, \texttt{append\_checklist}, and \texttt{update\_checklist}) are used more often in the grouped and individual configurations. In these settings, separate extraction tasks must repeatedly inspect and update a shared checklist.

\section{\fwnameAgent in the Medical Domain}
\label{sec:agent-in-medical}

To test whether \fwnameAgent can generalize beyond legal cases, we extend it to the medical domain \citep{joseph2024factpico}. Specifically, we apply \fwnameAgent to extract checklist items from medical reviews, which are long documents that synthesize evidence on specific health questions, such as whether an intervention is effective for a disease. We choose this domain because of the availability of recent medical reviews from 2025 and plain language summaries (PLS) written by health professionals.

\vspace{2pt}
\noindent \textbf{Experiment Setup.}
We collect 10 medical reviews from 2025 from Cochrane, an organization that publishes systematic reviews of research on human health and care.
The selected reviews average about 60K tokens, on the higher end of typical review length, making the task harder.
To define the evaluation checklist, we start from Cochrane's plain language summary (PLS) guidelines, which are designed to make medical evidence understandable to non-specialist readers. Because the guidelines describe the information expected in a PLS, checklist construction requires biomedical literacy and careful comparison across examples, rather than expert clinical judgment. We therefore recruit an undergraduate biology major with relevant domain background, who uses the guidelines as a starting point and refines the checklist by cross-checking the information included across multiple PLS examples.
This process takes 8 hours.
The checklist contains 29 items, covering information such as the health condition, treatment, and benefits, with definitions in Appendix \ref{app:medical_checklist_definitions}.
The student then extracts checklist items from the PLS of the 10 reviews, yielding 427 checklist-item annotations.
To adapt this setting into a multi-document format, we split each review by section, such as background, methods, and results, and treat each section as a separate document. Since \fwnameAgent is designed to be extensible, we only change the checklist definition and otherwise use the same experimental setup as in the legal domain.

\begin{figure}[t]
  \centering
  \includegraphics[width=\linewidth]{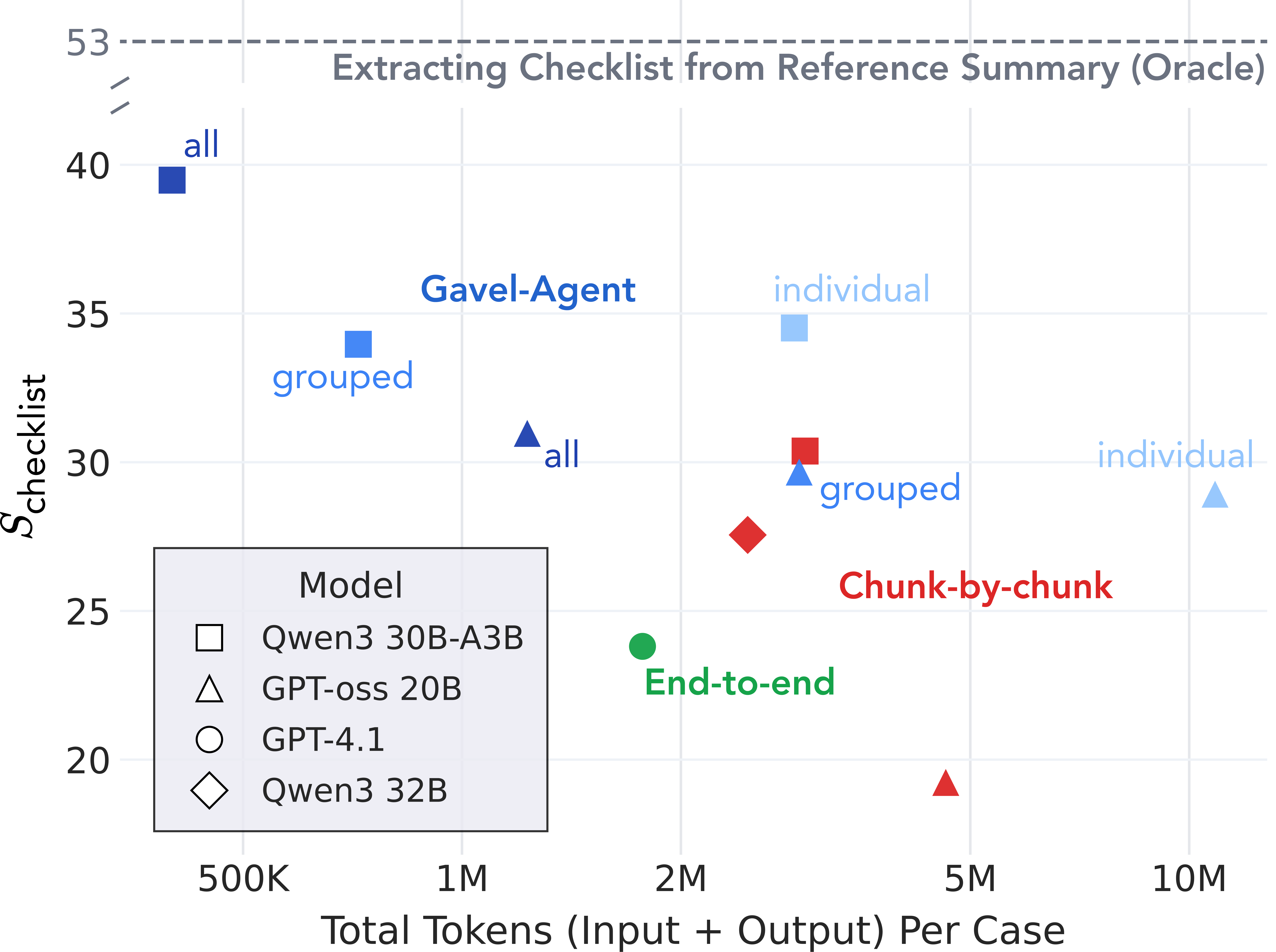}
    \vspace{-20pt}     
  \caption{$S_\text{checklist}$ versus total token usage for different checklist extraction methods on medical review. \fwnameAgent achieves top performance while using 77\% fewer tokens.}
  \label{fig:meta_eval_from_docs_total_tokens_medical}
  \vspace{-15pt}
\end{figure}

\vspace{2pt}
\noindent \textbf{Results.}
Figure \ref{fig:meta_eval_from_docs_total_tokens_medical} compares checklist extraction performance and token usage on medical reviews. \fwnameAgent performs best with an $S_\text{checklist}$ score of 39.5, evaluated against human checklist items extracted from the PLS. At the same time, it uses substantially fewer tokens: 77\% fewer than the end-to-end method and 83\% fewer than the chunk-by-chunk method. As in the legal domain, \fwnameAgent achieves higher precision but lower recall than other methods (see Figure \ref{fig:precision_vs_recall_from_docs_medical}).

\section{Conclusion}

We introduce \fwname, which includes \fwnameRef, a reference-based framework for evaluating long-context legal summarization that augments checklist evaluation with residual fact and writing-style evaluation, and \fwnameAgent, an agent scaffold for reference-free evaluation that extracts checklists directly from case documents. Using \fwnameRef, we evaluate 12 frontier LLMs on 2025 legal cases spanning 32K--512K tokens. We find that current models mainly fail by omitting key information rather than hallucinating, especially as case length increases. Models perform well on simple single-value items but struggle with multi-value and rare items. To meta-evaluate \fwname, we collect 160 hours of human annotations. We find that \fwnameAgent achieves competitive performance while using 36--59\% fewer tokens than end-to-end and chunk-by-chunk approaches. Finally, we show that \fwnameAgent extends to the medical domain, achieving the best performance with at least 77\% fewer tokens than baselines.

\section*{Limitations}
This work primarily focuses on the evaluation of legal summarization rather than on improving summarization models themselves. Exploring methods that directly improve legal summarization---such as first extracting structured checklists from case documents and then generating summaries conditioned on those checklists---could further enhance summary quality, and we leave this direction to future work.
Due to cost constraints, we do not apply \fwnameAgent on the strongest closed-source models, such as GPT-5.5 or Claude 4.7. Nevertheless, our results show that even with an open-source model like Qwen3 30B, \fwnameAgent approaches the performance of end-to-end extraction using GPT-4.1, suggesting substantial headroom for agent-based approaches.
Finally, our experiments indicate that a single agent handling all 26 checklist items performs poorly, as this setting effectively turns the task into a long-horizon problem. Future work could explore better agent architectures, such as planning agent or using LLMs to automatically design and spawn specialized sub-agents, to better handle long-horizon tasks.

\subsubsection*{Acknowledgments}
We thank Alexey Plagov, Sara Takagi, Jerry Lou Zheng, and Shannon Shen for their contributions. We are also grateful to Charlotte Alexander, Betsy DiSalvo, Doug Downey, and Jessy Junyi Li for valuable discussions. We thank Duong Minh Le, Jonathan Zheng, and Geyang Guo for their feedback. This research is supported in part by a Google Faculty Academic Research Award, the NIH Award R01LM014600, and the NSF CAREER Award IIS-2144493. Any opinions, findings, conclusions, or recommendations expressed in this material are those of the authors and do not necessarily reflect the views of Google, the National Institutes of Health, the National Science Foundation, or the U.S. Government. We also thank OpenAI for providing API credits to support this work.

\bibliography{custom}

\clearpage
\appendix

\section{Large Language Model Usage in Paper Writing}
We use LLMs solely for language polishing purposes: grammar correction and paraphrasing to improve clarity and readability. We do not use LLMs to generate new content. All semantic content and scientific contributions originate entirely from the authors.
\section{Additional Related Work}
\label{app:additional_related_work}

\textbf{Legal Summarization.}
Several datasets exist for this task. \citet{shukla2022legal} release Indian and UK Supreme Court cases with human-written summaries, \citet{elaraby-litman-2022-arglegalsumm} provide Canadian court opinions paired with expert summaries, and \citet{heddaya2024casesumm} collect U.S. Supreme Court opinions with official summaries. These resources focus on single-document summarization with inputs under 16K tokens. Multi-LexSum \citep{shen2022multi} and ExpertLongBench \citep{ruan2025expertlongbench} extend this to multi-document setting using cases from the Civil Rights Litigation Clearinghouse (CRLC), which offers public access to U.S. civil rights cases.
Following them, we collect cases from CRLC, focusing on 2025 filings to reduce data contamination. We evaluate 12 frontier LLMs with \fwnameRef on five length bins (32K–512K tokens) and provide fine-grained analysis beyond the aggregate scores reported in prior benchmarks \cite{yen2024helmet,ruan2025expertlongbench}.

\vspace{2pt}
\noindent\textbf{LLM Agent Scaffolds.}
Modern LLM agents are designed as autonomous problem-solvers that plan actions and invoke tools in a multi-step loop for tasks such as web browsing \citep{gur2023real}, coding \citep{yang2024swe}, or general-purpose reasoning. Several open-source scaffolds have been introduced \citep{xie2023openagents,wang2025openhands,lu2025octotools,qiu2025alita}.
For long-context processing, recent approaches segment documents into chunks or convert them into graph structures \citep{chen2023walking,sun-etal-2024-pearl,li2024graphreader,zhao-etal-2024-longagent,zhang2024chain}, which we adopt as our chunk-by-chunk method.
Inspired by how human experts read documents---skimming titles, prioritizing files, and searching for keywords rather than read exhaustively---we develop \fwnameAgent, an autonomous scaffold that equips models with six tools for navigating legal documents.

\section{Checklist Definitions}
\label{app:checklist_definitions}

The following are the definitions of the 26 checklist items used in our work, which are adapted from ExpertLongBench \citep{ruan2025expertlongbench}. We group them into 9 groups.

\begin{enumerate}[leftmargin=*,itemsep=1pt,parsep=1pt,topsep=2pt]

\item[\textbf{A.}] \textbf{Basic Case Information}
\begin{enumerate}[label=\arabic*.,itemsep=0pt,parsep=0pt,topsep=1pt]
    \item \textbf{Filing Date}: The date when the lawsuit was first initiated with the court.
    \item \textbf{Parties}: Description of each plaintiff and defendant involved, including relevant positions or offices held. Use specific terms (e.g., ``The city'', ``The parents'') rather than generic terms (e.g., ``The defendant'', ``The plaintiffs'').
    \item \textbf{Class Action or Individual Plaintiffs}: Whether the case involves class action plaintiffs or individual plaintiffs with descriptions.
    \item \rev{\textbf{Type of Counsel}: The type(s) of counsel representing each side. Use brief category labels (e.g., private counsel, public interest nonprofit, government counsel, pro se) and include specific organizations (if applicable) in parentheses (e.g., Public interest nonprofit (ACLU)).}
\end{enumerate}

\item[\textbf{B.}] \textbf{Legal Foundation}
\begin{enumerate}[label=\arabic*.,resume,itemsep=0pt,parsep=0pt,topsep=1pt]
    \item \rev{\textbf{Cause of Action}:  The legal vehicle(s) used to bring the claims (the ``how'' of suing), such as statutes that create a private/enforcement right of action (e.g., 42 U.S.C. § 1983, Title II ADA, FTCA) or judge-made vehicles (e.g., Bivens).}
    \item \rev{\textbf{Statutory/Constitutional Basis}: The substantive rights and sources of law allegedly violated (the 'what' was violated), such as specific constitutional provisions/clauses (e.g., Fourteenth Amendment---Equal Protection, First Amendment---Freedom of Association, Eighth Amendment) and statutory rights (e.g., ADA Title II, Rehab Act § 504).}
    \item \rev{\textbf{Remedy Sought}: What each party asks the court to grant, not what the court ordered or what the parties settled. Include both sides if the defendant seeks relief.}
\end{enumerate}

\item[\textbf{C.}] \textbf{Judge Information}
\begin{enumerate}[label=\arabic*.,resume,itemsep=0pt,parsep=0pt,topsep=1pt]
    \item \rev{\textbf{Judge Name}: The first and last name of the judge(s) involved in the case. Do not include Supreme Court Justices.}
\end{enumerate}

\item[\textbf{D.}] \textbf{Related Cases}
\begin{enumerate}[label=\arabic*.,resume,itemsep=0pt,parsep=0pt,topsep=1pt]
    \item \textbf{Consolidated Cases}: Cases that were combined with this case for joint proceedings.
    \item \textbf{Related Cases}: Other cases referenced or connected to this case, listed by case code number.
\end{enumerate}

\item[\textbf{E.}] \textbf{Filings and Proceedings}
\begin{enumerate}[label=\arabic*.,resume,itemsep=0pt,parsep=0pt,topsep=1pt]
    \item \textbf{Important Filings}: Significant motions filed, including temporary restraining orders, preliminary injunctions, motions to dismiss, and motions for summary judgment.
    \item \textbf{Court Rulings}: Judicial decisions on important filings such as motions to dismiss, summary judgment, preliminary injunctions, class certification, and attorneys' fees (excluding amended complaints and statements of interest).
    \item \textbf{Reported Opinions}: Citations of reported opinions using shortened Bluebook format (e.g., ``2020 WL 4218003''), without case name, court, or date unless from a different case.
    \item \textbf{Trials}: Information about trial proceedings including scheduling, outcomes, and related motions or rulings.
    \item \textbf{Appeals}: Whether appeals were filed, which parties appealed, to which court, and the outcomes.
\end{enumerate}

\item[\textbf{F.}] \textbf{Decrees}
\begin{enumerate}[label=\arabic*.,resume,itemsep=0pt,parsep=0pt,topsep=1pt]
    \item \rev{\textbf{Significant Terms}: The substantive obligations ordered by the court. This includes consent decrees and stipulated judgments/injunctions because they are entered as court orders.
    \item \rev{\textbf{Decree Dates}: All decree-related dates such as entry date, modification/amendment dates (of the order), suspension/stay dates, partial termination dates, and full termination/vacatur dates. Decrees include injunctions, consent decrees, or stipulated judgments/injunctions.}}
    \item \rev{\textbf{Duration}: The duration of all decrees obligations (each as a separate entry). A `decree' is any formal order or judgment issued by a court such as an injunction, consent decree, or stipulated judgment/injunction, as opposed to a negotiated agreement between parties.}
\end{enumerate}

\item[\textbf{G.}] \textbf{Settlements}
\begin{enumerate}[label=\arabic*.,resume,itemsep=0pt,parsep=0pt,topsep=1pt]
    \item \rev{\textbf{Settlement Terms}: The substantive obligations the parties agree to in a settlement that is not entered as a court order. A settlement may be court-approved or enforced, but as long as it is not entered as an order, it is a settlement.}
    \item \rev{\textbf{Settlement Date}: All settlement-related dates (each as a separate entry) such as execution/signing date(s), court approval date (if approved but not entered as an order), amendment dates, enforcement/retention dates without incorporation (e.g., court retains jurisdiction over the settlement but does not enter it as an order), and termination/expiration of the settlement agreement (if contractual).}
    \item \rev{\textbf{Duration}: The duration of all settlements obligations (each as a separate entry). A 'settlement' is any negotiated agreement between parties that resolves a dispute, as opposed to a formal order or judgment issued by a court.}
    \item \rev{\textbf{Court Enforcement}: Whether the settlement (not entered as an order/judgment) is court-enforced. Answer Yes if the court explicitly retains jurisdiction to enforce the settlement without incorporating it into an order/judgment (e.g., Kokkonen retention). Answer No if it's a private agreement with no retained jurisdiction.}
    \item \rev{\textbf{Enforcement Disputes}:  The disputes about enforcing a settlement (a negotiated agreement not entered as a court order)---e.g., motions to enforce/contempt or requests invoking retained jurisdiction---each as a separate value with date, movant, issue, and outcome (or pending).}
\end{enumerate}

\item[\textbf{H.}] \textbf{Monitoring}
\begin{enumerate}[label=\arabic*.,resume,itemsep=0pt,parsep=0pt,topsep=1pt]
    \item \textbf{Monitor Name}: Name of any court-appointed monitor or special master.
    \item \textbf{Monitor Reports}: Monitor's findings regarding defendant compliance with court orders, including which terms are being met.
\end{enumerate}

\item[\textbf{I.}] \textbf{Context}
\begin{enumerate}[label=\arabic*.,resume,itemsep=0pt,parsep=0pt,topsep=1pt]
    \item \textbf{Factual Basis}: The underlying facts and evidence supporting the legal claims, including: (i) details of relevant events (what, when, where, who), (ii) supporting evidence (physical, documentary, testimonial), and (iii) background context.
\end{enumerate}

\end{enumerate}

\section{Writing Style Similarity Evaluation Details}
\label{appendix:writing_style_definitions}

The following are the definitions of the five aspects used in our writing style similarity evaluation. Each aspect is rated on a 1–5 Likert scale, where 5 indicates identical and 1 indicates completely different.

\begin{enumerate}[leftmargin=*,itemsep=0.4em]

\item \textbf{Readability \& Jargon Level} \\
Compare the reading level and the balance of legal jargon vs.\ plain language. Consider terminology density and accessibility to non-legal readers.
\begin{itemize}[leftmargin=*,itemsep=0em,topsep=0.2em]
  \item[5] Nearly identical reading level and jargon density; same balance of technical/plain language throughout.
  \item[4] Very similar complexity with minor differences in terminology or occasional variance in technical language.
  \item[3] Moderate differences in accessibility; one is noticeably more technical in places but overall similar.
  \item[2] Significantly different complexity; one is consistently more technical or more accessible.
  \item[1] Completely different target audiences (e.g., one for legal professionals, the other for the general public).
\end{itemize}

\item \textbf{Narrative Order} \\
Compare whether events are presented in the same sequence (chronological vs.\ thematic) and the ordering of key facts and arguments.
\begin{itemize}[leftmargin=*,itemsep=0em,topsep=0.2em]
  \item[5] Identical sequence of information; same events, facts, and arguments in the same order.
  \item[4] Same overall flow with 1--2 elements reordered; core structure preserved.
  \item[3] Similar general structure but several sections reordered; recognizable yet rearranged.
  \item[2] Different organizational approaches with some overlap (mix of chronological and thematic).
  \item[1] Completely different information architecture (e.g., one chronological, the other organized by issues).
\end{itemize}

\item \textbf{Sentence Structure \& Voice} \\
Compare sentence variety, active vs.\ passive voice, and tense consistency.
\begin{itemize}[leftmargin=*,itemsep=0em,topsep=0.2em]
  \item[5] Nearly identical sentence patterns, voice usage, and tense choices throughout.
  \item[4] Very similar style with occasional differences in sentence complexity or voice.
  \item[3] Moderate variation; one favors longer/shorter sentences or more active/passive constructions.
  \item[2] Noticeably different styles; consistent differences in sentence variety and voice preferences.
  \item[1] Completely different approaches (e.g., one varied and active; the other uniform and passive).
\end{itemize}

\item \textbf{Formatting \& Layout} \\
Compare use of headings, bullet/numbered lists, paragraphing, and other structural cues.
\begin{itemize}[leftmargin=*,itemsep=0em,topsep=0.2em]
  \item[5] Identical formatting choices; same use of headings, lists, and paragraph breaks.
  \item[4] Very similar structure with minor variations (e.g., one extra heading or different list style).
  \item[3] Similar approach but noticeable differences in execution (e.g., both use headings but at different levels/frequency).
  \item[2] Different formatting philosophies; one is much more structured than the other.
  \item[1] Completely different (e.g., one heavily formatted; the other continuous prose).
\end{itemize}

\item \textbf{Citation \& Reference Style} \\
Compare presence, position, and formatting of case/statute citations or footnotes (inline vs.\ separate), citation density, and conventions.
\begin{itemize}[leftmargin=*,itemsep=0em,topsep=0.2em]
  \item[5] Identical citation approach; same style, frequency, and positioning.
  \item[4] Very similar practices with minor formatting differences or occasional variation in placement.
  \item[3] Similar philosophy but different execution (e.g., both cite cases but differ in density/positioning).
  \item[2] Different approaches; one is substantially more reference-heavy or uses a different citation style.
  \item[1] Completely different or incomparable (e.g., one with extensive citations, the other with none).
\end{itemize}

\end{enumerate}
\section{Annotation Details}
\label{app:annotation_details}

\paragraph{Annotator Recruitment.}
We recruit four in-house annotators who are native English speakers and U.S.-based undergraduate students with basic familiarity with legal cases. All annotators are trained by the authors: we review the 26 checklist items together, ensure that everyone understands the legal terms involved (e.g., decree, settlement, ruling), and walk through example annotations. Because their task is to extract checklist items from case summaries that are written for lay readers rather than to provide legal judgments or read case documents, we do not require formal legal training once they clearly understand each checklist item and its definition. All annotators provided informed consent to the release of their annotations for research purposes.

\paragraph{Annotation Procedure.}
\rev{To evaluate LLMs' ability to \textit{extract checklist items}, we annotated 40 long case summaries (avg. 1,130 words) to stress-test the models: if the LLM can accurately extract checklist items from these longer summaries, it should perform at least as well on the shorter ones used in the main model evaluation.} Since extracting all 26 checklist items from scratch is time-consuming, annotators start from GPT-5’s extractions.
Using our paragraph-by-paragraph review interface modified from Thresh \citep{heineman-etal-2023-thresh}, annotators add missing values, correct extractions and supporting text, or delete incorrect values. Each summary annotation takes approximately one hour. Figures \ref{fig:checklist_example_1} to \ref{fig:checklist_example_10} show an example of our annotations on a case summary, covering all 26 checklist items.
In total, we collect \rev{70} summary-level annotations covering \rev{5,442} item-level annotations, where the ten longest summaries (averaging \rev{1,695} words) receive triple annotations, with adjudication by a fourth annotator. The remaining \rev{30} summaries receive single annotations.
To evaluate LLMs' ability to \textit{compare checklist values}, annotators assess 150 item pairs from model and reference summaries (100 multi-value, 50 single-value), drawn from diverse LLMs for generalizability.
For single-value pairs, they perform 4-class classification: equal, A contains B, B contains A, or different. For multi-value pairs, they match elements from list A to list B. Annotations are aggregated by majority vote: for single-value items, we take the class with $\geq$ two votes (no cases had all three labels differ); for multi-value items, we keep matches identified by $\geq$ two annotators.
To evaluate LLM's ability to \textit{extract residual facts} from residual spans, an annotator annotates for 100 instances with 281 residual facts annotated.
To evaluate LLM's ability to \textit{rate writing style similarity}, we annotate 25 model-reference summary pairs. Annotators rate similarity across five style aspects using 1-5 Likert scales, with three annotations per pair. Final scores are the median across annotators.
All annotators are paid \$18 USD per hour, with a total cost of \rev{\$3K} USD.

\begin{figure}
  \centering
  \includegraphics[width=\linewidth]{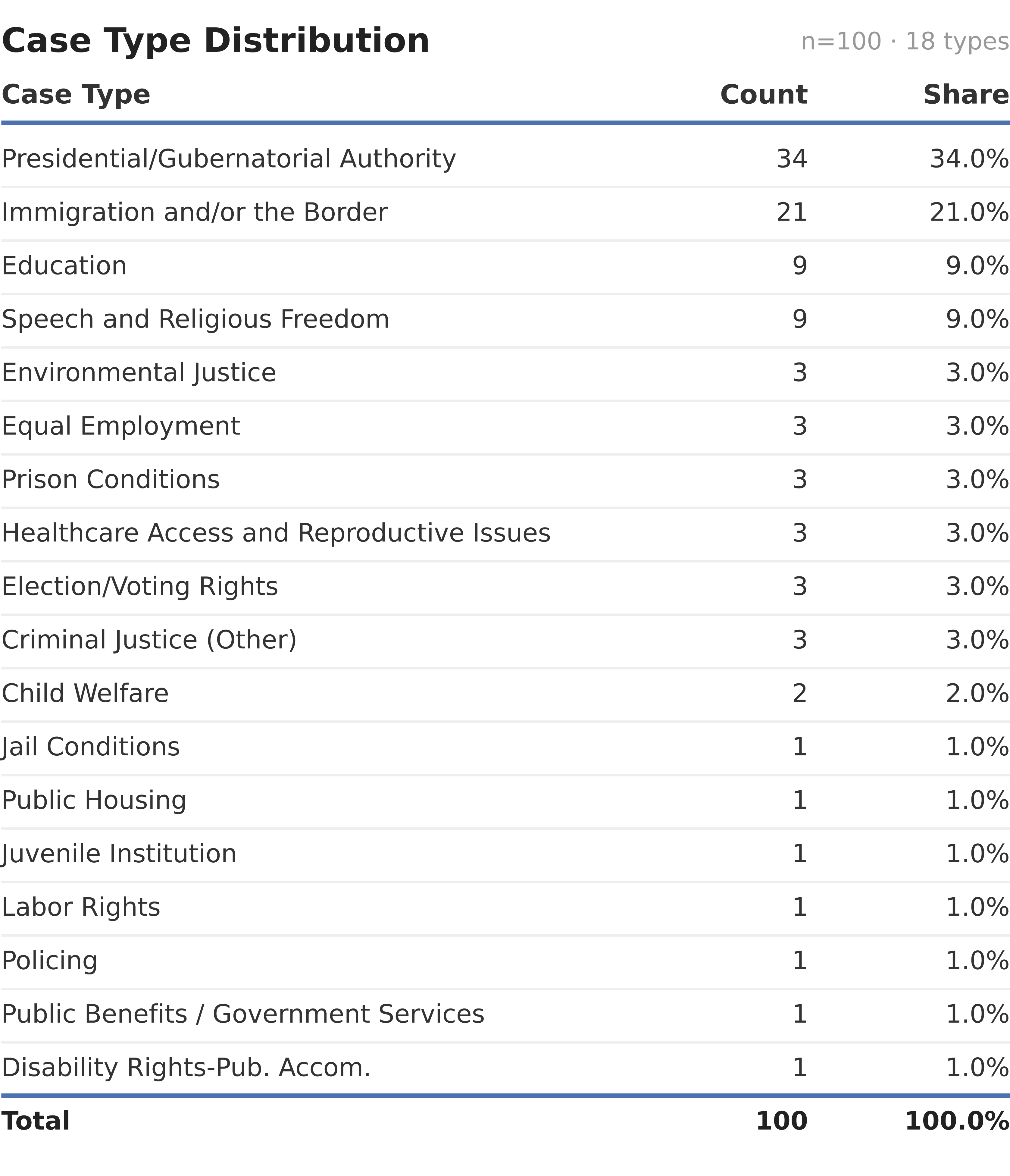}
  \vspace{-15pt}
  \caption{Distribution of case types across the 100 cases used to evaluate LLM summarization performance in Section \ref{sec:evaluation-of-llm}.}
  \label{fig:case_type_distribution}
\end{figure}

\begin{figure*}[!t]
  \centering
  \includegraphics[width=0.99\textwidth]{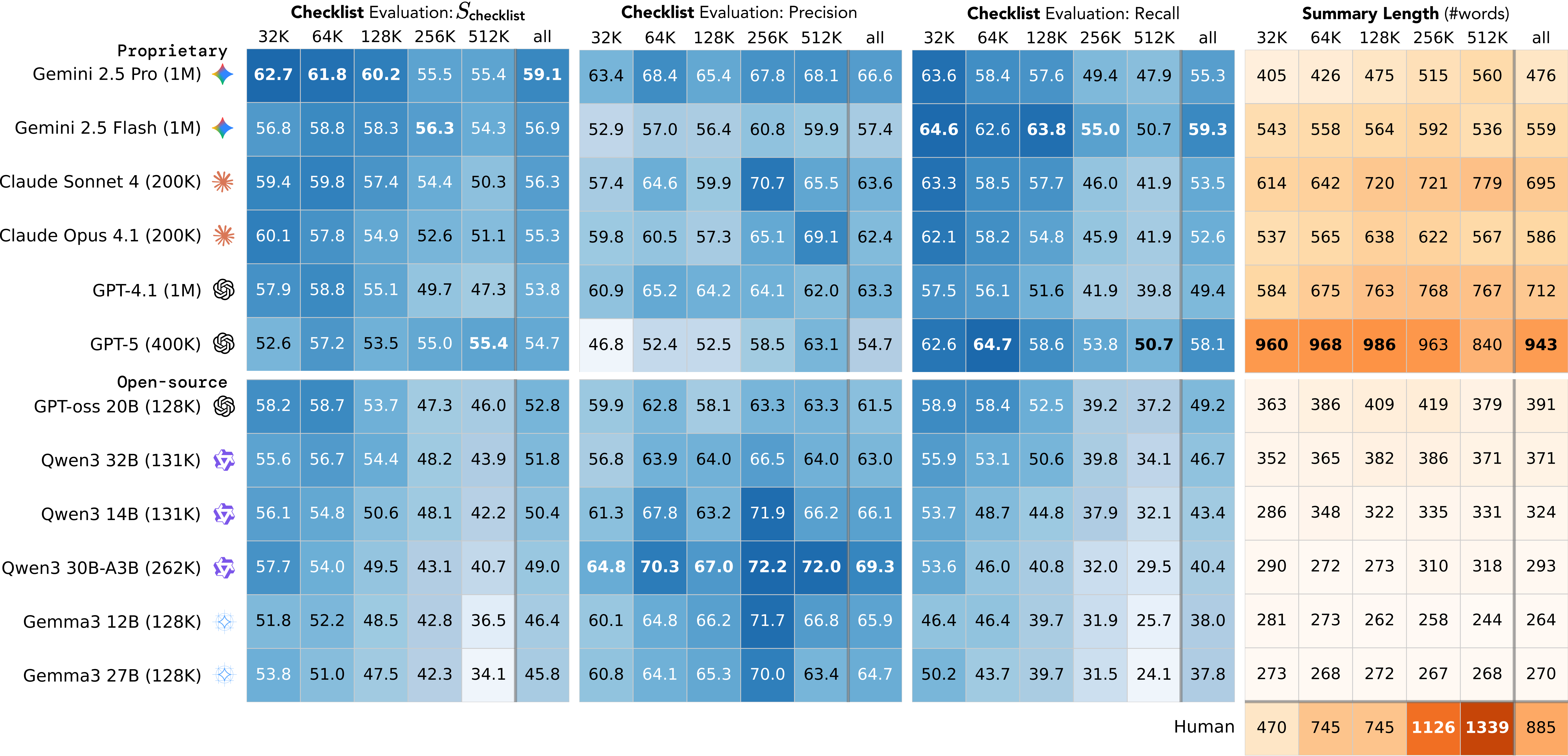}
  \vspace{-5pt}
  \caption{Checklist evaluation for the 12 evaluated LLMs, showing F$_1$ ($S\text{checklist}$), precision, recall, and summary length by case-length bin. Models generally achieve higher precision than recall, suggesting that under-specification, rather than hallucination, is the main failure mode. In the 256K and 512K bins, LLM-generated summaries are much shorter than human summaries and omit substantially more information with much lower recall.}
  \label{fig:precision_recall_length_heatmap}
  \vspace{-12pt}
\end{figure*}

\paragraph{Inter-Annotator Agreement.}
For checklist extraction, the \rev{ten} longest summaries receive triple annotations. Agreement is measured as the average pairwise $S_\text{checklist}$ score across annotators, reaching \rev{83.6} (using Gemma3 27B as the comparison model). For checklist comparison, single-value pairs achieve moderate agreement with Fleiss’ $\kappa=0.57$, while multi-value matching yields an average pairwise F1 of 0.82, indicating high consistency. For writing style similarity, Krippendorff’s $\alpha$ \citep{krippendorff2011computing} across the five aspects averages 0.32. \rev{We also measure a ``two-agree'' metric: overall, at least two annotators agree with each other on the rating 94.4\% of the time, and all three annotators choose different ratings only 5.6\% of the time. This indicates that most instances of writing-style rating show clear majority agreement, and full disagreement is rare.}

\begin{figure*}[!t]
  \centering
  \includegraphics[width=0.99\textwidth]{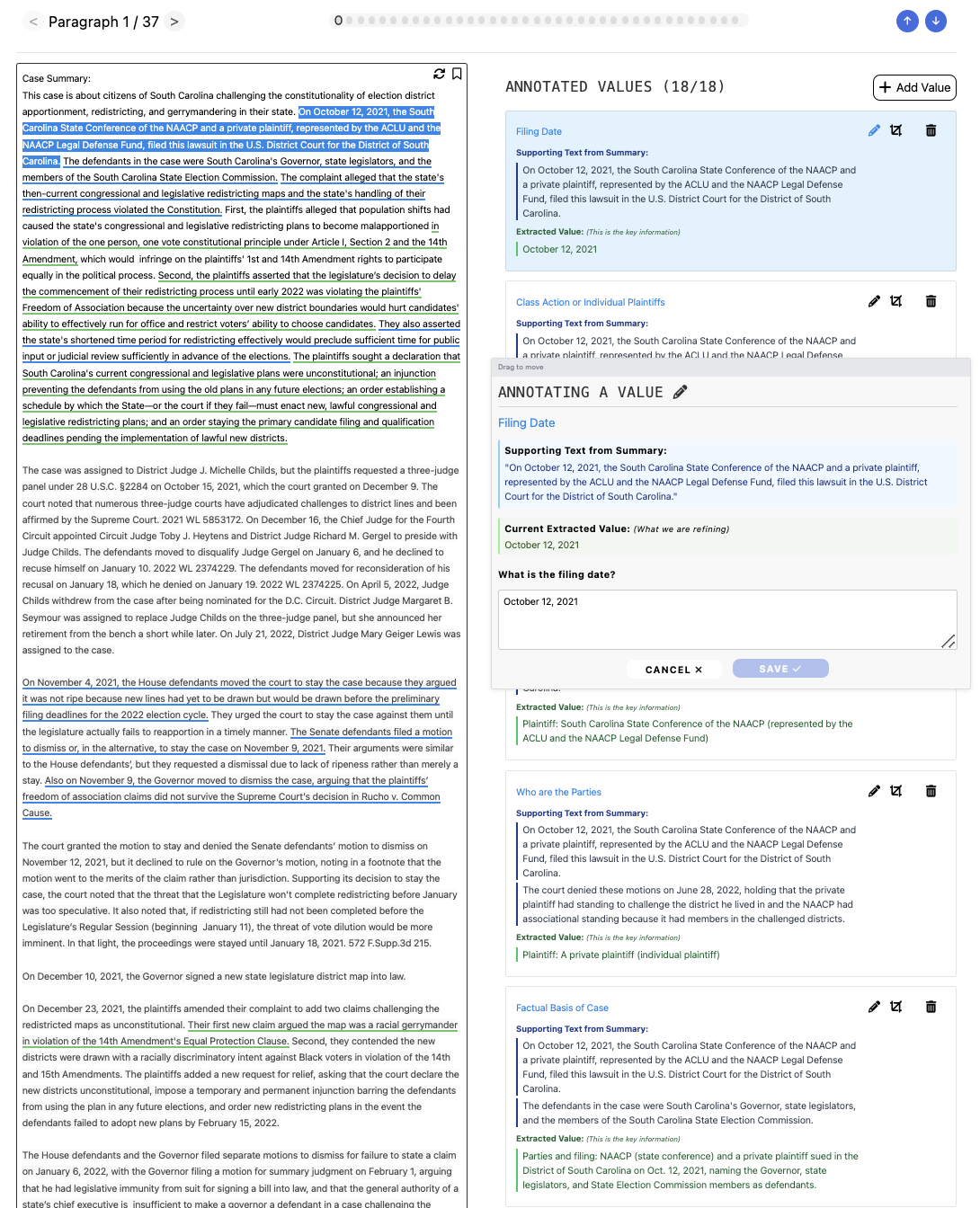}
  \vspace{-5pt}
  \caption{Screenshot of the annotation interface for checklist extraction from summaries. Annotators can add, remove, or modify checklist item values, with the process carried out paragraph by paragraph to ensure each sentence is carefully reviewed.}
  \label{fig:interface-checklist-extraction}
  \vspace{0pt}
\end{figure*}

\begin{figure*}[!t]
  \centering
  \includegraphics[width=0.99\textwidth]{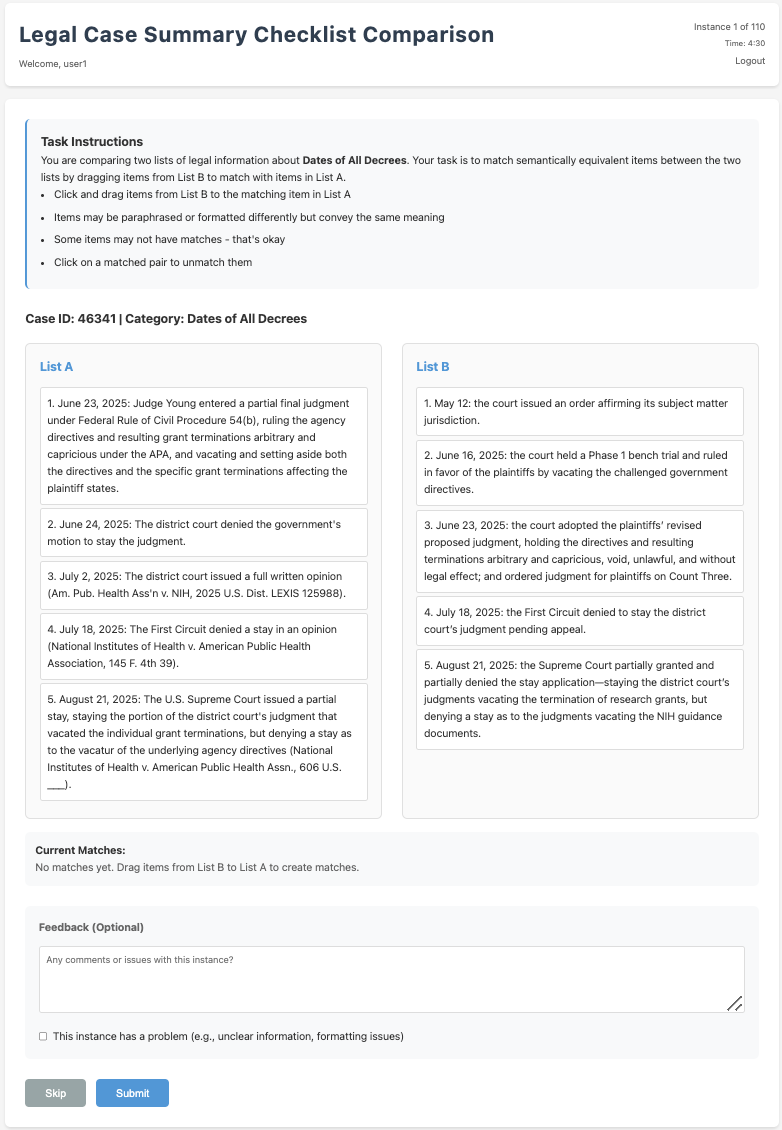}
  \vspace{-5pt}
  \caption{Screenshot of the annotation interface for checklist comparison. Annotators match items between two lists in a list-wise comparison. For string-wise comparison, where both values are strings, the middle component becomes a radio selection with four options: equal, A contains B, B contains A, or different.}
  \label{fig:interface-checklist-comparison}
  \vspace{0pt}
\end{figure*}

\begin{figure*}[!t]
  \centering
  \includegraphics[width=0.99\textwidth]{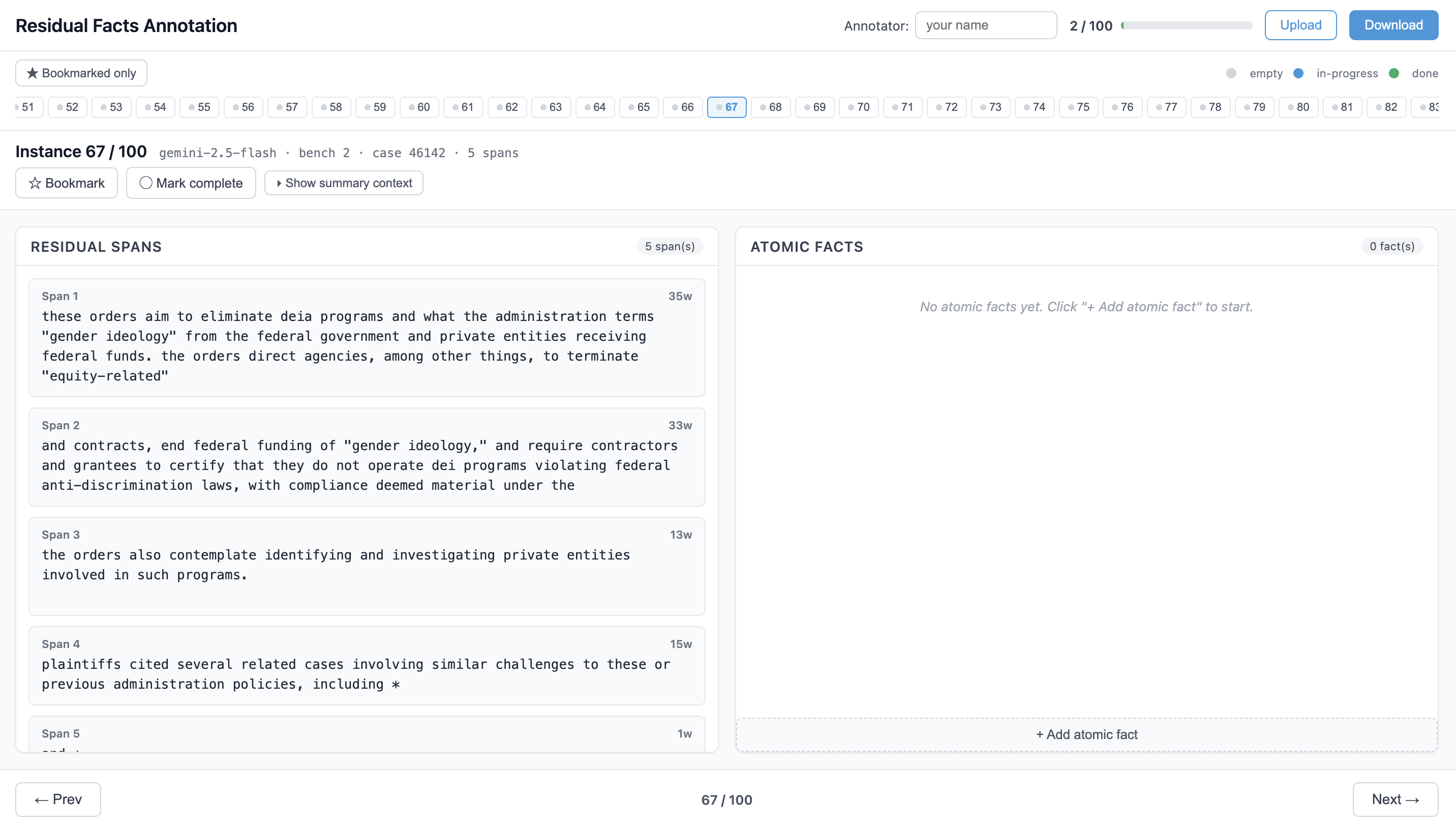}
  \vspace{-5pt}
  \caption{Screenshot of the residual-fact extraction interface, where annotators extract atomic facts from a given list of residual spans.}
  \label{fig:residual-fact-extraction}
  \vspace{0pt}
\end{figure*}

\begin{figure*}[!t]
  \centering
  \includegraphics[width=0.99\textwidth]{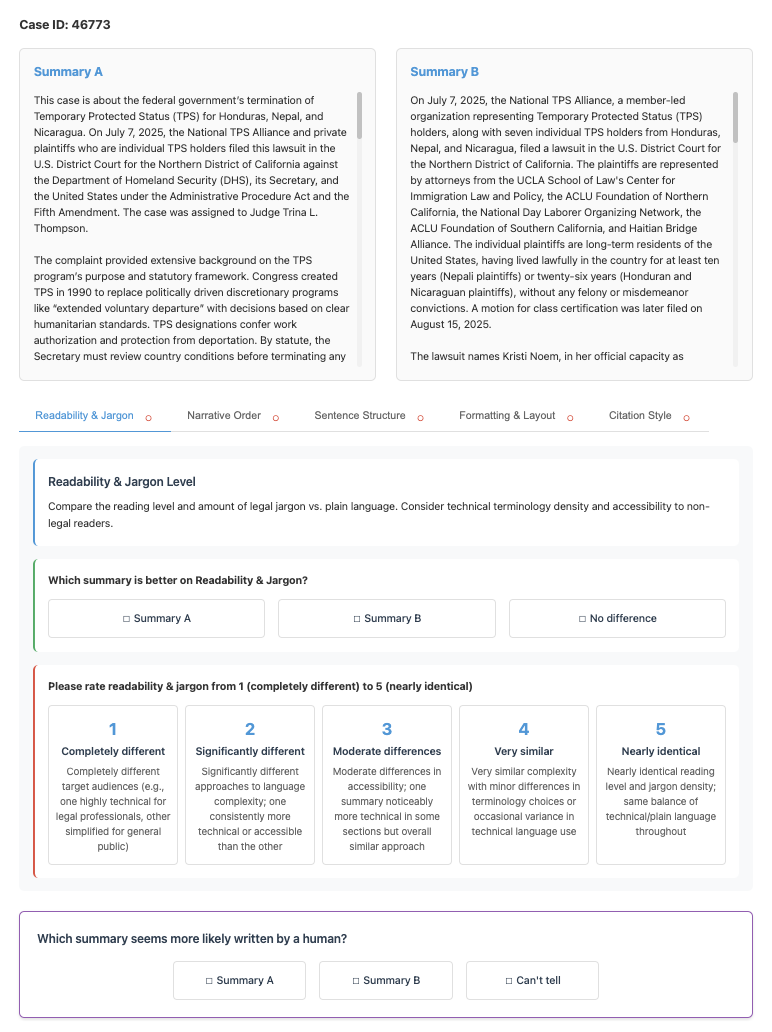}
  \vspace{-5pt}
  \caption{Screenshot of the annotation interface for rating writing style similarity. Annotators compare two summaries, providing ratings on five aspects and answering auxiliary questions such as which summary they prefer.}
  \label{fig:interface-writing-style}
  \vspace{0pt}
\end{figure*}

\paragraph{Annotation Interfaces.}
Figures \ref{fig:interface-checklist-extraction}, \ref{fig:interface-checklist-comparison}, \ref{fig:residual-fact-extraction} and \ref{fig:interface-writing-style} display screenshots of our human annotation interfaces for checklist extraction, checklist comparison, residual-fact extraction, and writing style similarity rating, respectively. The collected data are used for the meta-evaluation of \fwnameRef and for evaluating checklist extraction from case documents methods.
\section{Further Analysis}

Figure \ref{fig:case_type_distribution} shows the distribution of case types across the 100 cases used to evaluate LLM summarization, covering 18 diverse subject areas.

\begin{figure}
  \centering
  \includegraphics[width=\linewidth]{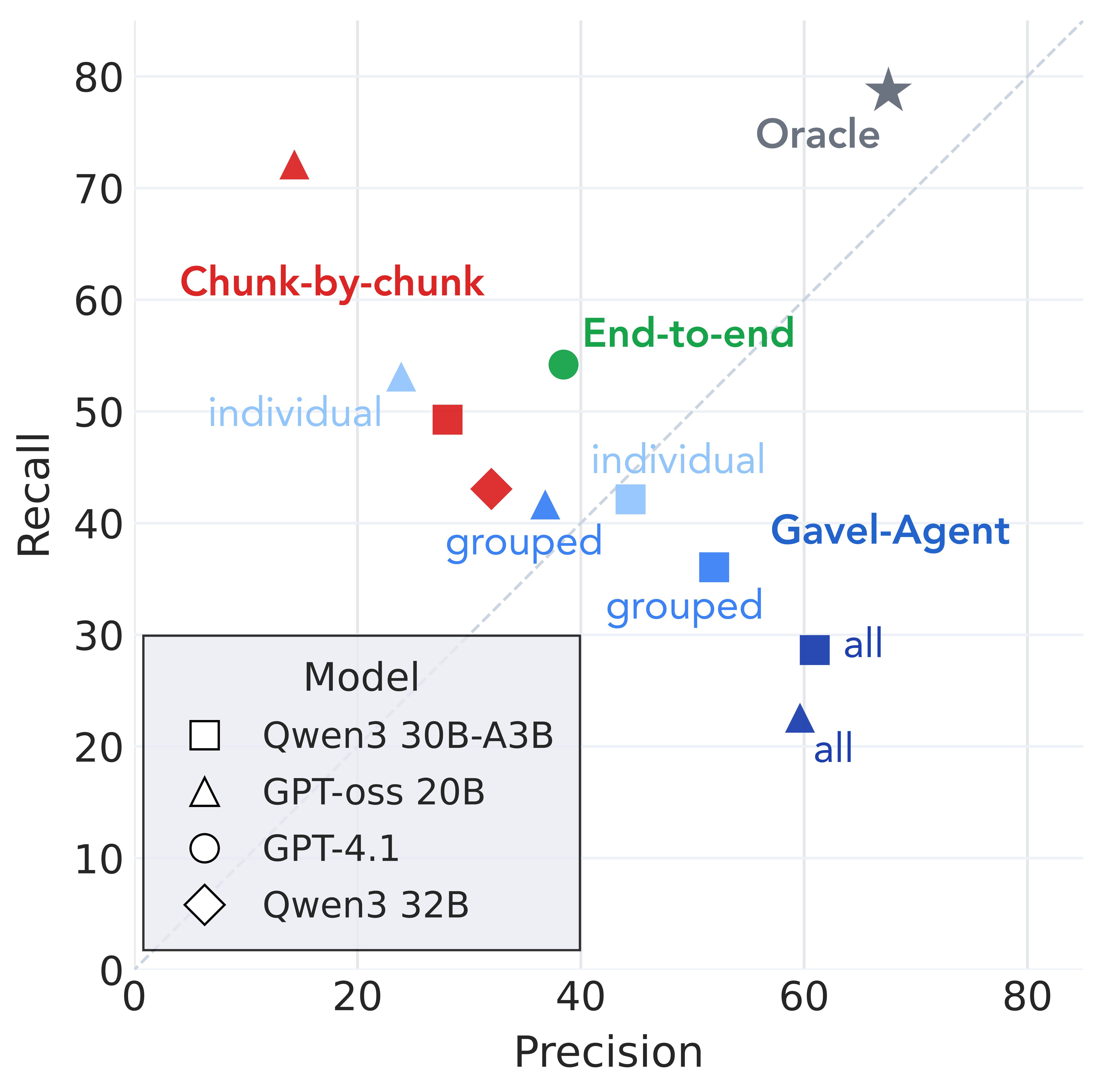}
  \vspace{-15pt}
  \caption{Precision versus recall for different methods extracting from case documents.}
  \label{fig:precision_vs_recall_from_docs}
\end{figure}

\begin{figure}
  \centering
  \includegraphics[width=\linewidth]{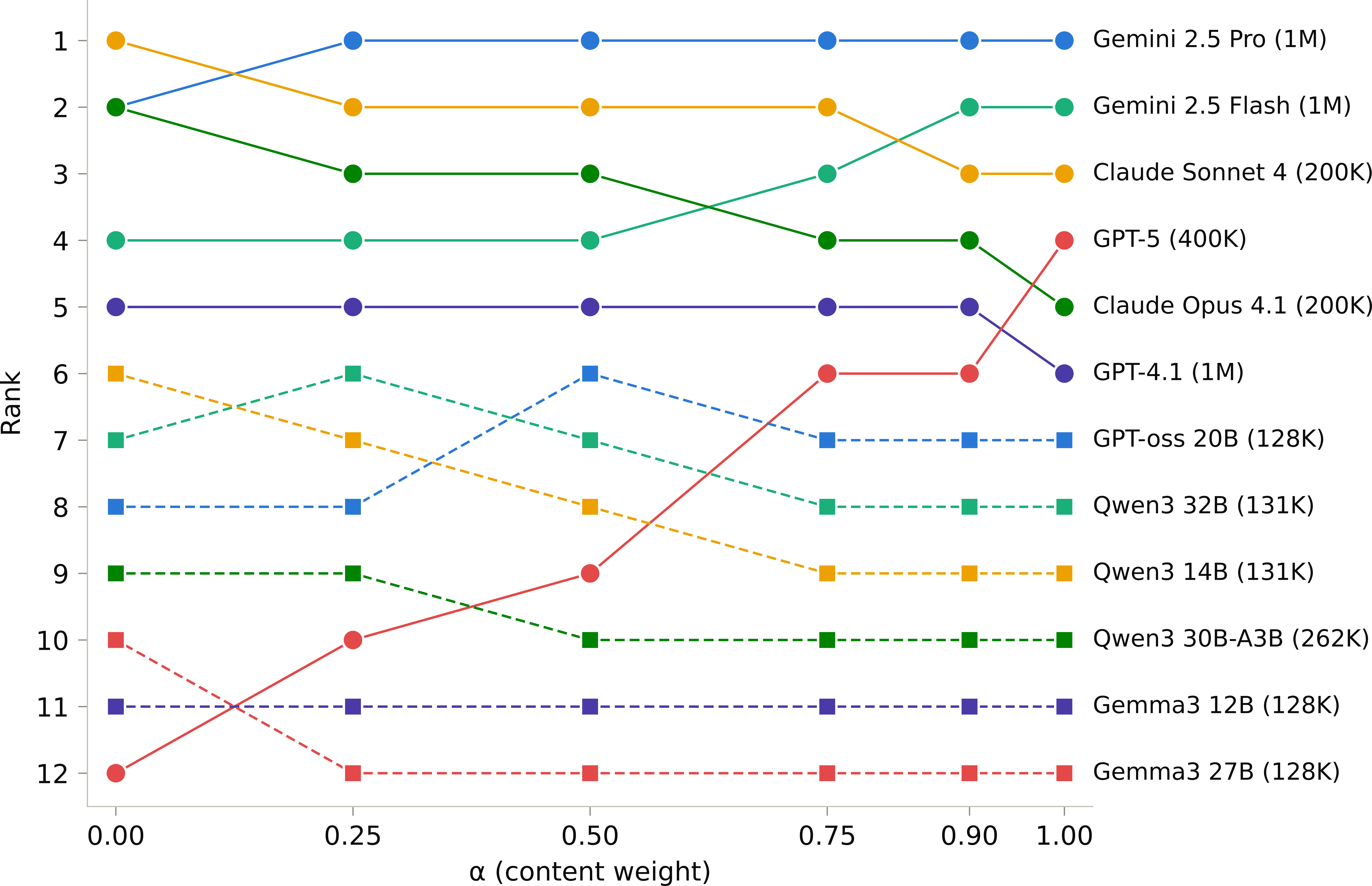}
  \vspace{-15pt}
  \caption{Sensitivity of Gavel-Ref model rankings to the content–style weighting parameter $\alpha$. Scores are recomputed across all 100 cases. Model rankings remain stable across different values of $\alpha$.}
  \label{fig:alpha_sensitivity_rank}
\end{figure}

\begin{figure*}[!t]
  \centering
  \includegraphics[width=0.99\textwidth]{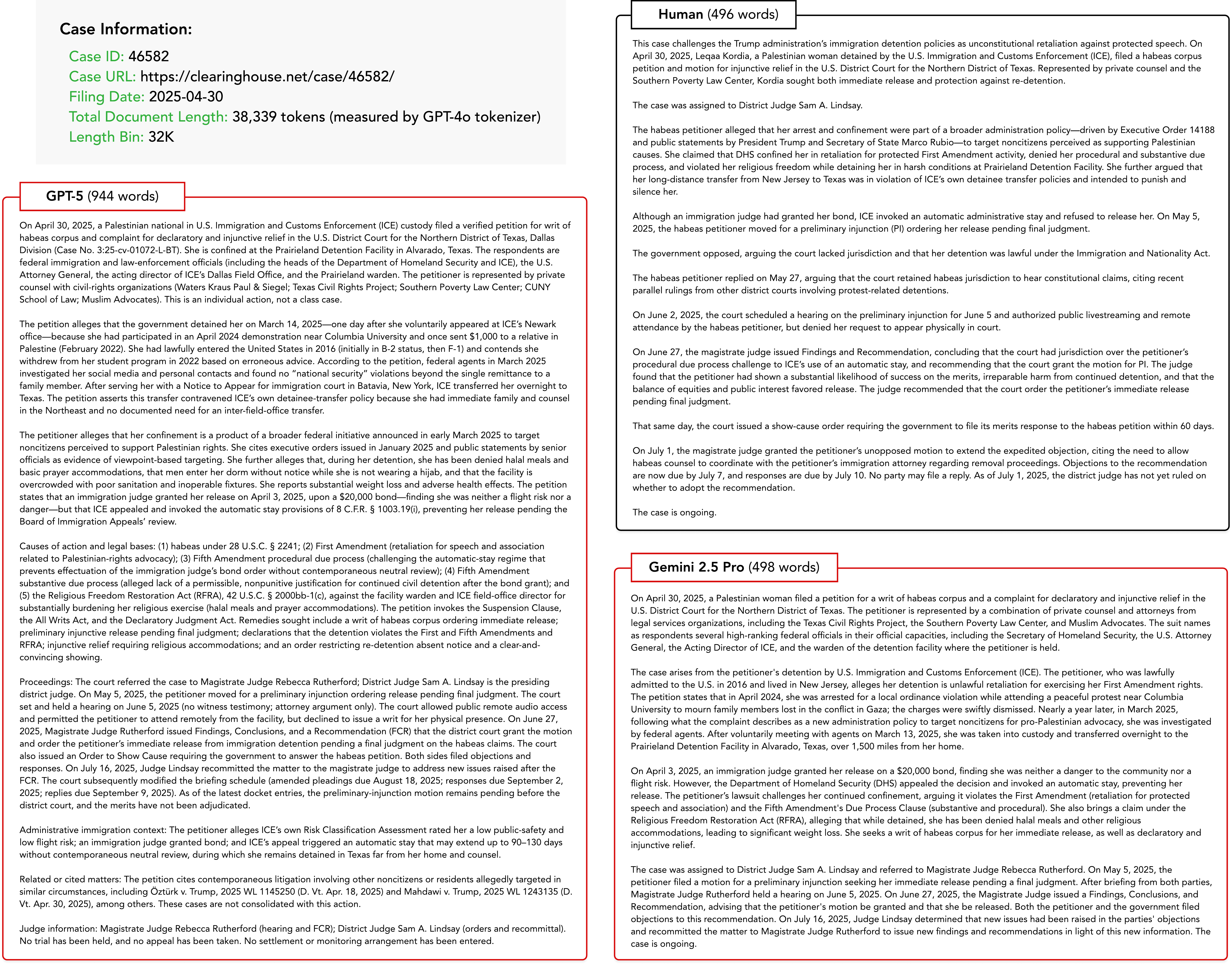}
  \vspace{-5pt}
  \caption{Example summaries from GPT-5, Gemini 2.5 Pro, and a human reference for a case in the 32K bin. This illustrates why GPT-5 produces very long summaries (as seen in Figure \ref{fig:precision_recall_length_heatmap}) even for short cases.}
  \label{fig:summary_example}
  \vspace{-12pt}
\end{figure*}

Figure~\ref{fig:precision_recall_length_heatmap} shows checklist F$_1$ ($S_\text{checklist}$), precision, recall, and average summary length for each LLM across case-length bins. Overall, models are more likely to omit key information than to hallucinate, as precision is consistently higher than recall. This issue becomes more severe as case length increases: recall drops substantially, while precision often remains stable or even increases.
Compared to human summaries, LLMs only approach human length in the 32K–128K bins; for 256K and 512K cases, all models produce much shorter summaries than humans. In general, open-source models generate noticeably shorter summaries than proprietary models.
Among all models, GPT5 is an outlier: it consistently produces very long summaries (often over 900 words) even for short cases (32K–128K), substantially longer than the human references. Figure~\ref{fig:summary_example} shows a typical example. GPT-5 often writes in a highly verbose, list-style format rather than a narrative, which we hypothesize is related to its “high” thinking mode.
We also compute instance-level correlations between summary length and $S_\text{\fwnameRef}$. Overall, we observe a moderate positive correlation (Pearson $r=0.31$, Spearman $\rho=0.36$, Kendall’s $\tau=0.24$), but this is largely driven by weaker open-source models that both underperform and produce shorter summaries.
When we separate proprietary and open-source models, the correlations become much smaller: within proprietary models, Pearson $r=-0.11$, Spearman $\rho=-0.13$, and Kendall’s $\tau=0.09$; within open-source models, Pearson $r=0.20$, Spearman $\rho=0.20$, and Kendall’s $\tau=0.14$. This suggests that, once we control for model family, summary length alone explains only a small fraction of the performance differences.

\begin{figure}
  \centering
  \includegraphics[width=\linewidth]{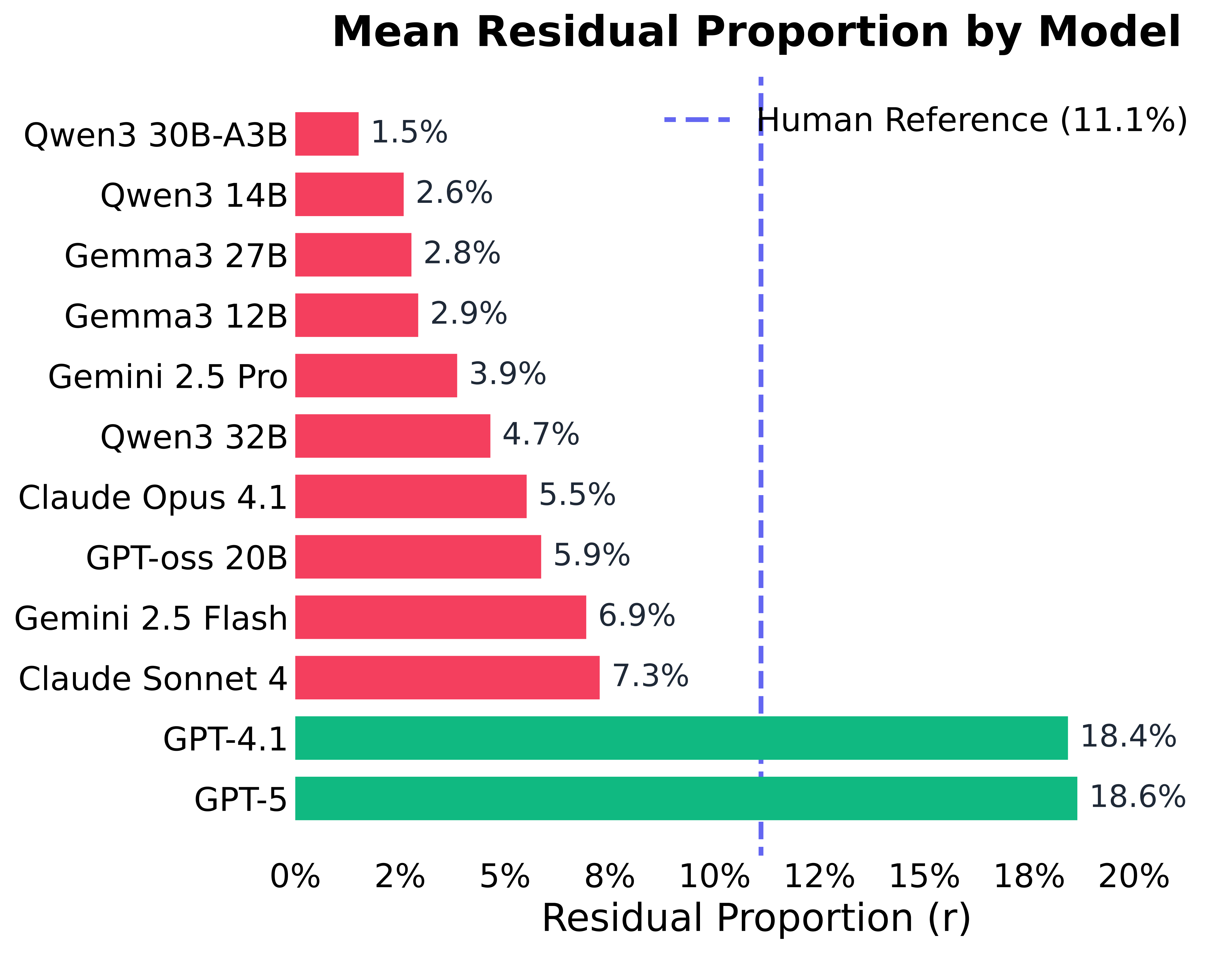}
  \vspace{-15pt}
  \caption{Average residual ratio $r$ across model summaries, as in Equation \ref{eq:gavel-ref-score}, which is the proportion of residual content in the summary. GPT-4.1 and GPT-5 contain more residual information than human summaries, while other models contain less.}
  \label{fig:model_residual_proportion}
\end{figure}

Figure~\ref{fig:alpha_sensitivity_rank} presents a sensitivity analysis over $\alpha \in {0, 0.25, 0.5, 0.75, 0.9, 1.0}$. For each value of $\alpha$, we recompute the overall Gavel-Ref score across all 100 cases for the all 12 models: $S_{Gavel-Ref}=(1-r)\alpha S_{checklist}+r\alpha S_{residual}+(1-\alpha)S_{style}.$
Here, $\alpha$ controls the relative weight assigned to content coverage versus writing style.
The rankings are stable across weighting choices. Gemini 2.5 Pro ranks first for every $\alpha \geq 0.25$; only the style-only setting ($\alpha=0$) changes the top-ranked model. For all $\alpha \geq 0.25$, the system-level Kendall's $\tau$, measured relative to the $\alpha=0.9$ ranking, is at least 0.733 among the proprietary models and 0.758 across all 12 models. The remaining rank changes occur mainly among nearly tied models and GPT-5, whose writing-style score is substantially lower because it generates overly verbose summaries for cases in the 32K–128K length bins.

\begin{figure}
  \centering
  \includegraphics[width=\linewidth]{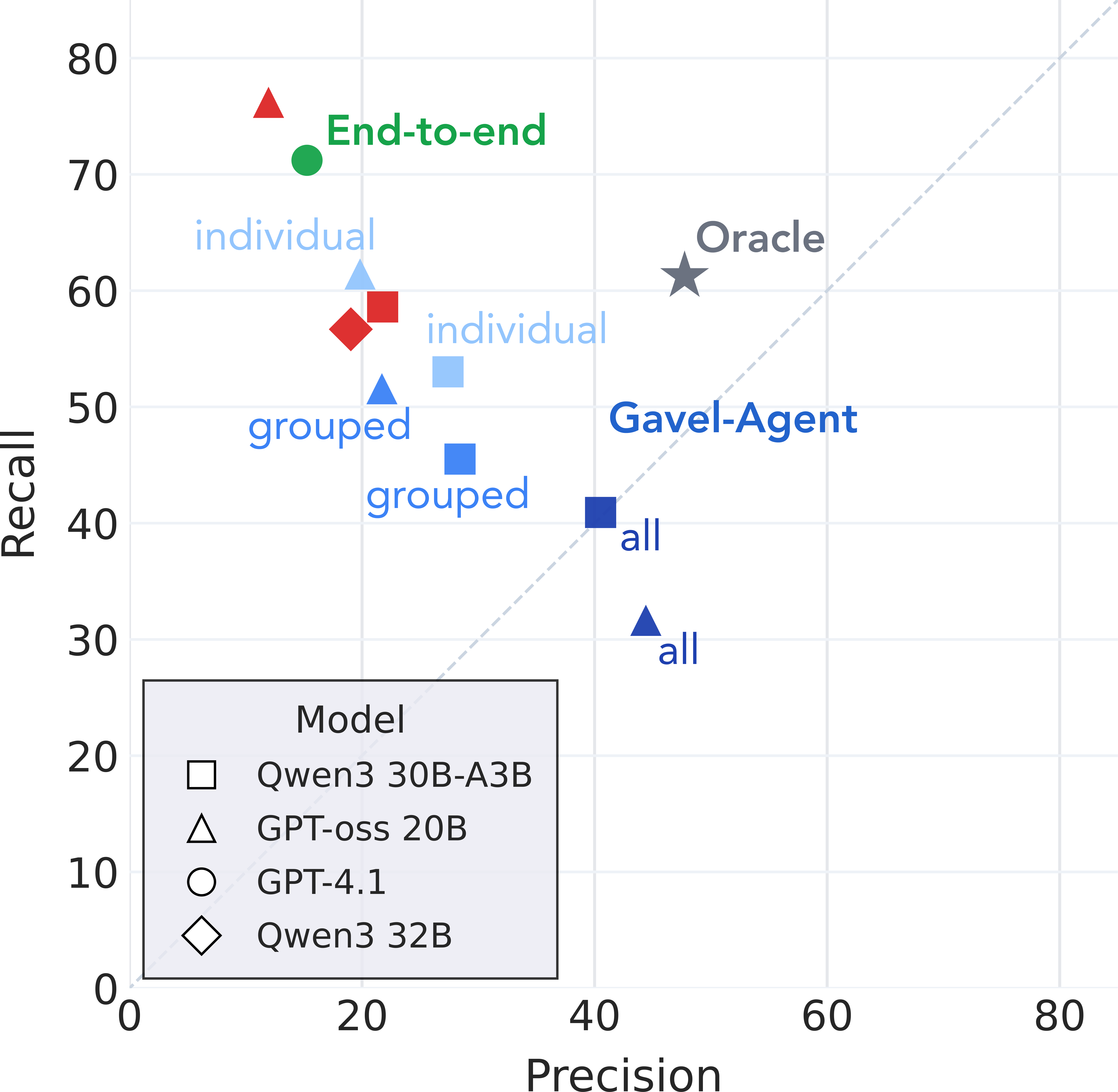}
  \vspace{-15pt}
  \caption{Precision versus recall for different methods extracting from medical review.}
  \label{fig:precision_vs_recall_from_docs_medical}
\end{figure}

Figure \ref{fig:model_residual_proportion} shows the average residual ratio $r$ across model summaries. GPT-4.1 and GPT-5 have the highest residual ratios, followed by human summaries and then the remaining models.

\begin{figure*}[!t]
  \centering
  \includegraphics[width=0.99\textwidth]{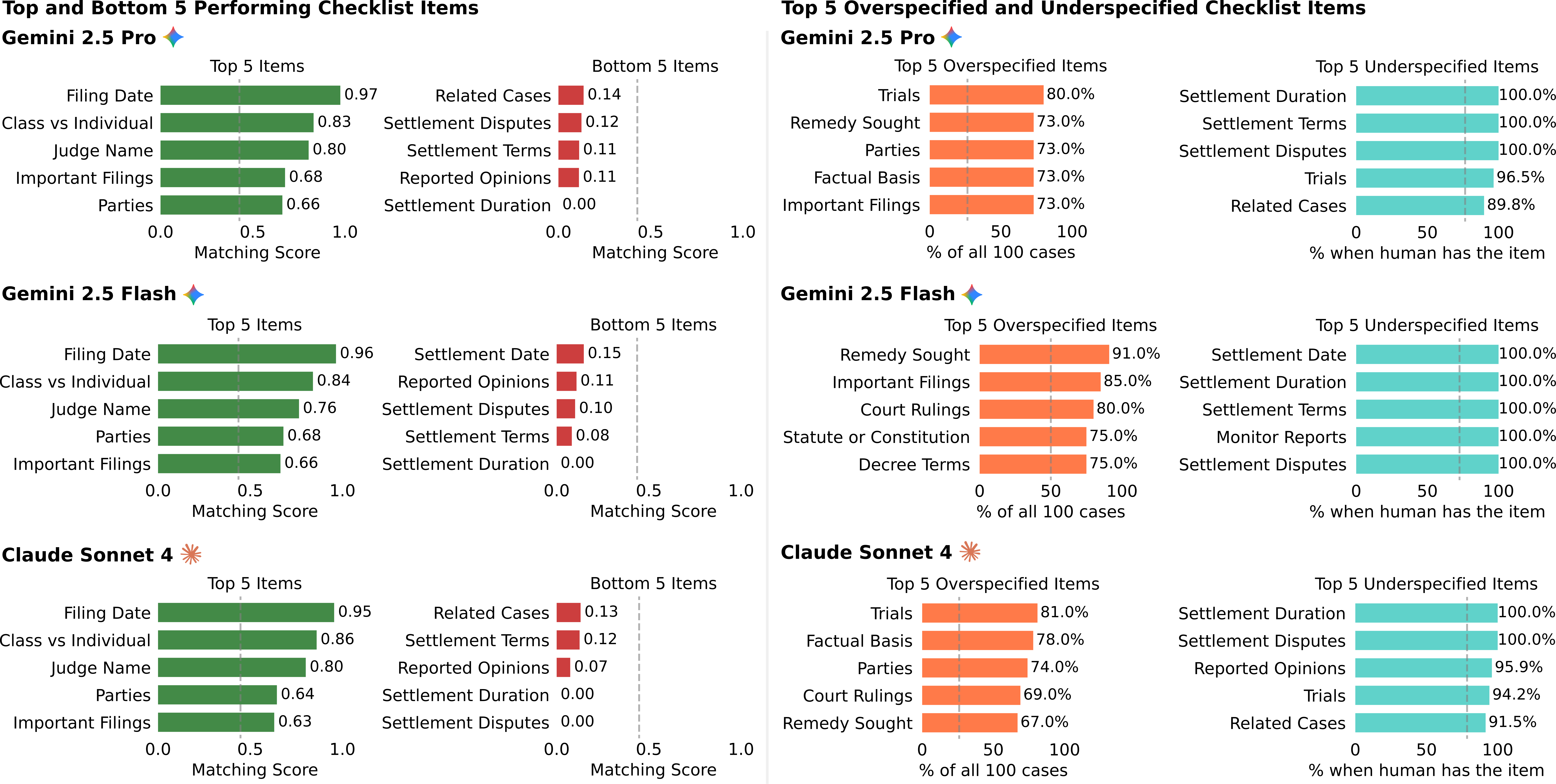}
  \vspace{-5pt}
  \caption{Performance breakdown for the top-3 models in checklist evaluation (Gemini 2.5 Pro, Gemini 2.5 Flash, and Claude Sonnet 4): top/bottom 5 checklist items by matching score and most frequently over/under-specified items. Overspecification measured as frequency across all 100 cases; underspecification as frequency among cases where human summary includes that item.}
  \label{fig:llm_eval_by_items_full}
  \vspace{0pt}
\end{figure*}

Figure \ref{fig:llm_eval_by_items_full} presents the item-level performance for the top 3 models in checklist evaluation---Gemini 2.5 Flash, Pro and Claude Sonnet 4---showing their top and bottom 5 checklist items plus consistently over- and under-specified items.
All three models exhibit high similar performance patterns across items.

\begin{figure*}[t]
  \centering
  \begin{subfigure}[b]{0.24\textwidth}
    \centering
    \includegraphics[width=\linewidth]{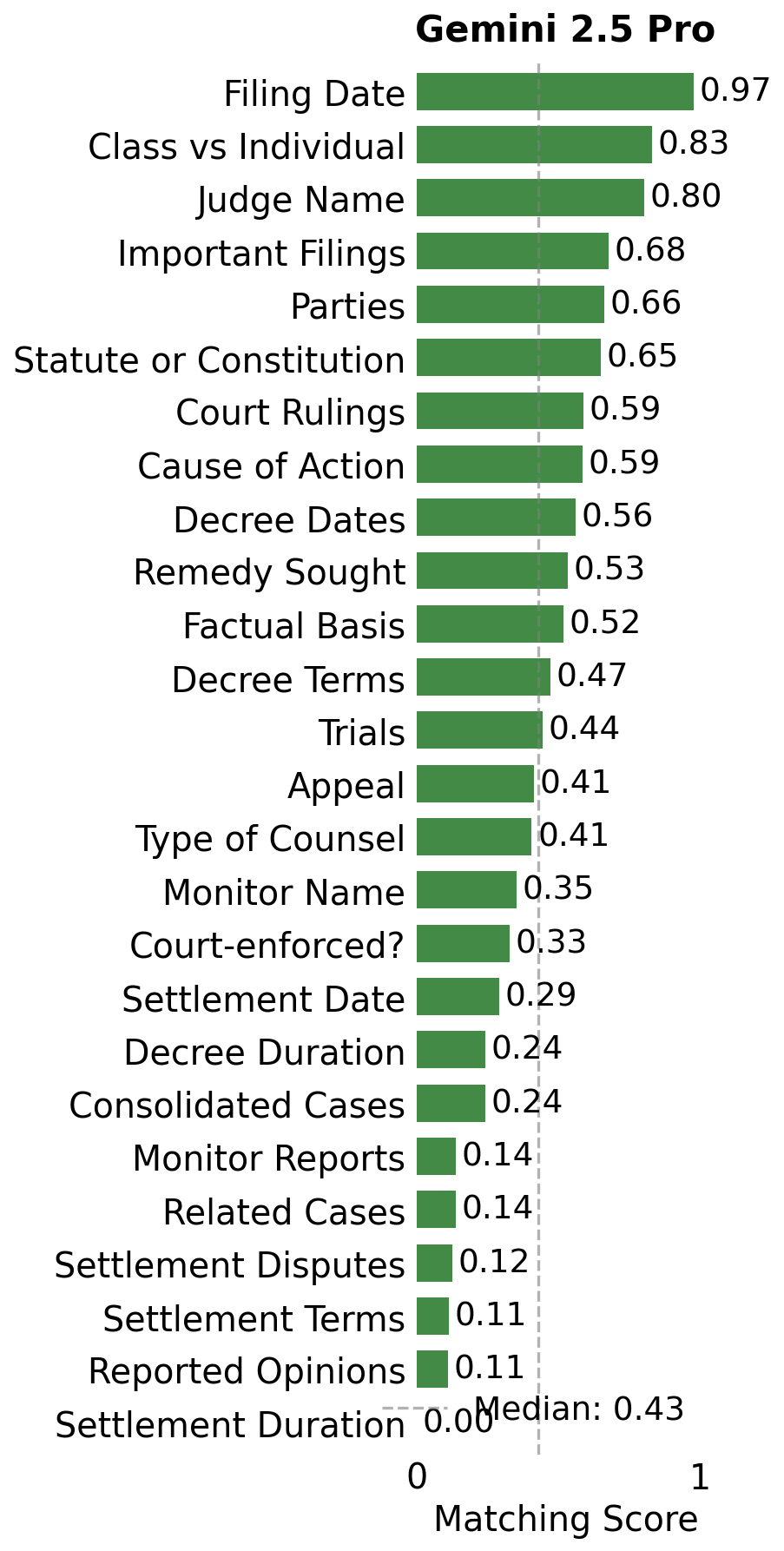}
  \end{subfigure}
  \hfill
  \begin{subfigure}[b]{0.24\textwidth}
    \centering
    \includegraphics[width=\linewidth]{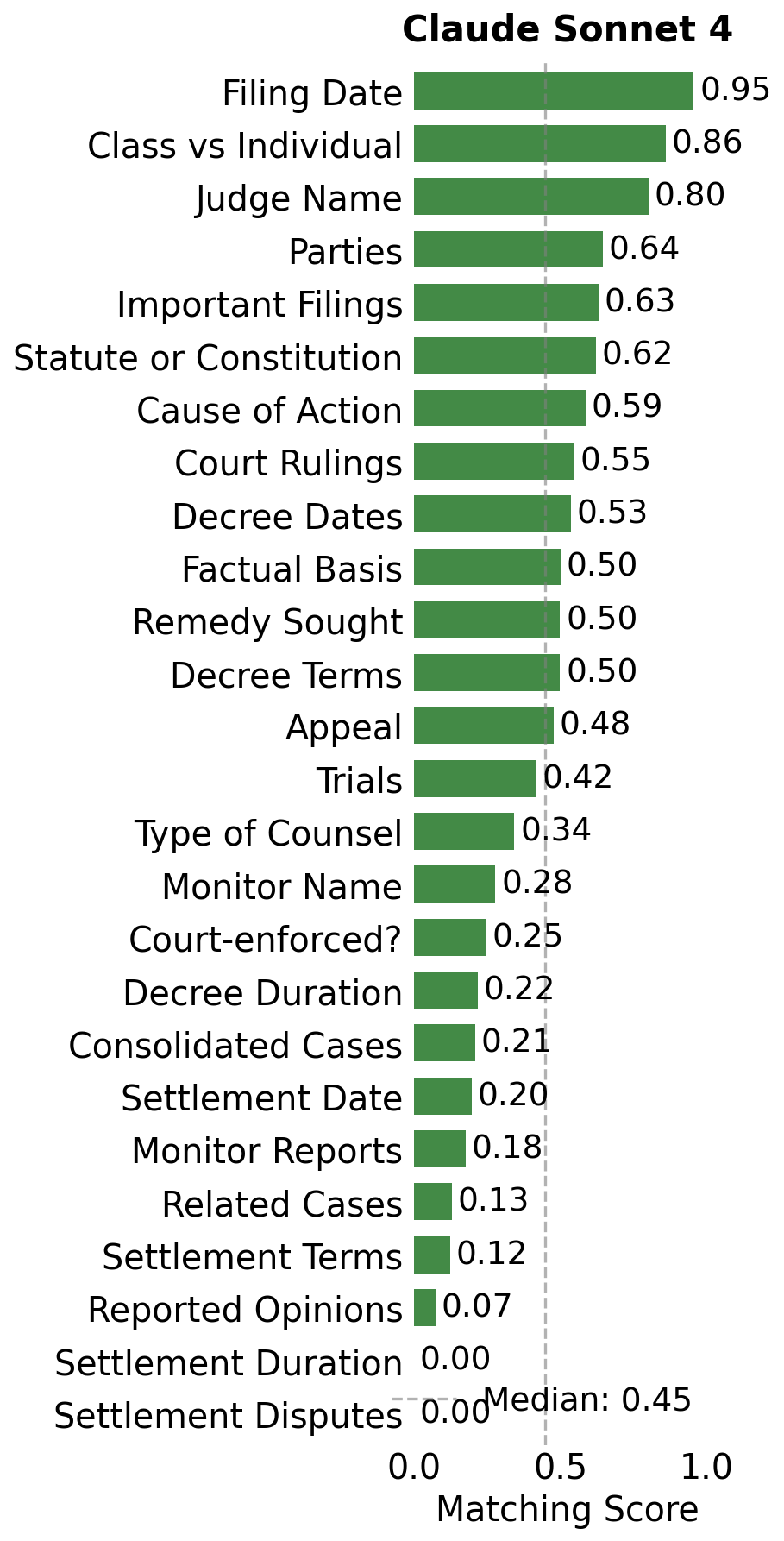}
  \end{subfigure}
  \hfill
  \begin{subfigure}[b]{0.24\textwidth}
    \centering
    \includegraphics[width=\linewidth]{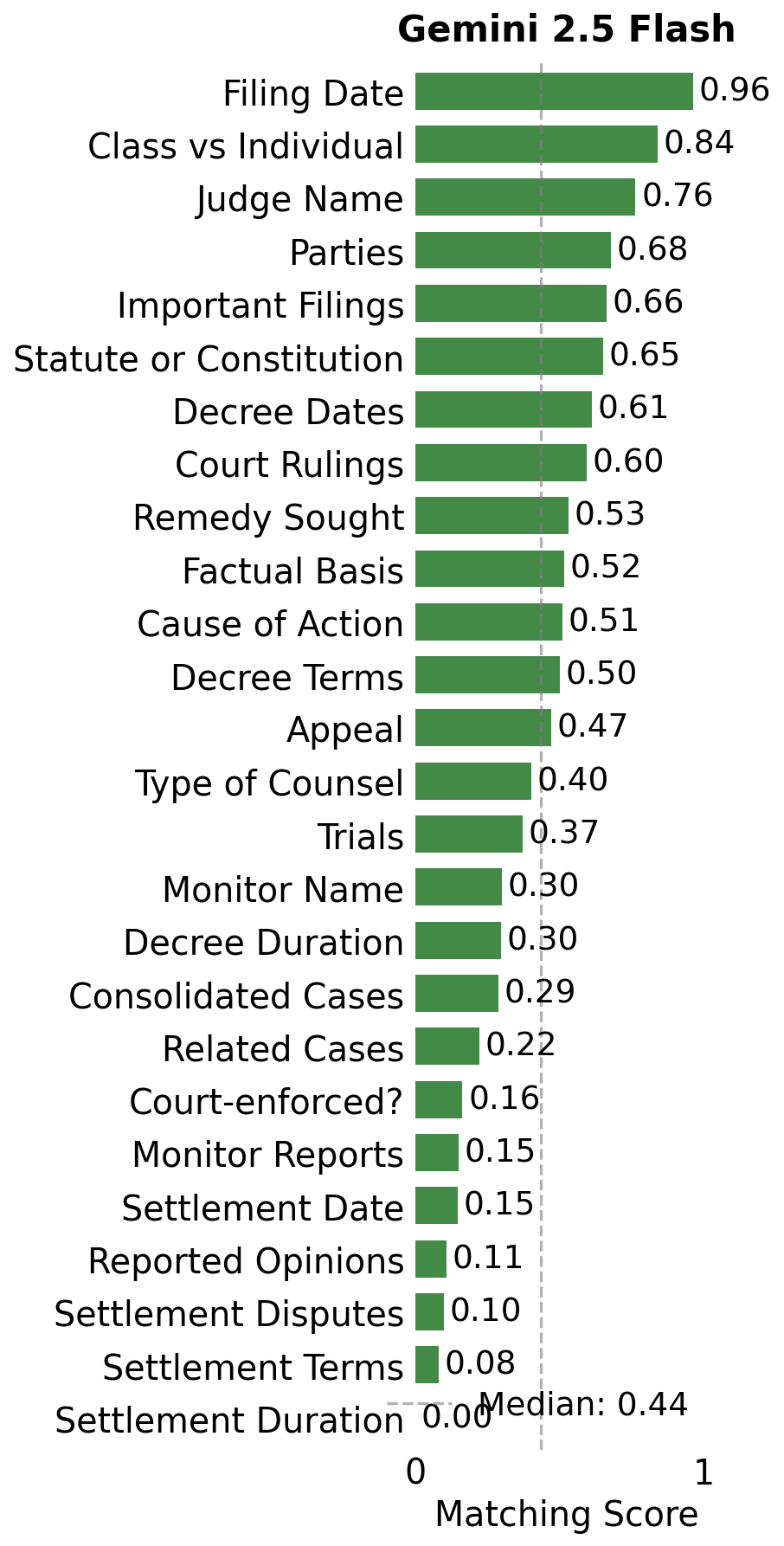}
  \end{subfigure}
  \hfill
  \begin{subfigure}[b]{0.24\textwidth}
    \centering
    \includegraphics[width=\linewidth]{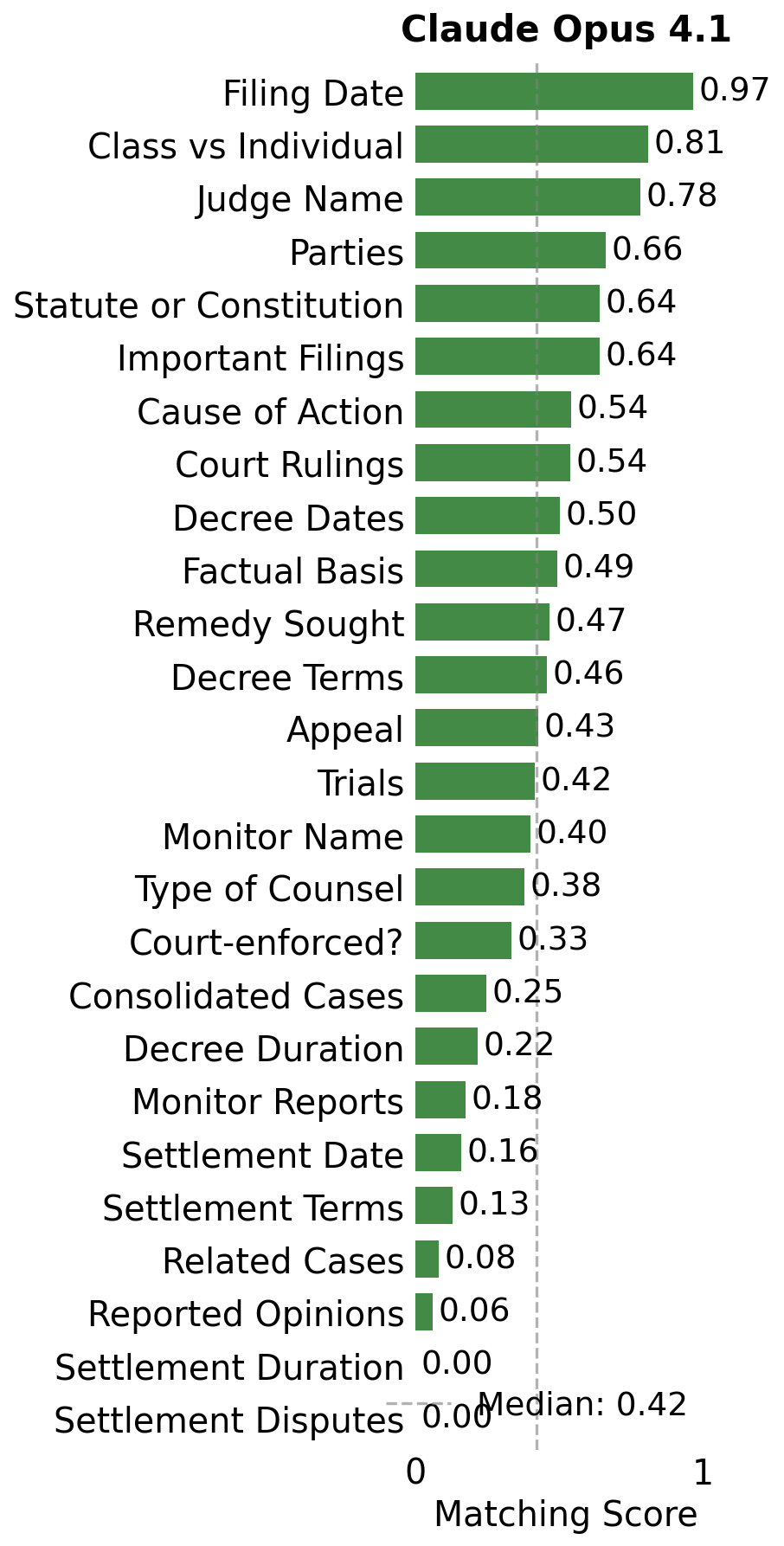}
  \end{subfigure}
  \vspace{-4pt}
  \caption{Checklist item-level performance for each LLM in the checklist evaluation. The metric is the matching score $m_i$. This figure shows results for Gemini 2.5 Pro, Claude Sonnet 4, Gemini 2.5 Flash, and Claude Opus 4.1.}
  \label{fig:model_checklist_item_level_1}
\end{figure*}

\begin{figure*}[t]
  \centering
  \begin{subfigure}[b]{0.24\textwidth}
    \centering
    \includegraphics[width=\linewidth]{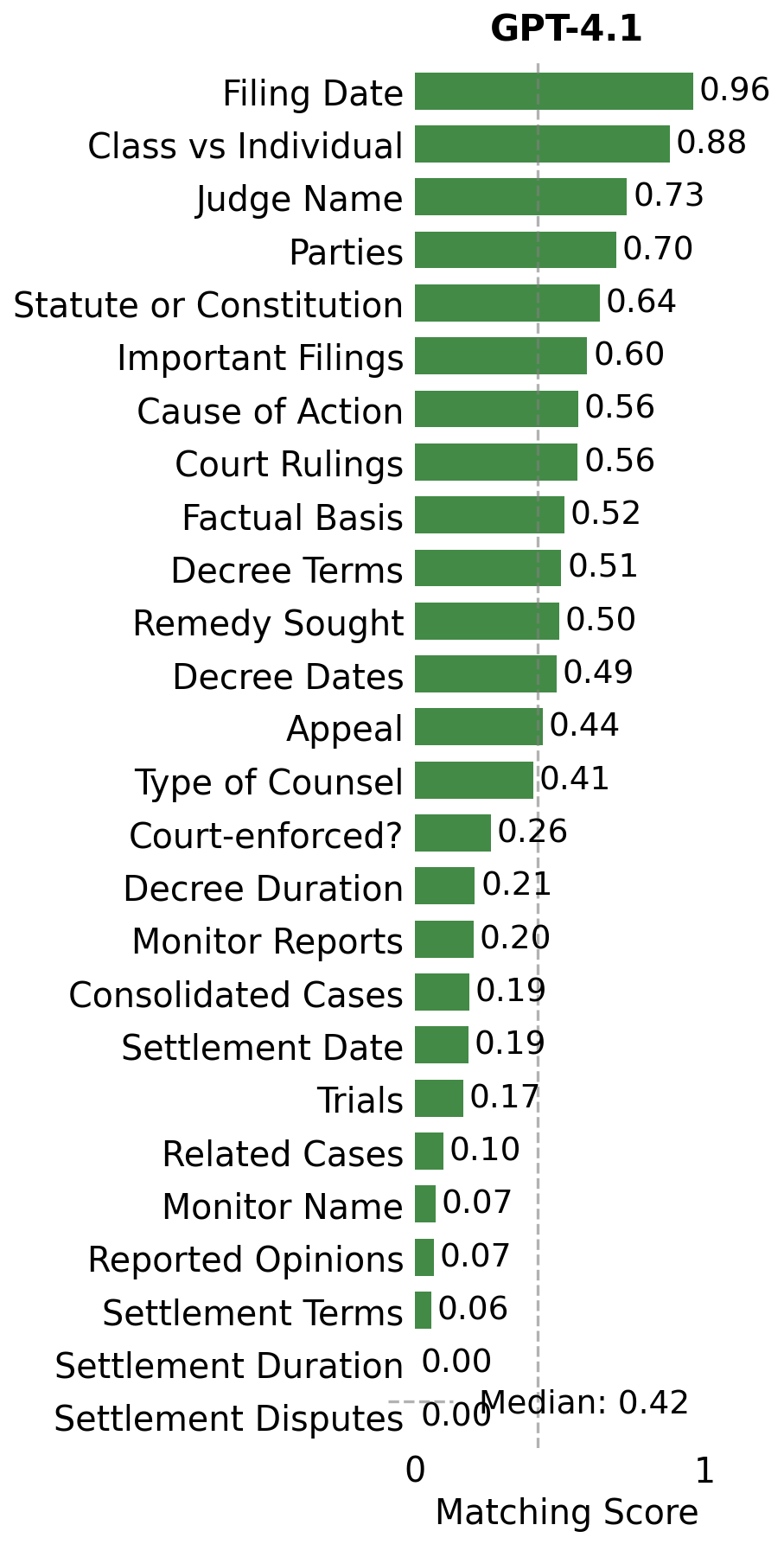}
  \end{subfigure}
  \hfill
  \begin{subfigure}[b]{0.24\textwidth}
    \centering
    \includegraphics[width=\linewidth]{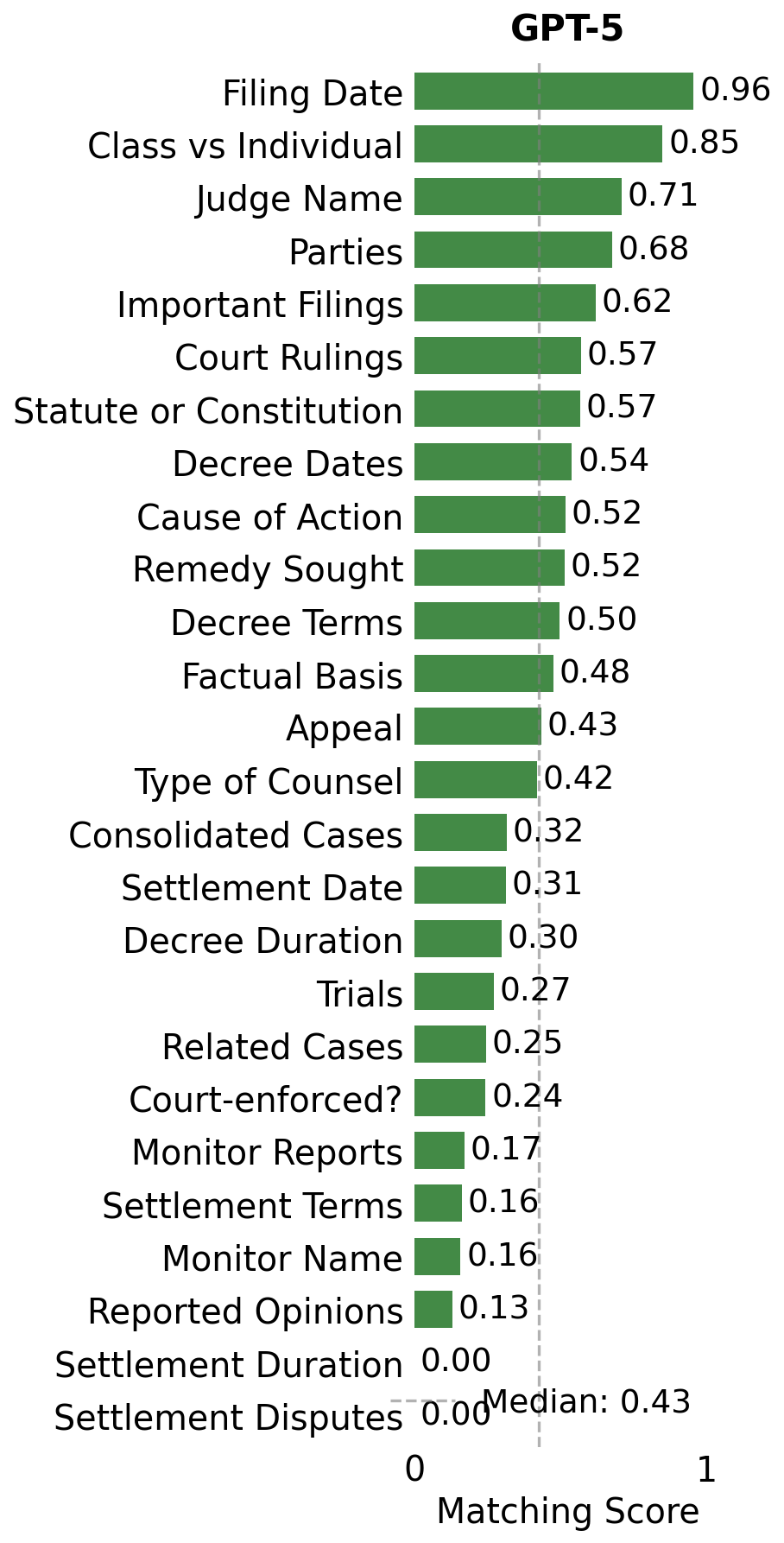}
  \end{subfigure}
  \hfill
  \begin{subfigure}[b]{0.24\textwidth}
    \centering
    \includegraphics[width=\linewidth]{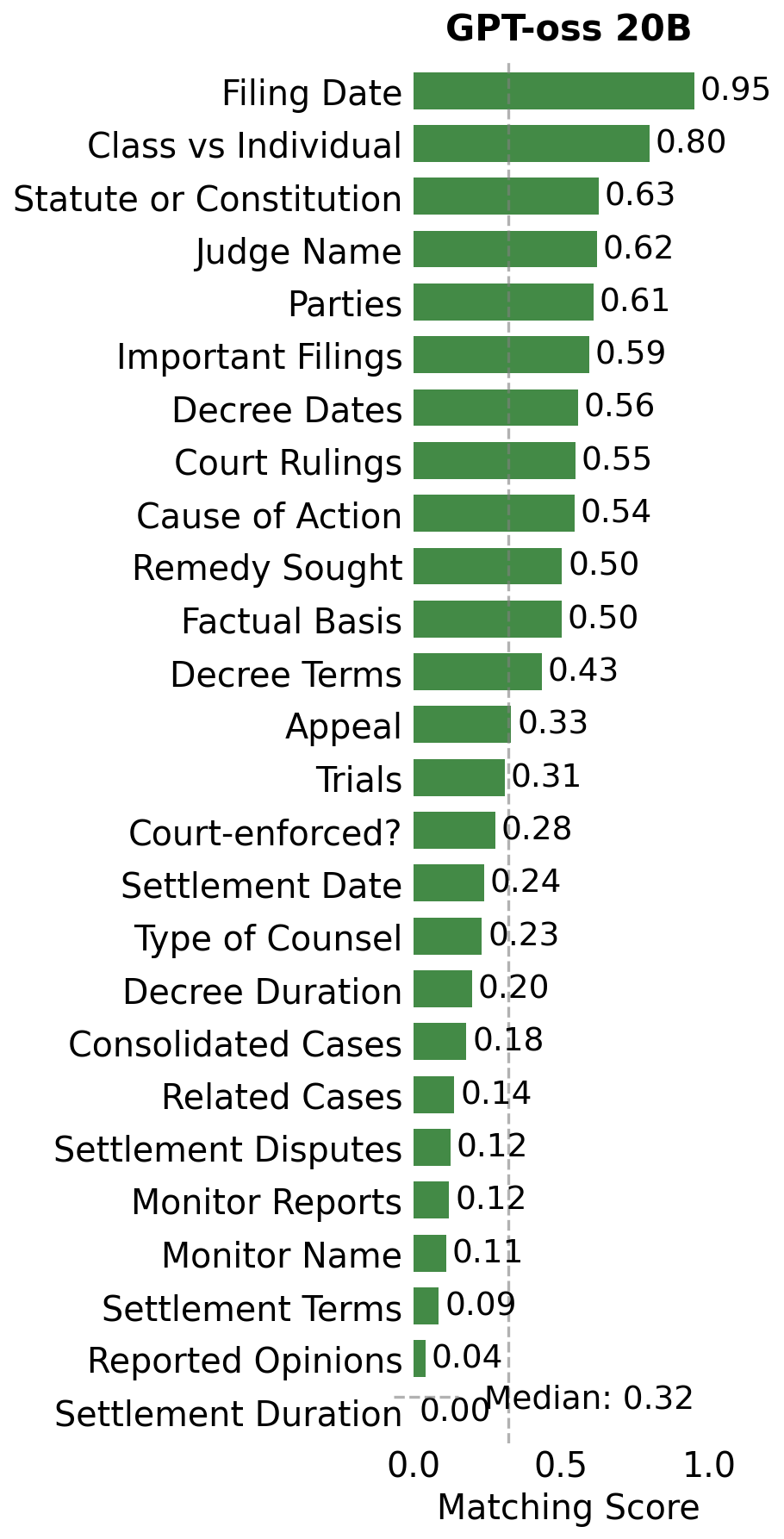}
  \end{subfigure}
  \hfill
  \begin{subfigure}[b]{0.24\textwidth}
    \centering
    \includegraphics[width=\linewidth]{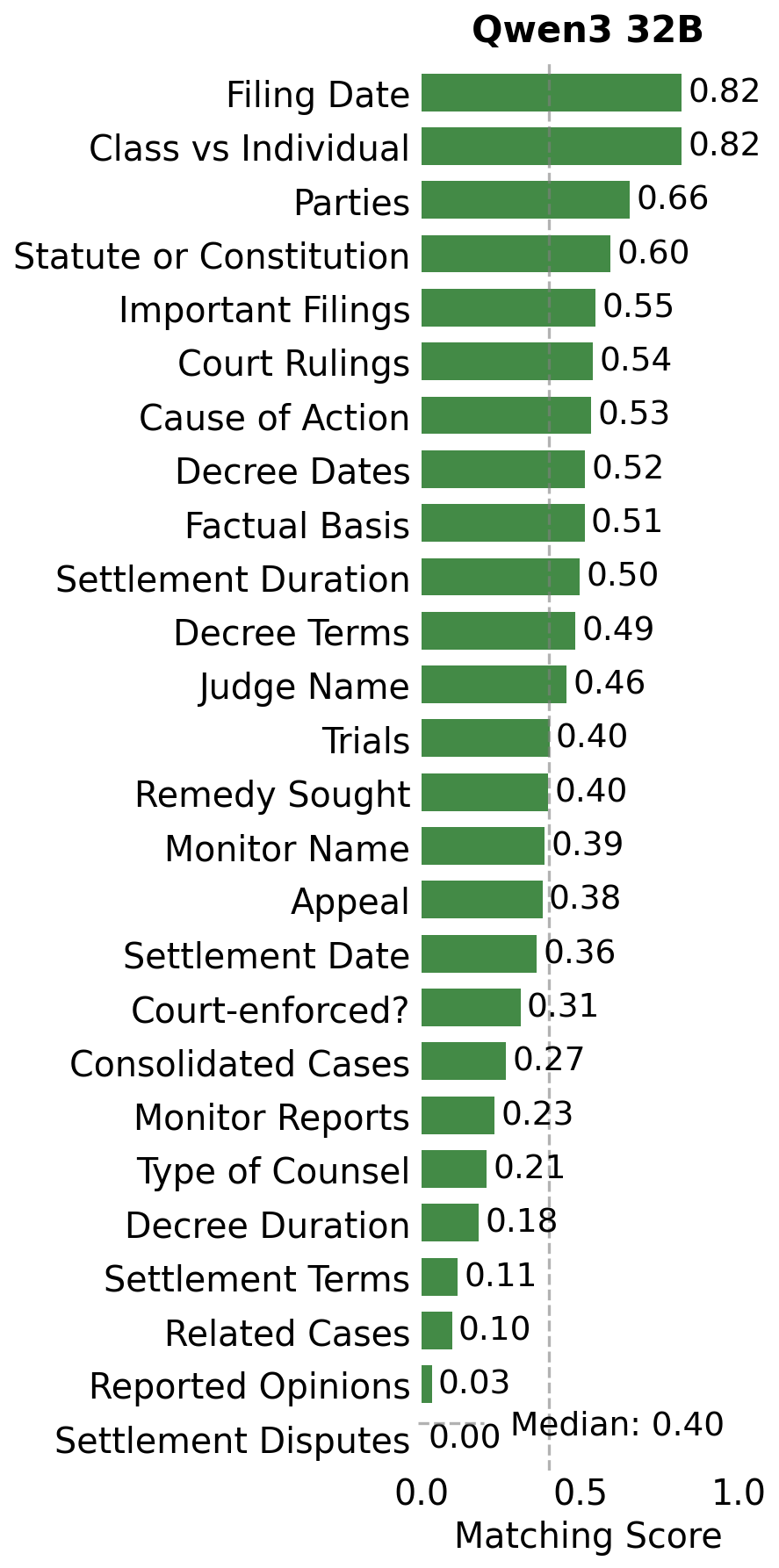}
  \end{subfigure}
  \vspace{-4pt}
  \caption{Checklist item-level performance for each LLM in the checklist evaluation. The metric is the matching score $m_i$. This figure shows results for GPT-4.1, GPT-5, GPT-oss 20B, Qwen3 32B.}
  \label{fig:model_checklist_item_level_2}
\end{figure*}

\begin{figure*}[t]
  \centering
  \begin{subfigure}[b]{0.24\textwidth}
    \centering
    \includegraphics[width=\linewidth]{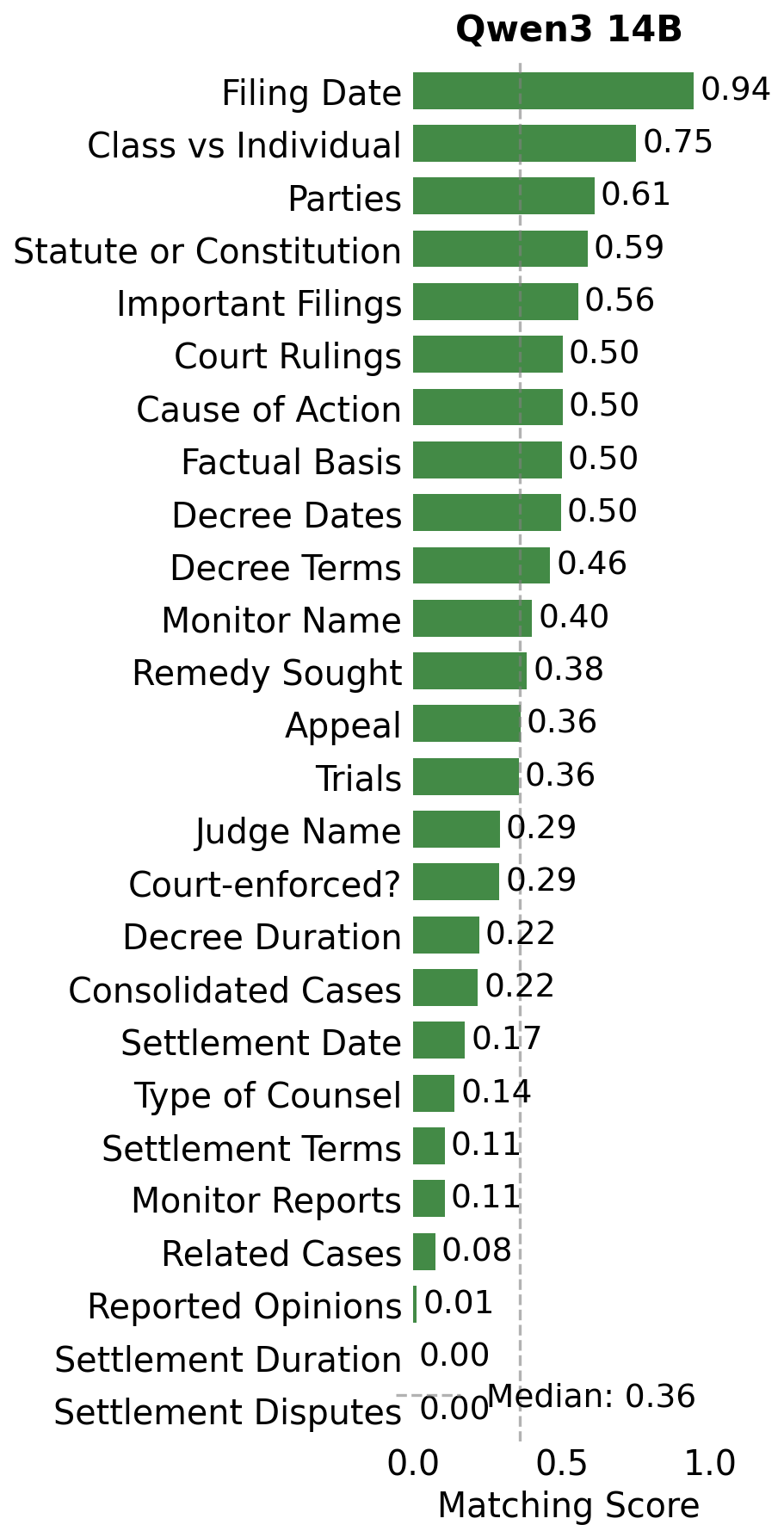}
  \end{subfigure}
  \hfill
  \begin{subfigure}[b]{0.24\textwidth}
    \centering
    \includegraphics[width=\linewidth]{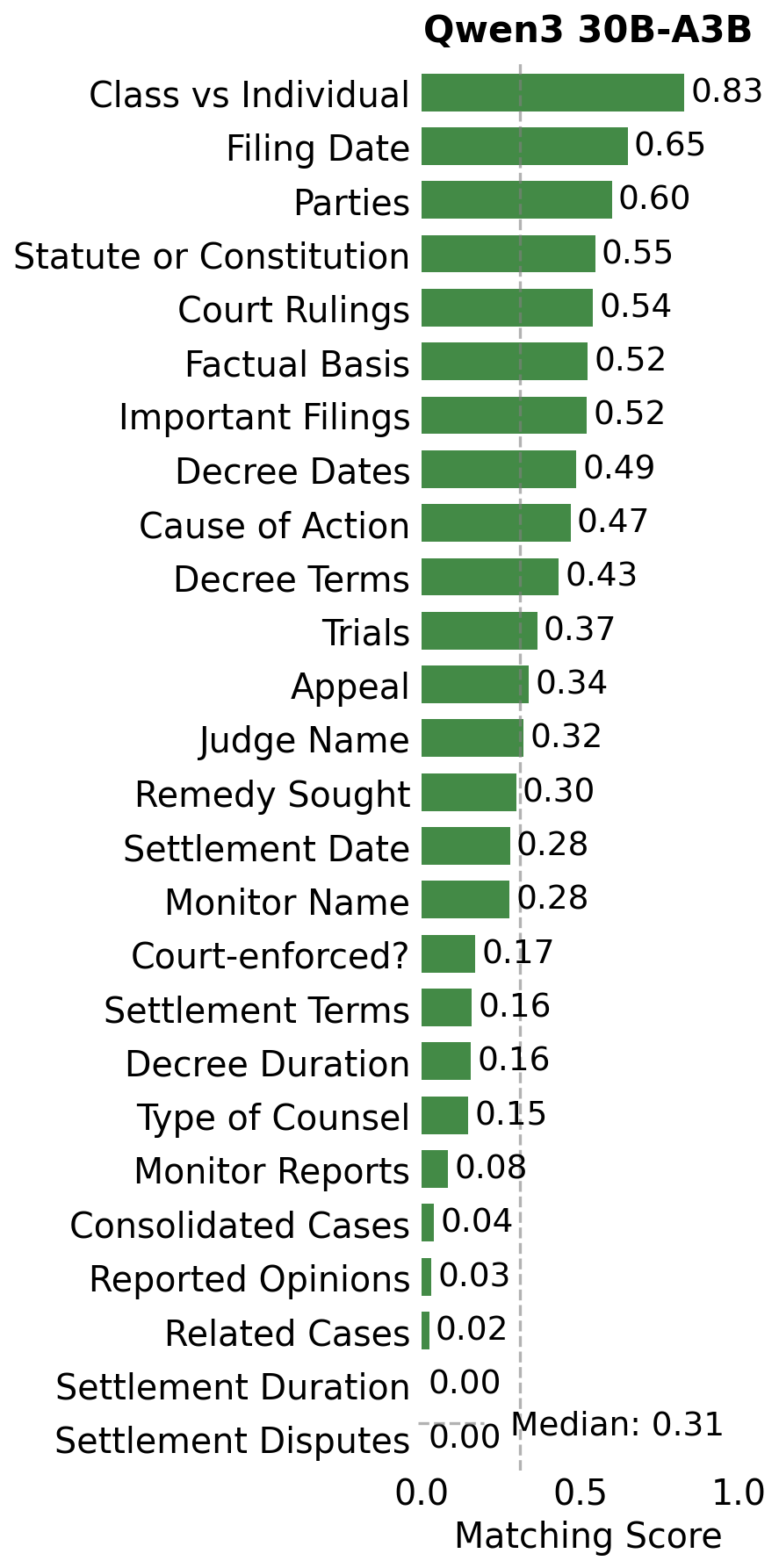}
  \end{subfigure}
  \hfill
  \begin{subfigure}[b]{0.24\textwidth}
    \centering
    \includegraphics[width=\linewidth]{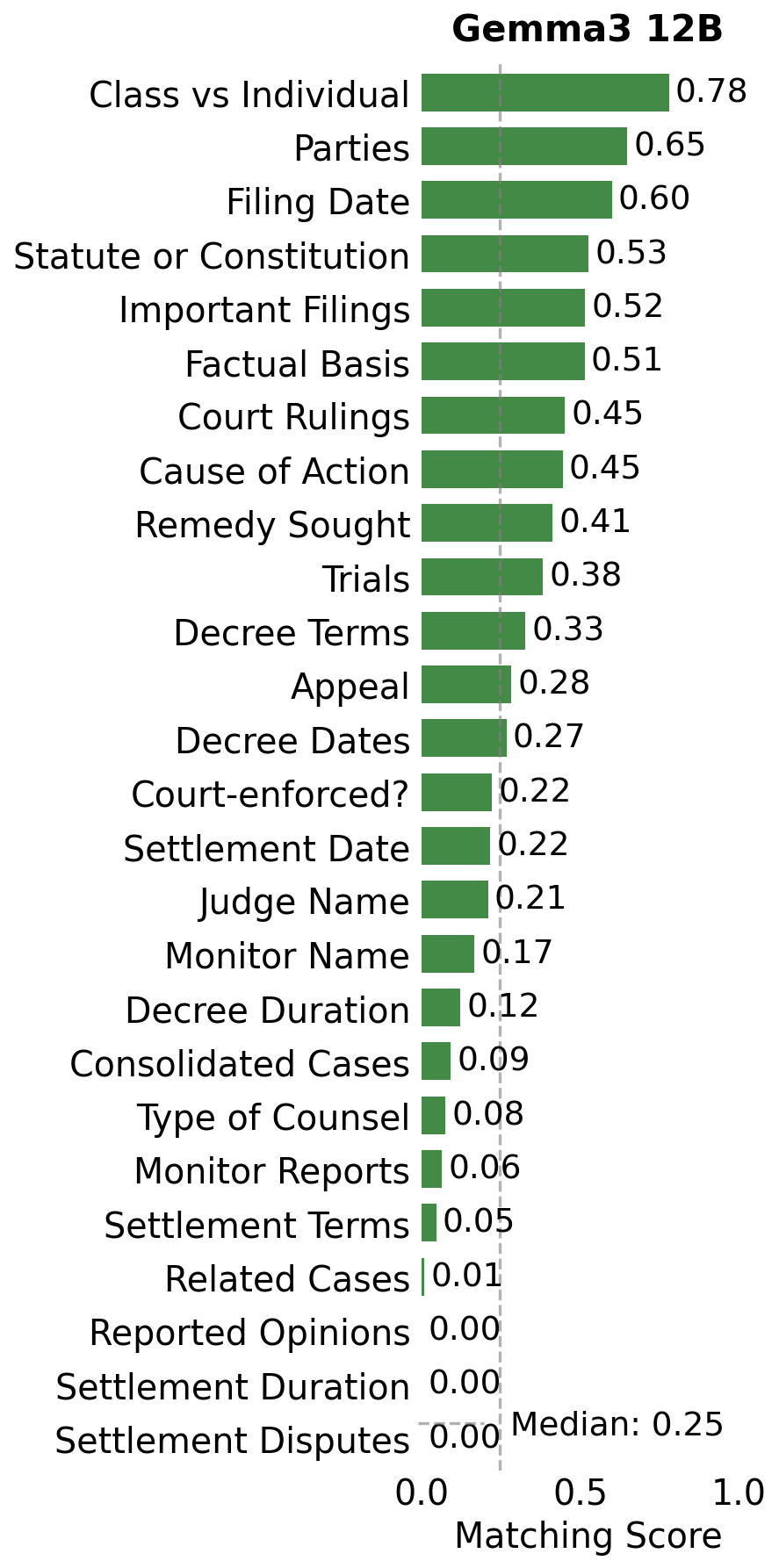}
  \end{subfigure}
  \hfill
  \begin{subfigure}[b]{0.24\textwidth}
    \centering
    \includegraphics[width=\linewidth]{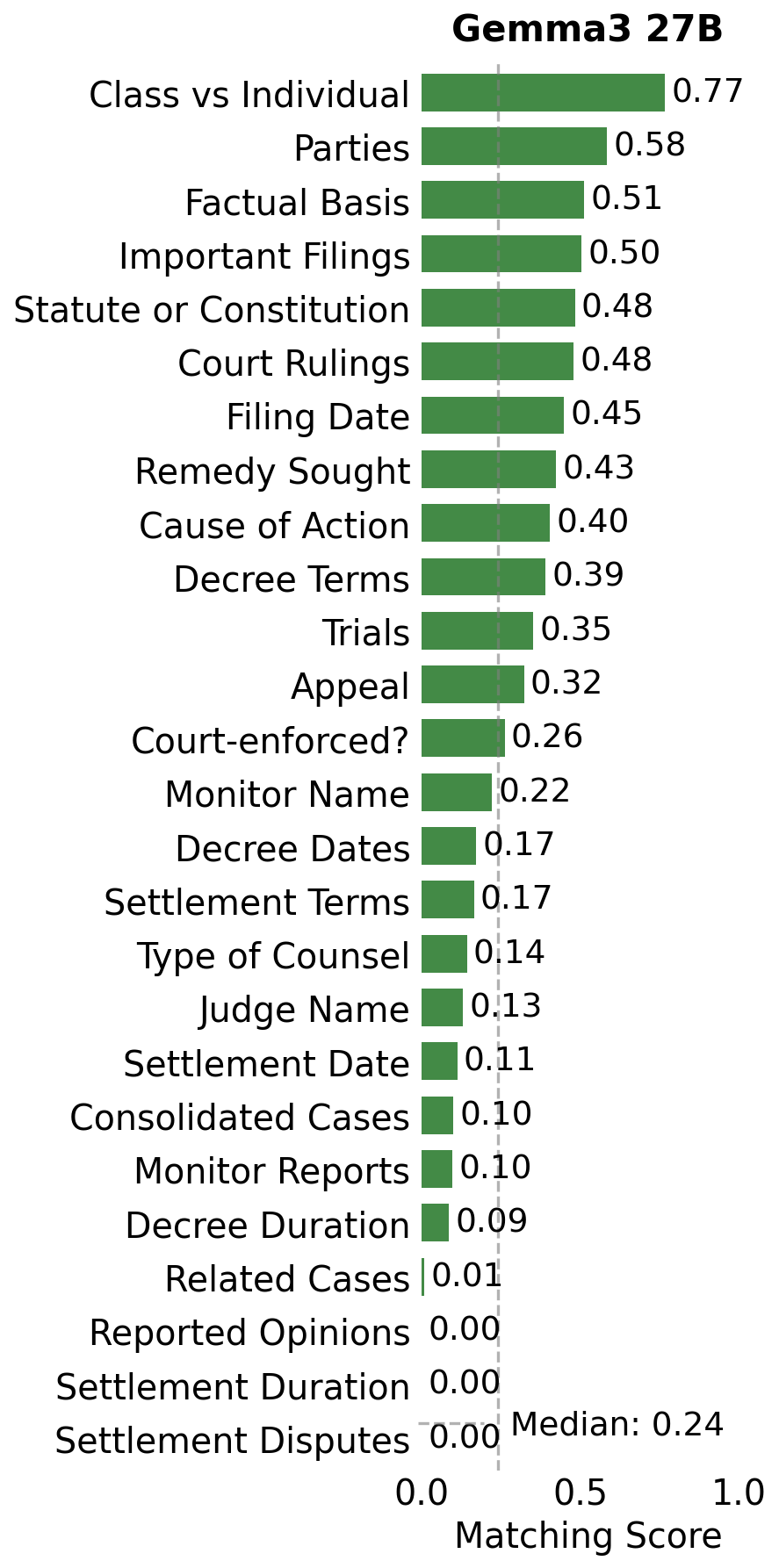}
  \end{subfigure}
  \vspace{-4pt}
  \caption{Checklist item-level performance for each LLM in the checklist evaluation. The metric is the matching score $m_i$. This figure shows results for Qwen3 14B, Qwen3 30B-A3B, Gemma3 12B and Gemma3 27B.}
  \label{fig:model_checklist_item_level_3}
\end{figure*}

\rev{Figures \ref{fig:model_checklist_item_level_1}, \ref{fig:model_checklist_item_level_2}, and \ref{fig:model_checklist_item_level_3} present the checklist item-level performance for each of the 12 LLMs we evaluate.}

\begin{figure*}[!t]
  \centering
  \includegraphics[width=0.99\textwidth]{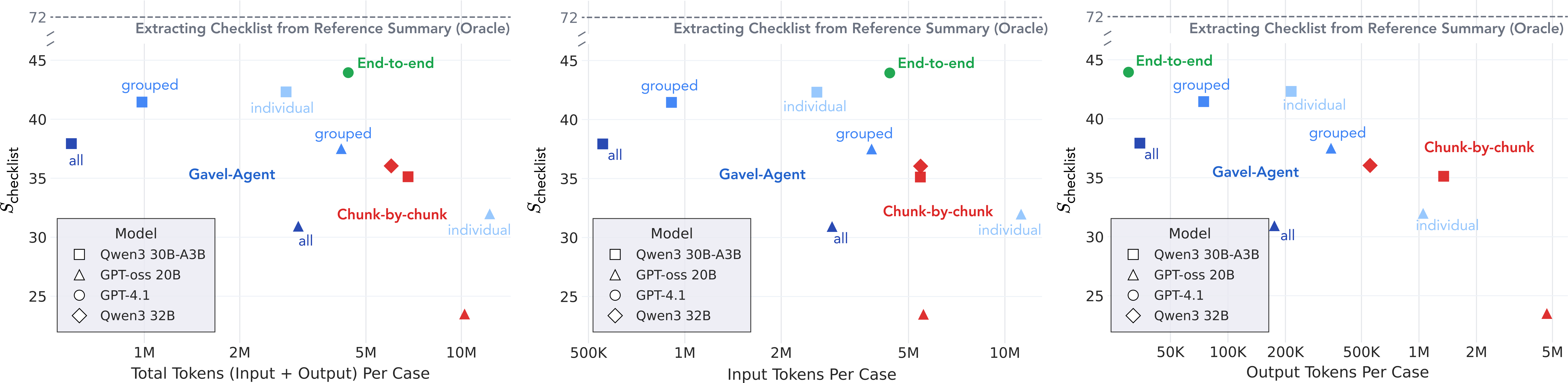}
  \vspace{-5pt}
  \caption{$S_\text{checklist}$ versus total token, input token, and output token usage for different methods extracting from case documents. Oracle is LLM (GPT-5, the best extractor we tested among 5 models) extracting checklist directly from human summary of the case.}
  \label{fig:meta_eval_from_docs_full}
  \vspace{0pt}
\end{figure*}

\begin{figure*}[!t]
  \centering
  \includegraphics[width=0.99\textwidth]{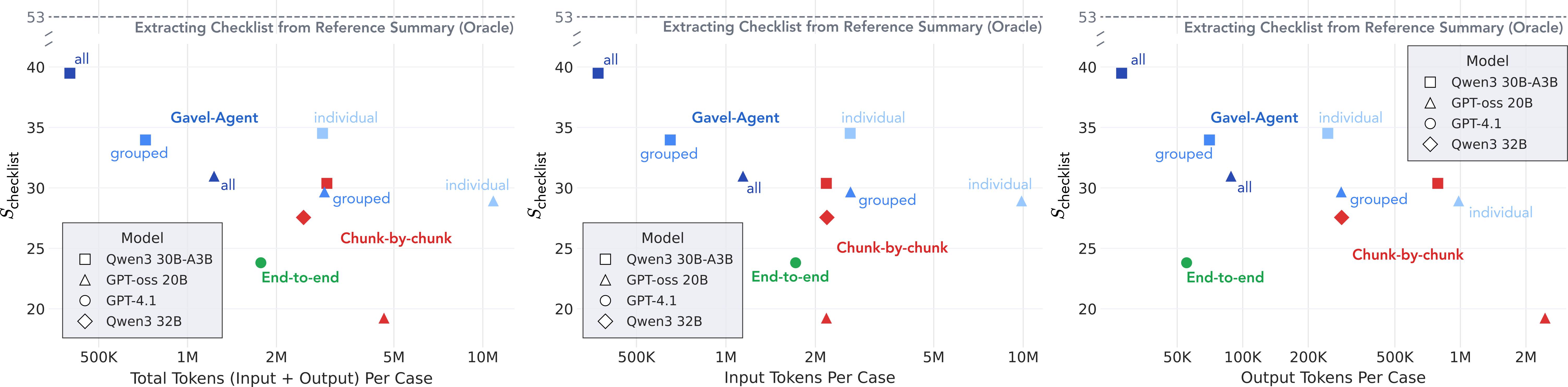}
  \vspace{-5pt}
  \caption{$S_\text{checklist}$ versus total token, input token, and output token usage for different methods extracting from medical review. Oracle is an LLM (Qwen3 32B, the best extractor we tested among 5 models) extracting checklist directly from human plain language summary of the review.}
  \label{fig:meta_eval_from_docs_full_medical}
  \vspace{0pt}
\end{figure*}

Figure \ref{fig:meta_eval_from_docs_full} presents the checklist extraction performance $S_\text{checklist}$ versus total, input, output token usage for each method extracting checklist from case documents. Figure \ref{fig:precision_vs_recall_from_docs} shows the precision vs recall for each method.
Figures \ref{fig:meta_eval_from_docs_full_medical} and \ref{fig:precision_vs_recall_from_docs_medical} present the same analyses for the medical domain, where methods extract checklists from a medical review.
 
\begin{figure*}[!t]
  \centering
  \includegraphics[width=0.99\textwidth]{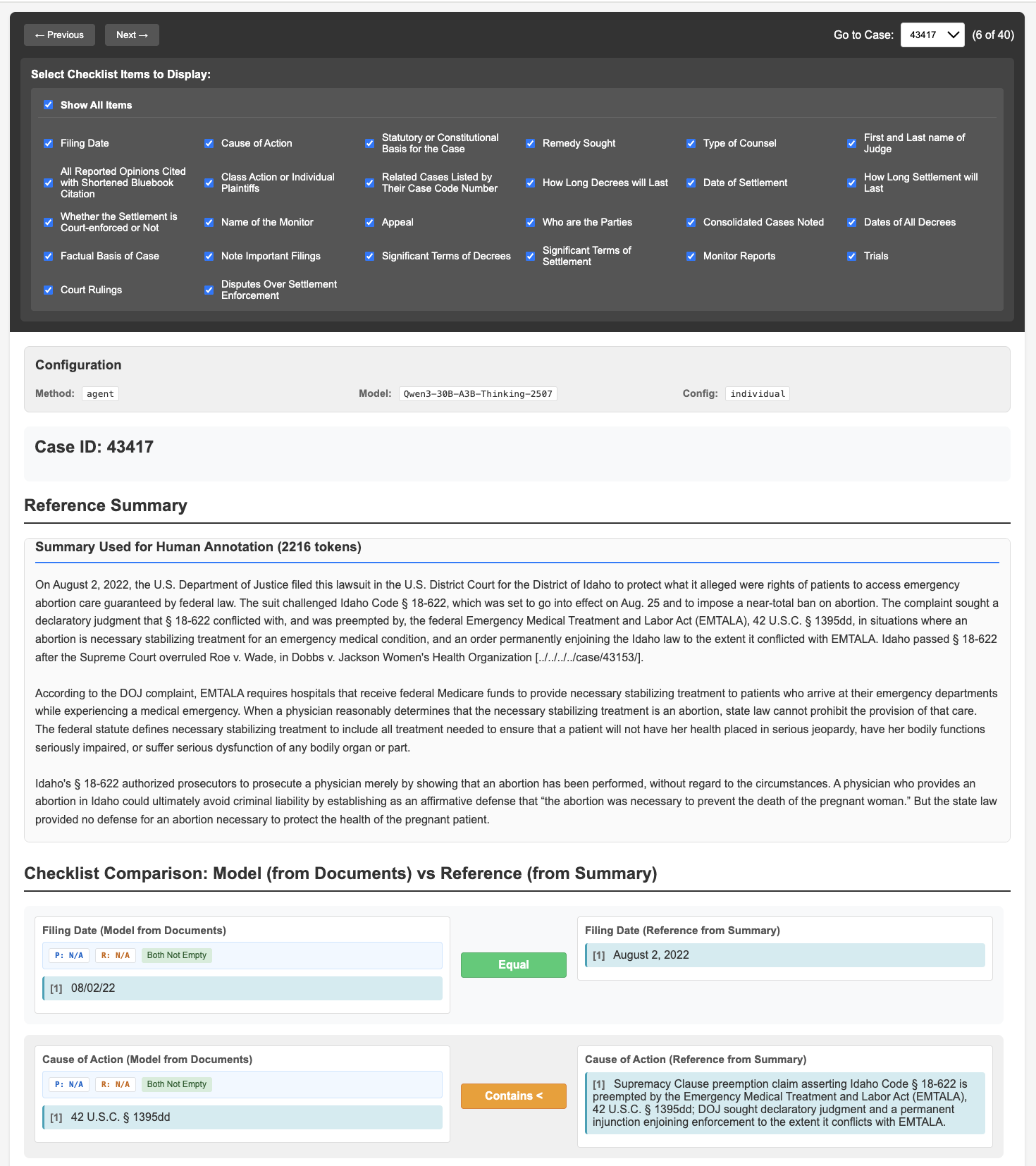}
  \vspace{-5pt}
  \caption{Screenshot of a visualization for one case, comparing checklists extracted directly from case documents by \fwnameAgent with Qwen3 30B-A3B (26 individual agents configuration) against the human-annotated checklist extracted from the case summary (figure 1 of 10).}
  \label{fig:checklist_example_1}
  \vspace{0pt}
\end{figure*}

\begin{figure*}[!t]
  \centering
  \includegraphics[width=0.99\textwidth]{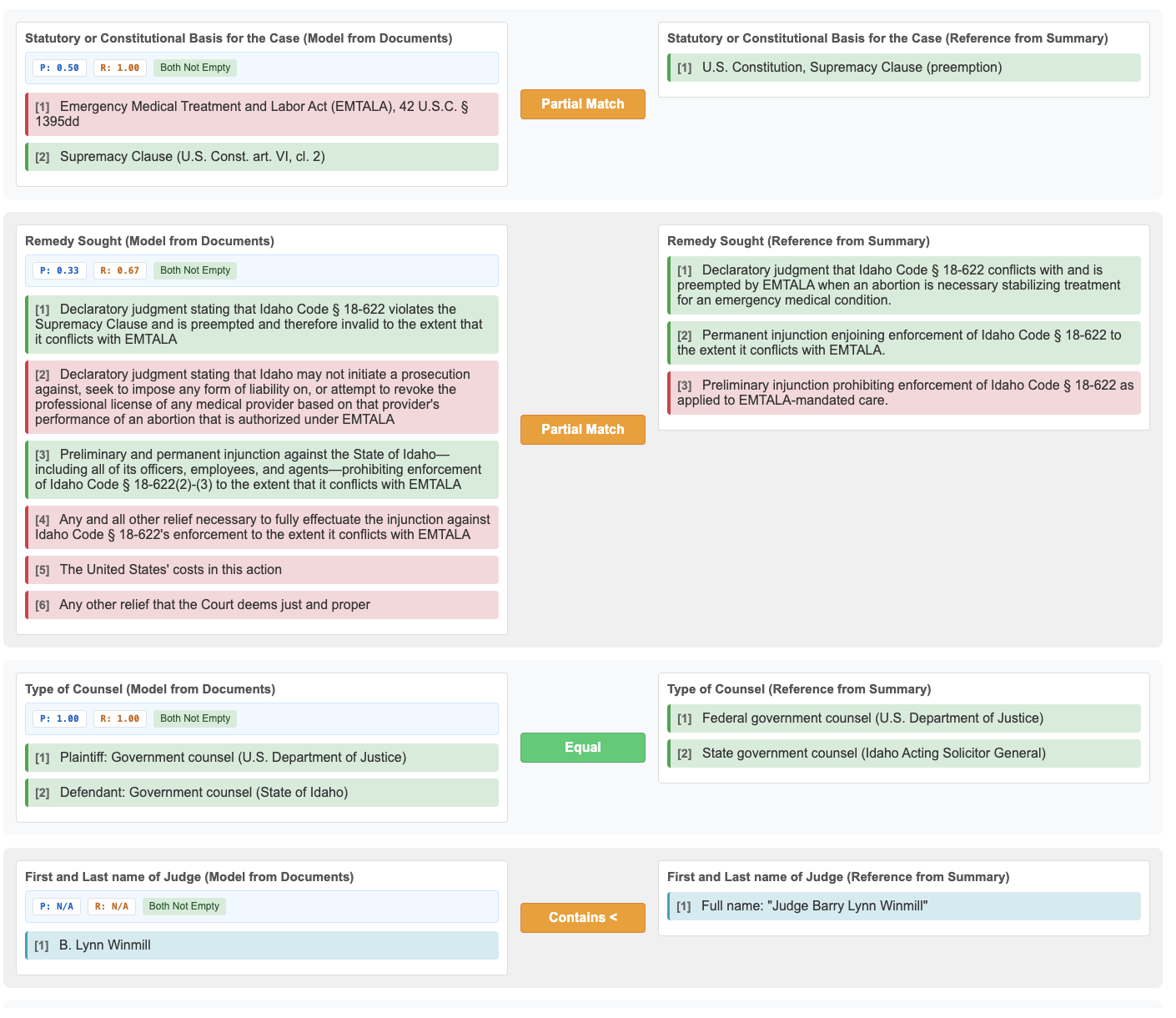}
  \vspace{-5pt}
  \caption{Screenshot of a visualization for one case, comparing checklists extracted directly from case documents by \fwnameAgent with Qwen3 30B-A3B (26 individual agents configuration) against the human-annotated checklist extracted from the case summary (figure 2 of 10).}
  \label{fig:checklist_example_2}
  \vspace{0pt}
\end{figure*}

\begin{figure*}[!t]
  \centering
  \includegraphics[width=0.99\textwidth]{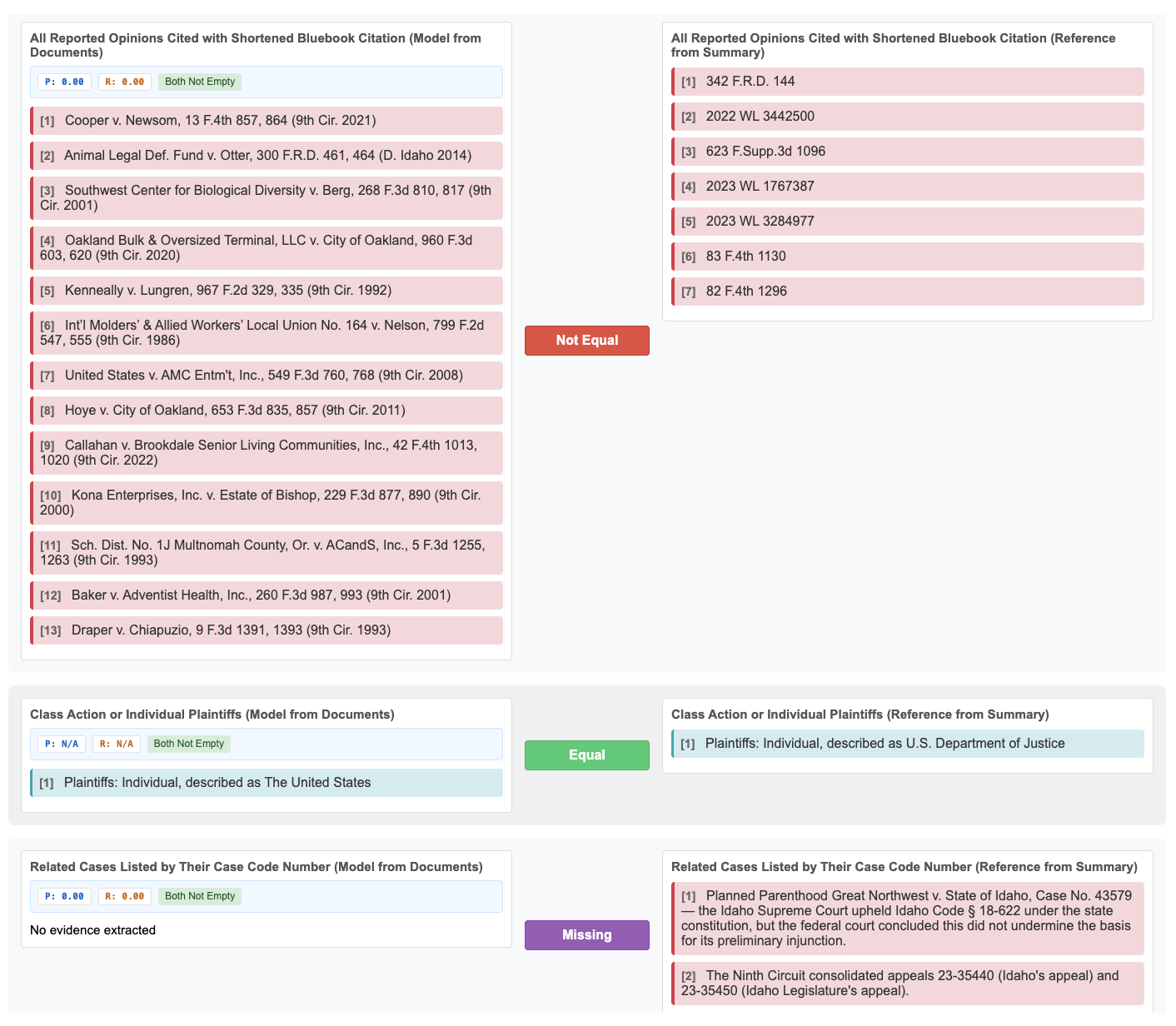}
  \vspace{-5pt}
  \caption{Screenshot of a visualization for one case, comparing checklists extracted directly from case documents by \fwnameAgent with Qwen3 30B-A3B (26 individual agents configuration) against the human-annotated checklist extracted from the case summary (figure 3 of 10).}
  \label{fig:checklist_example_3}
  \vspace{0pt}
\end{figure*}

\begin{figure*}[!t]
  \centering
  \includegraphics[width=0.99\textwidth]{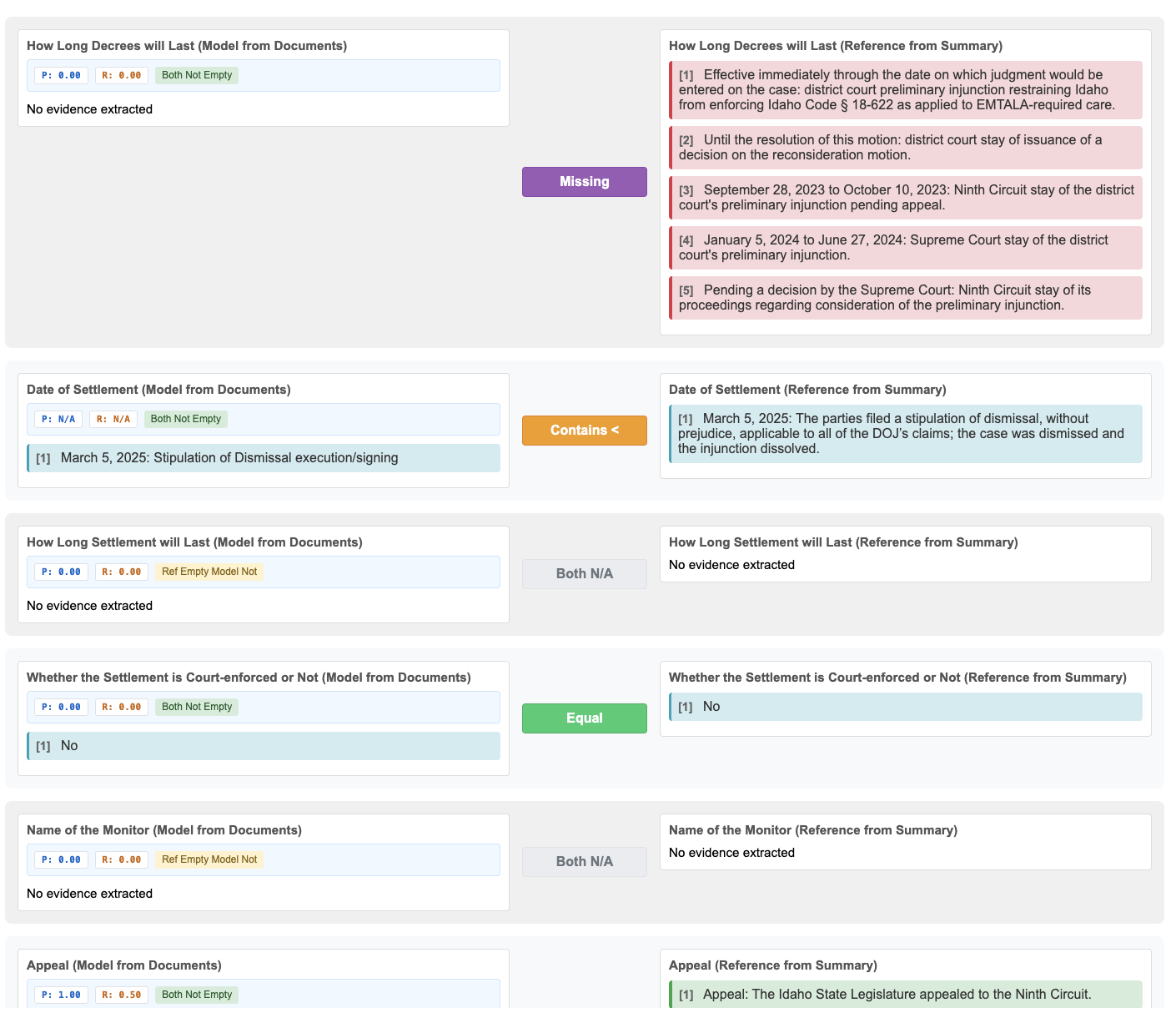}
  \vspace{-5pt}
  \caption{Screenshot of a visualization for one case, comparing checklists extracted directly from case documents by \fwnameAgent with Qwen3 30B-A3B (26 individual agents configuration) against the human-annotated checklist extracted from the case summary (figure 4 of 10).}
  \label{fig:checklist_example_4}
  \vspace{0pt}
\end{figure*}

\begin{figure*}[!t]
  \centering
  \includegraphics[width=0.99\textwidth]{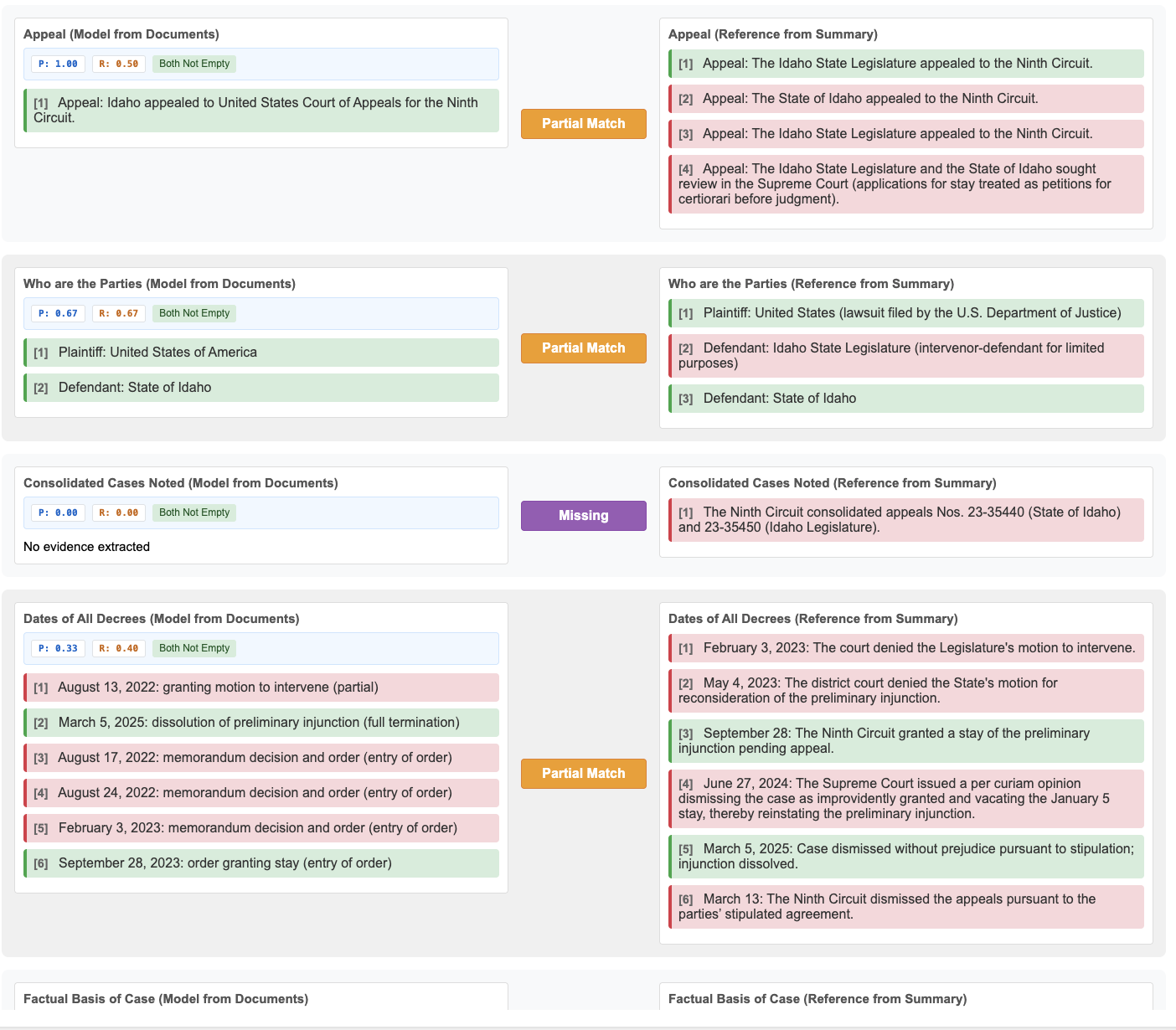}
  \vspace{-5pt}
  \caption{Screenshot of a visualization for one case, comparing checklists extracted directly from case documents by \fwnameAgent with Qwen3 30B-A3B (26 individual agents configuration) against the human-annotated checklist extracted from the case summary (figure 5 of 10).}
  \label{fig:checklist_example_5}
  \vspace{0pt}
\end{figure*}

\begin{figure*}[!t]
  \centering
  \includegraphics[width=0.99\textwidth]{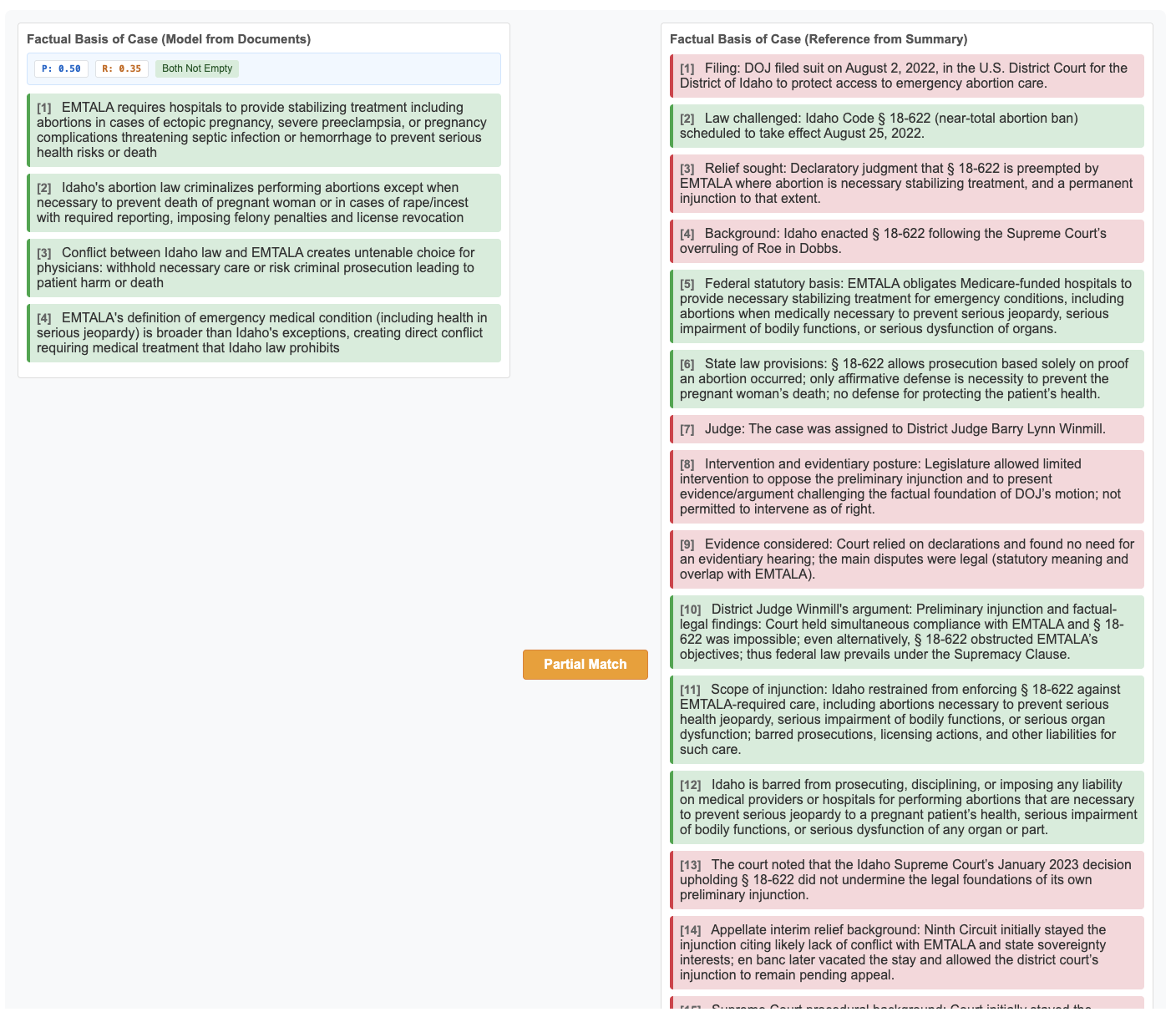}
  \vspace{-5pt}
  \caption{Screenshot of a visualization for one case, comparing checklists extracted directly from case documents by \fwnameAgent with Qwen3 30B-A3B (26 individual agents configuration) against the human-annotated checklist extracted from the case summary (figure 6 of 10).}
  \label{fig:checklist_example_6}
  \vspace{0pt}
\end{figure*}

\begin{figure*}[!t]
  \centering
  \includegraphics[width=0.99\textwidth]{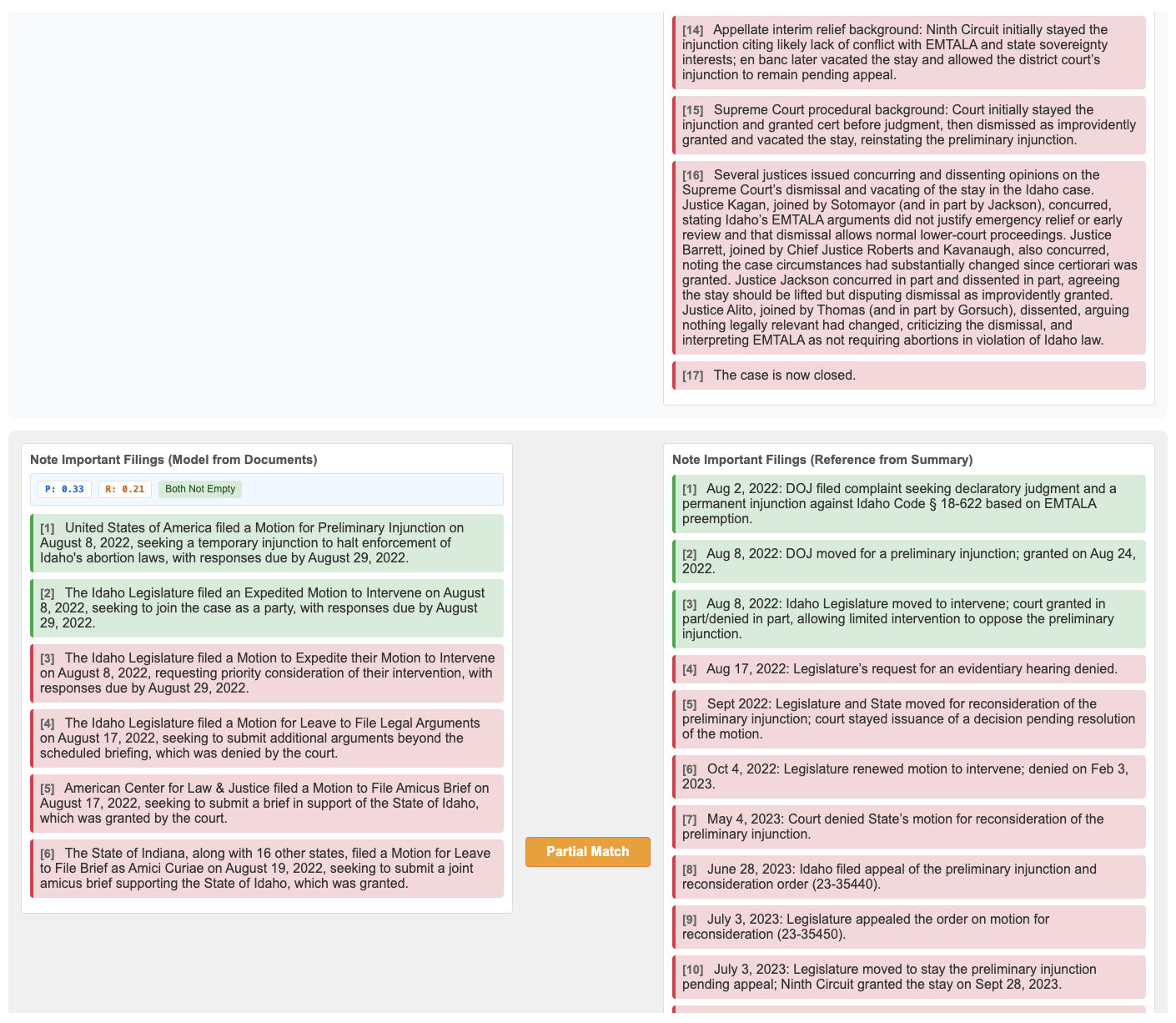}
  \vspace{-5pt}
  \caption{Screenshot of a visualization for one case, comparing checklists extracted directly from case documents by \fwnameAgent with Qwen3 30B-A3B (26 individual agents configuration) against the human-annotated checklist extracted from the case summary (figure 7 of 10).}
  \label{fig:checklist_example_7}
  \vspace{0pt}
\end{figure*}

\begin{figure*}[!t]
  \centering
  \includegraphics[width=0.99\textwidth]{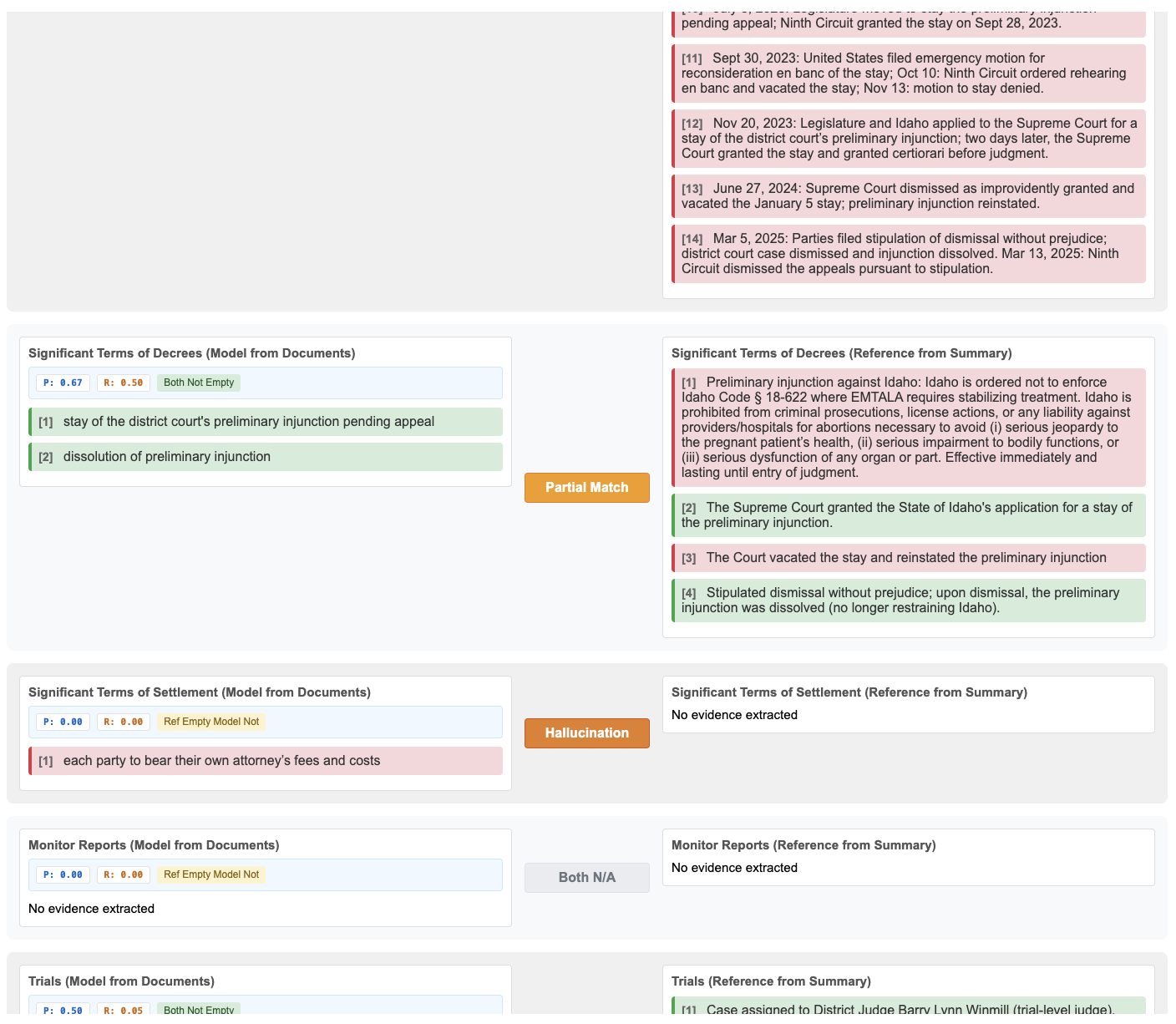}
  \vspace{-5pt}
  \caption{Screenshot of a visualization for one case, comparing checklists extracted directly from case documents by \fwnameAgent with Qwen3 30B-A3B (26 individual agents configuration) against the human-annotated checklist extracted from the case summary (figure 8 of 10).}
  \label{fig:checklist_example_8}
  \vspace{0pt}
\end{figure*}

\begin{figure*}[!t]
  \centering
  \includegraphics[width=0.99\textwidth]{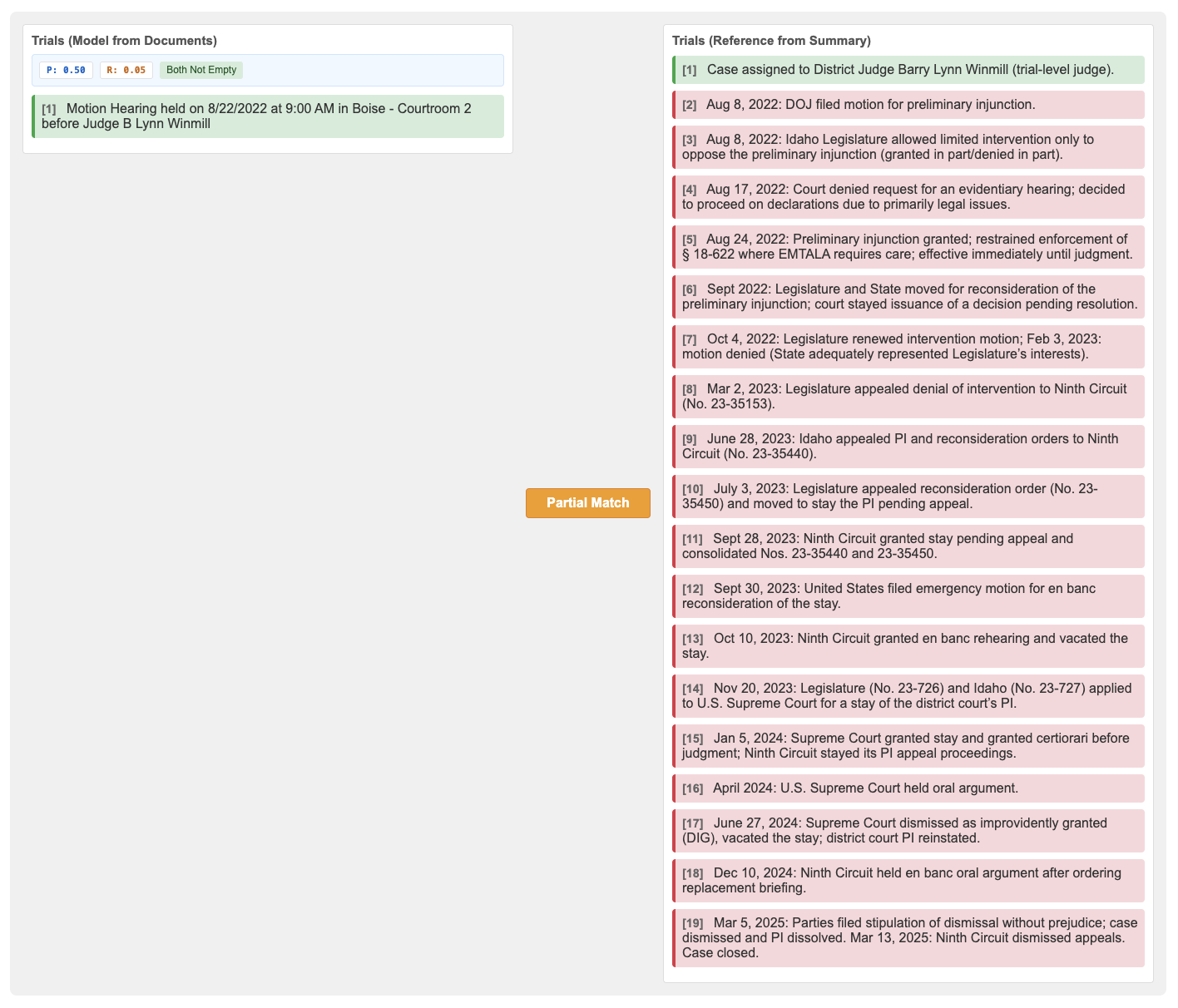}
  \vspace{-5pt}
  \caption{Screenshot of a visualization for one case, comparing checklists extracted directly from case documents by \fwnameAgent with Qwen3 30B-A3B (26 individual agents configuration) against the human-annotated checklist extracted from the case summary (figure 9 of 10).}
  \label{fig:checklist_example_9}
  \vspace{0pt}
\end{figure*}

\begin{figure*}[!t]
  \centering
  \includegraphics[width=0.99\textwidth]{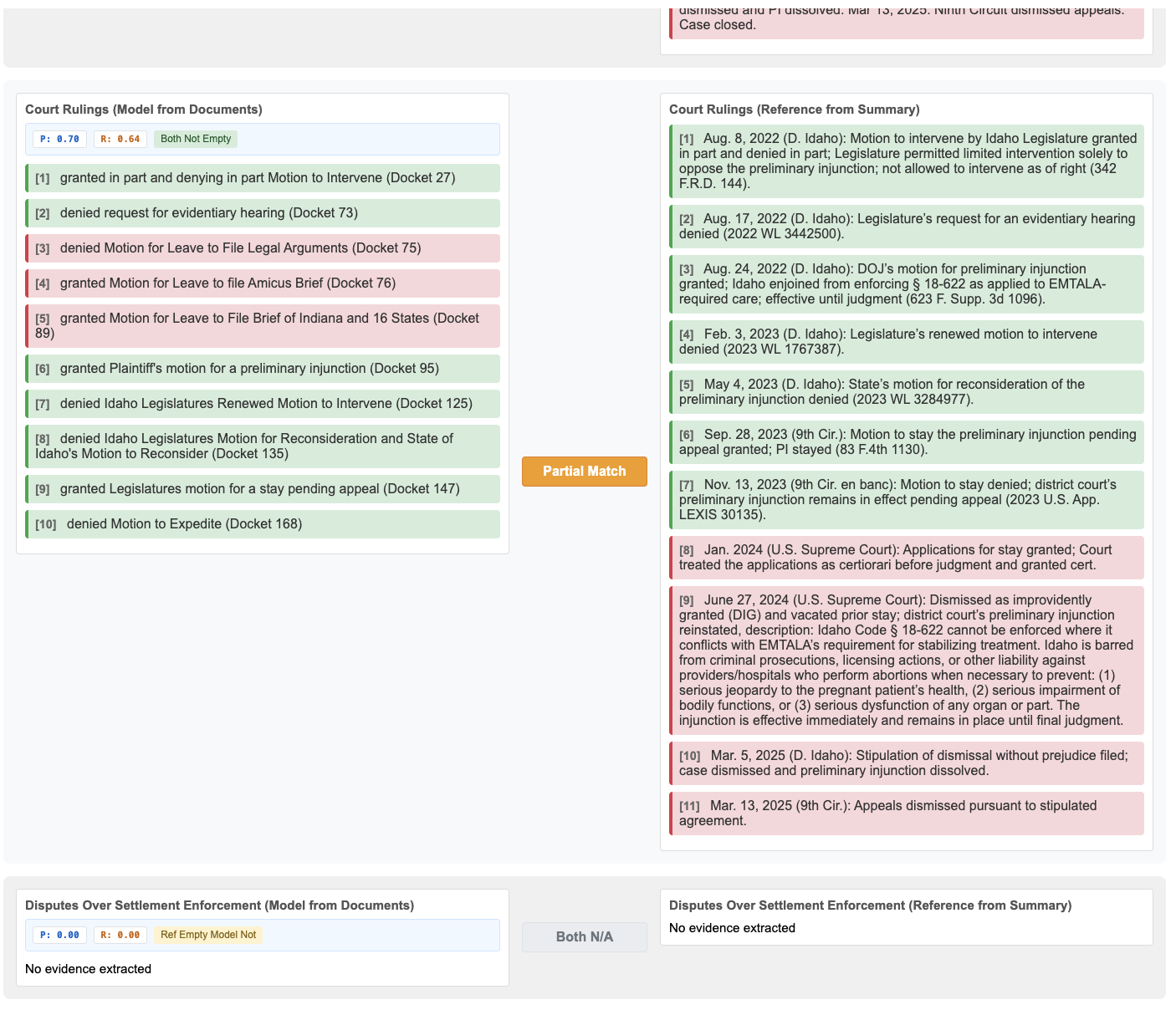}
  \vspace{-5pt}
  \caption{Screenshot of a visualization for one case, comparing checklists extracted directly from case documents by \fwnameAgent with Qwen3 30B-A3B (26 individual agents configuration) against the human-annotated checklist extracted from the case summary (figure 10 of 10).}
  \label{fig:checklist_example_10}
  \vspace{0pt}
\end{figure*}

\rev{Figures \ref{fig:checklist_example_1} to \ref{fig:checklist_example_10} show a randomly sampled case, comparing checklists extracted directly from case documents by \fwnameAgent with Qwen3 30B-A3B (26-agent configuration) against the human-annotated checklist extracted from the case summary.}

\begin{figure*}[!t]
  \centering
  \includegraphics[width=0.99\textwidth]{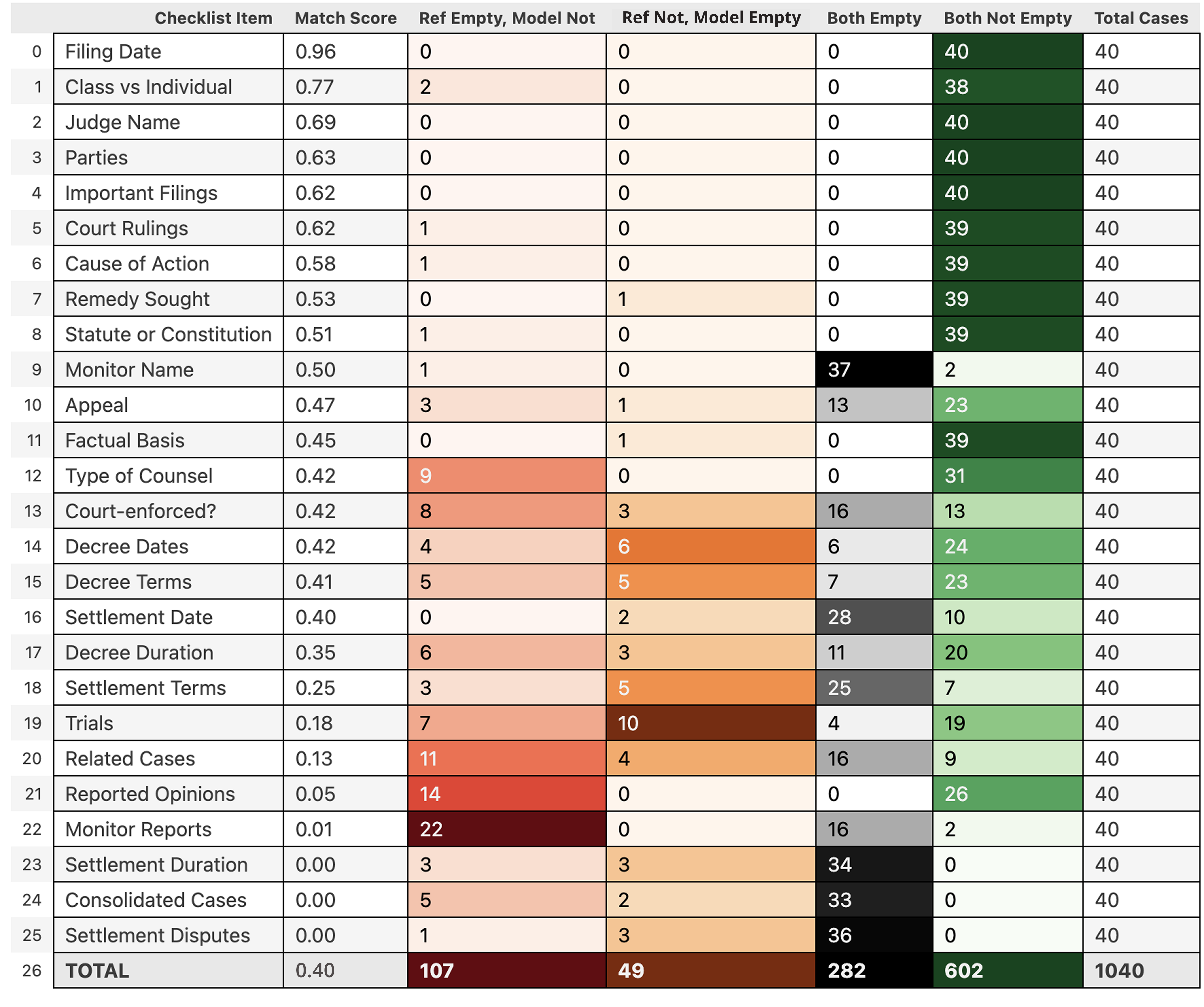}
  \vspace{-5pt}
  \caption{Checklist item–level performance and statistics for end-to-end checklist extraction from full case documents using GPT-4.1. The table reports the matching score $m_i$ for each checklist item, along with counts of reference–model value occurrences. For example, “Ref Empty, Model Not” denotes number of cases where the human reference value is empty but the model extracts some value.}
  \label{fig:by-item-from-documents-end-to-end}
  \vspace{0pt}
\end{figure*}

\begin{figure*}[!t]
  \centering
  \includegraphics[width=0.99\textwidth]{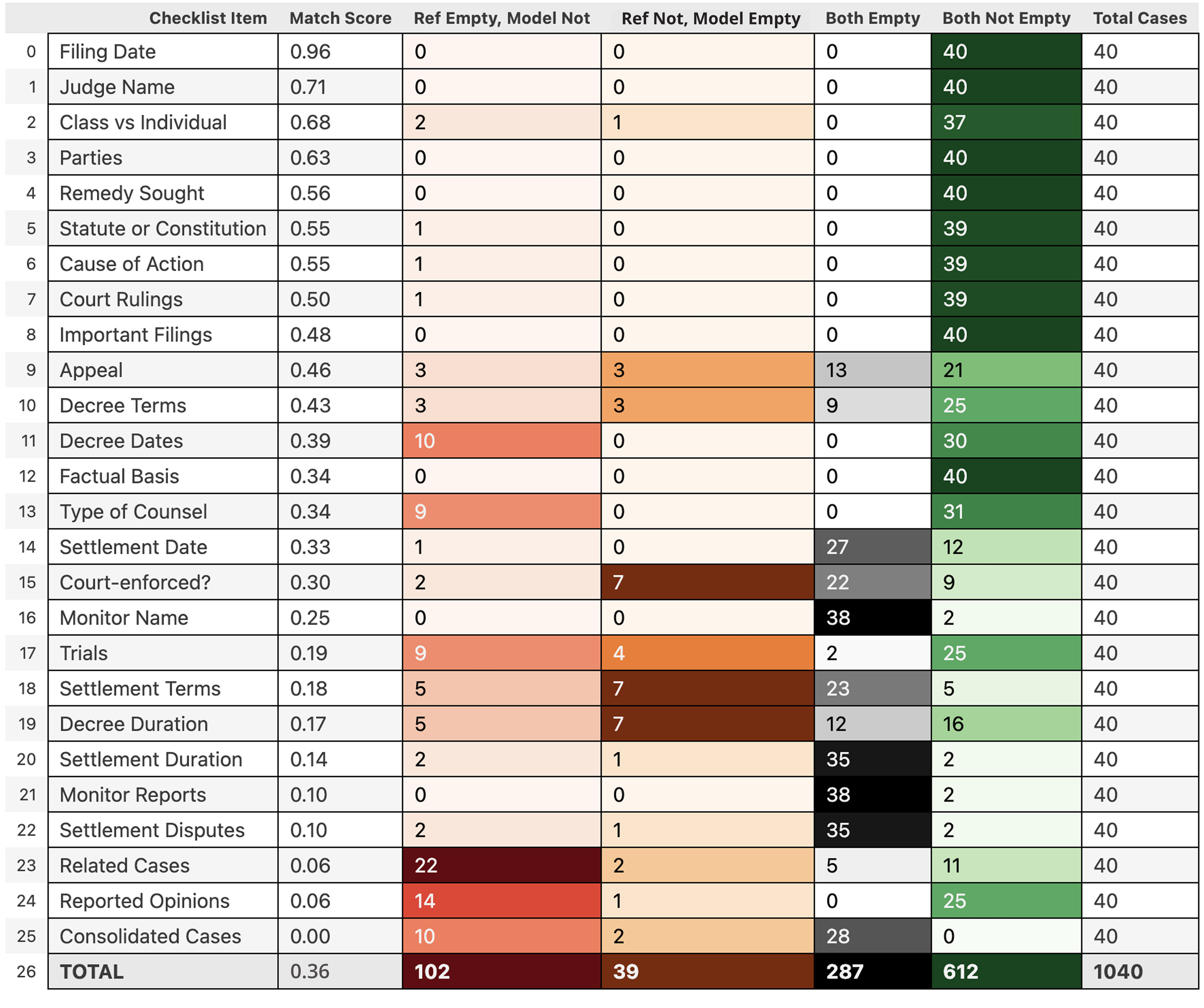}
  \vspace{-5pt}
  \caption{Checklist item–level performance and statistics for \fwnameAgent checklist extraction from full case documents using Qwen3 30B-A3B with 26 individual agent setup. The table reports the matching score $m_i$ for each checklist item, along with counts of reference–model value occurrences. For example, “Ref Empty, Model Not” denotes number of cases where the human reference value is empty but the model extracts some value.}
  \label{fig:by-item-from-documents-agent}
  \vspace{0pt}
\end{figure*}

\begin{figure*}[!t]
  \centering
  \includegraphics[width=0.99\textwidth]{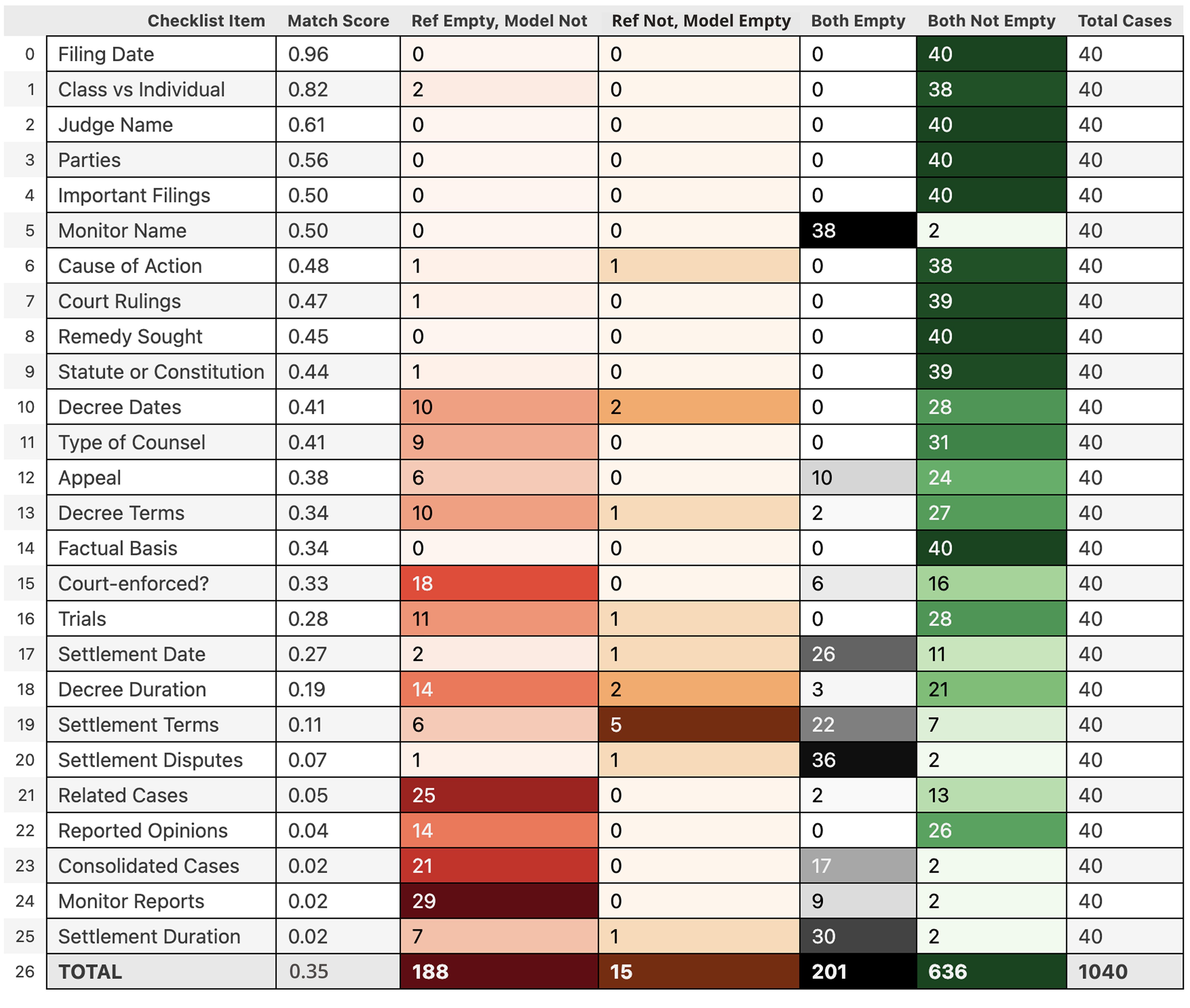}
  \vspace{-5pt}
  \caption{Checklist item–level performance and statistics for chunk-by-chunk checklist extraction from full case documents using Qwen3 30B-A3B. The table reports the matching score $m_i$ for each checklist item, along with counts of reference–model value occurrences. For example, “Ref Empty, Model Not” denotes number of cases where the human reference value is empty but the model extracts some value.}
  \label{fig:by-item-from-documents-chunk-by-chunk}
  \vspace{0pt}
\end{figure*}

Figures \ref{fig:by-item-from-documents-end-to-end} to \ref{fig:by-item-from-documents-chunk-by-chunk} show checklist item–level performance and statistics for checklist extraction from case documents, compared against human-extracted checklists derived from case summaries. Compared to end-to-end extraction and \fwnameAgent, the chunk-by-chunk method is more prone to over-extraction, as evidenced by a substantially higher number of cases where the human reference is empty but the model extracts a value (``Ref Empty, Model Not'' column).

\section{Implementation Details}

For all language models, we use a temperature of 0.7 and top-p of 1, except for GPT-5 (where temperature cannot be changed and is fixed at 1) and Qwen3, for which we use a temperature of 0.6 and top-p of 0.95, following the official recommendations. For Gemini 2.5 Flash and Pro, we set the thinking budget to -1 (allowing the model to decide). For GPT-5, we use ``high'' thinking effort. For Claude Sonnet 4 and Opus 4.1, we set the thinking budget to 10,000.

We use the following versions of the proprietary models: gpt-4.1-2025-04-14, gpt-5-2025-08-07, claude-sonnet-4-20250514, claude-opus-4-1-20250805, gemini-2.5-flash (June 2025), and gemini-2.5-pro (June 2025). For open-source models, we use the instruction-tuned version of Gemma3 (Gemma3-it) and Qwen3-30B-A3B-Thinking-2507 for Qwen3 30B-A3B. Open-source models are run through vLLM \cite{kwon2023efficient} on 4 A40 GPUs. For all reasoning models such as Qwen3, we use the reasoning mode. Due to compute constraints, we could not run models larger than these, such as GPT-oss 120B.
The total API costs is \$1,800 USD.

For \fwnameAgent, we implement tool calls using each model’s native format: ChatML for Qwen3 and Harmony for GPT-oss.
\section{Medical Checklist Definitions}
\label{app:medical_checklist_definitions}

The following are the definitions of the 29 medical checklist items used in our work. We group them into 6 categories.

\begin{enumerate}[leftmargin=*,itemsep=1pt,parsep=1pt,topsep=2pt]

\item[\textbf{A.}] \textbf{Clinical Question Structure}
\begin{enumerate}[label=\arabic*.,itemsep=0pt,parsep=0pt,topsep=1pt]
    \item \textbf{Condition Defined}: Description of the health condition, disease, disorder, or clinical problem being addressed.
    \item \textbf{Population Defined}: Specifies the population to whom the evidence applies, including demographic characteristics, age groups, clinical status, disease severity, or other relevant patient characteristics.
    \item \textbf{Intervention Defined}: Identifies the treatment, exposure, therapy, procedure, preventive measure, or diagnostic approach being evaluated.
    \item \textbf{Comparator Defined}: Identifies the comparison condition such as placebo, standard care, no treatment, alternative treatment, or control intervention.
    \item \textbf{Timing of Intervention Defined}: Specifies when, how frequently, or for how long the intervention or exposure was administered.
    \item \textbf{Purpose/Aim Stated}: Explicit statement of the objective, research question, or clinical aim of the review or study.
\end{enumerate}

\item[\textbf{B.}] \textbf{Outcomes Specification}
\begin{enumerate}[label=\arabic*.,resume,itemsep=0pt,parsep=0pt,topsep=1pt]
    \item \textbf{Primary Benefit Outcome Identified}: Explicit identification of the main beneficial outcome used to evaluate effectiveness (e.g., symptom reduction, disease remission, survival improvement).
    \item \textbf{Secondary Benefit Outcome Identified}: Additional beneficial outcomes evaluated separately from the primary outcome.
    \item \textbf{Harm/Safety Outcome Identified}: Explicit identification of adverse events, side effects, complications, or other unwanted outcomes associated with the intervention or exposure.
    \item \textbf{Outcome Timeframe Stated}: Specifies the timing or duration of outcome assessment (e.g., short-term, long-term, ``after 12 months,'' ``by two years'').
    \item \textbf{Outcome Definition Clarified}: Explains the meaning or clinical interpretation of an outcome measure using plain language, examples, or defining criteria.
\end{enumerate}

\item[\textbf{C.}] \textbf{Results Reporting}
\begin{enumerate}[label=\arabic*.,resume,itemsep=0pt,parsep=0pt,topsep=1pt]
    \item \textbf{Benefit Direction Stated}: Indicates whether the intervention or exposure increased, decreased, improved, worsened, or produced no meaningful difference in beneficial outcomes.
    \item \textbf{Harm Direction Stated}: Indicates whether harms, adverse events, or safety risks increased, decreased, or showed no meaningful difference.
    \item \textbf{Magnitude Descriptor Used}: Uses qualitative magnitude descriptors such as ``slightly,'' ``moderately,'' ``substantially,'' ``little to no,'' or ``large'' to characterize effect size or clinical importance.
    \item \textbf{Quantitative Data Provided}: Reports numerical findings such as percentages, means, confidence intervals, risk ratios, odds ratios, hazard ratios, or absolute effect estimates.
    \item \textbf{Evidence Absence Stated}: Explicit statement that evidence, studies, outcome data, or sufficient information were not available.
\end{enumerate}

\item[\textbf{D.}] \textbf{Certainty and Evidence Quality}
\begin{enumerate}[label=\arabic*.,resume,itemsep=0pt,parsep=0pt,topsep=1pt]
    \item \textbf{Certainty Level Stated}: Expresses the degree of confidence or certainty in the evidence using qualitative or formal evidence-certainty terminology (e.g., ``high certainty,'' ``low certainty,'' ``very uncertain'').
    \item \textbf{Reason for Downgrading Certainty Given}: Explains factors reducing confidence in the evidence such as study limitations, risk of bias, inconsistency, indirectness, imprecision, or small sample size.
    \item \textbf{Study Design Identified}: Specifies the study methodology or design used to generate evidence, such as randomized controlled trials, cohort studies, case-control studies, or observational studies.
    \item \textbf{Number of Studies Reported}: States the total number of included studies contributing evidence to the review or analysis.
    \item \textbf{Total Sample Size Reported}: States the total number of participants, patients, or observations included across studies.
    \item \textbf{Search Strategy Mentioned}: Mentions the process used to identify studies or evidence sources, including literature searches, databases searched, or systematic search methods.
    \item \textbf{Inclusion Criteria Implied or Stated}: Specifies eligibility criteria or characteristics required for studies, participants, interventions, or outcomes to be included.
    \item \textbf{Evidence Gaps Identified}: Notes areas where evidence is missing, insufficient, inconsistent, underpowered, or unavailable for particular outcomes or populations.
\end{enumerate}

\item[\textbf{E.}] \textbf{Contextual and Background Information}
\begin{enumerate}[label=\arabic*.,resume,itemsep=0pt,parsep=0pt,topsep=1pt]
    \item \textbf{Condition Background Explained}: Provides contextual information about the disease or condition, including mechanism, prevalence, progression, risk factors, or clinical significance.
    \item \textbf{Intervention Rationale Explained}: Explains the biological, clinical, or theoretical reasoning for why the intervention or exposure may be effective.
    \item \textbf{Definition of Technical Terms Provided}: Defines medical, statistical, or technical terminology using plain language explanations, parenthetical clarifications, or simplified descriptions.
\end{enumerate}

\item[\textbf{F.}] \textbf{Applicability and Currency}
\begin{enumerate}[label=\arabic*.,resume,itemsep=0pt,parsep=0pt,topsep=1pt]
    \item \textbf{Applicability/Generalizability Implied or Stated}: Indicates the populations, settings, or clinical contexts to which the findings are applicable, including limitations to external validity or generalizability.
    \item \textbf{Evidence Currency Stated}: Specifies how current or up-to-date the evidence review is, including search dates, publication windows, or evidence cut-off dates.
\end{enumerate}

\end{enumerate}
\section{Prompts}
\label{app:prompts}

The following lists the prompts used in our paper.

\paragraph{Prompts used in \fwnameRef.}
Figure~\ref{prompt:checklist_extraction} shows the prompt for extracting checklist items from summaries.
Figures~\ref{prompt:checklist_comparison_single_value} and \ref{prompt:checklist_comparison_multi_value} show the prompts for comparing single-value and multi-value checklist items, respectively.
Figure~\ref{prompt:residual_facts_extraction} shows the prompt for extracting residual facts not covered by checklist items or their supporting text.
Figure~\ref{prompt:writing_style_rating} shows the prompt for rating writing style similarity between two summaries across five aspects.

\paragraph{Prompt for summarization.}
Figure~\ref{prompt:summarization} shows the prompt for legal summarization.

\paragraph{Prompts for checklist extraction from case documents.}
Figures~\ref{prompt:end-to-end-extraction-part-1} and \ref{prompt:end-to-end-extraction-part-2} present the prompts for the end-to-end method.
Figure~\ref{prompt:chunk-by-chunk-extraction} presents the prompt for the chunk-by-chunk method.
Figures~\ref{prompt:agent-system-prompt-part-1}, \ref{prompt:agent-system-prompt-part-2}, and \ref{prompt:agent-system-prompt-part-3} present the system prompts used in \fwnameAgent.

\begin{figure*}[htbp]
  \centering
  \begin{AIbox}[width=\textwidth]{Prompt for Extracting Checklist from Summary} 
  \footnotesize
   \begin{Prompt}
You are assisting a lawyer in extracting key information from a legal case summary. Given a case summary, identify {checklist_item_definition}
# Note: Do not make assumptions or add information that is not presented in the summary.

# Case Summary
{case_summary}

# Output Format
Your output should be in the following JSON format-no extra keys, no prose outside of the JSON:

```
{{
  "reasoning": "<brief analysis of the case summary and how you identified the relevant information or determined that none was present>",
  "extracted": [
    {{
      "evidence": [
        "<verbatim snippet 1>",
        "<verbatim snippet 2 (if multiple snippets are relevant)>"
        // ...
      ],
      "value": "<extracted information from the evidence>"
    }}
    // ...
  ]
}}
```
## Definitions of each part
- `reasoning`: A brief analysis of the case summary and how you identified the relevant information or determined that none was present.
- `extracted`: A list of one or more objects, each representing a distinct piece of information relevant to the checklist item (e.g., multiple court rulings, decree dates, or cited opinions). Always use a list, even if there is only one item.
- `evidence`: One or more exact text snippets copied from the case summary that support the extracted information. Always return as a list of strings.
- `value`: The extracted information.

## Rules for the JSON schema
1. **extracted** and **evidence** is always a list, even if they hold a single object.
2. Copy the **evidence** exactly as it appears in the case summary-no rewriting.
3. If the case summary contains no relevant information, output the **extracted** as an empty list:

```
{{
  "reasoning": "<brief analysis>",
  "extracted": []
}}
```
    \end{Prompt}
  \end{AIbox}
  \vspace{-5pt}
  \caption[]{}
  \label{prompt:checklist_extraction}
\end{figure*}

\begin{figure*}[htbp]
  \centering
  \begin{AIbox}[width=\textwidth]{Prompt for Comparing Single-Value Checklist Item} 
   \begin{Prompt}
You are given two pieces of legal information (A and B) about **{checklist_category}**, extracted from two summaries of the same case. Your task is to compare these pieces of information based on their **semantic meaning** - that is, what they actually convey, regardless of how they are worded or formatted.

# Information to Compare
## Information A:
{information_A}

## Information B:
{information_B}

# Relationship Options
Determine which of these four relationships best describes how A and B relate to each other:
1. **"A contains B"** - A includes all the information in B, plus additional information
2. **"B contains A"** - B includes all the information in A, plus additional information
3. **"A equals B"** - A and B convey the same information (semantically equivalent)
4. **"A and B are different"** - A and B contain different or conflicting information

# Output Format
Structure your response as follows:
**Reasoning:** Provide your detailed analysis of how the two pieces of information relate to each other

**Final Answer:** State one of the four options: "A contains B", "B contains A", "A equals B", or "A and B are different"
       \end{Prompt}
  \end{AIbox}
  \vspace{-5pt}
  \caption[]{}
  \label{prompt:checklist_comparison_single_value}
\end{figure*}

\begin{figure*}[htbp]
  \centering
  \begin{AIbox}[width=\textwidth]{Prompt for Comparing Multi-Value Checklist Item} 
  \footnotesize
   \begin{Prompt}
You are given two lists of legal information (A and B) about **{checklist_category}**, extracted from two summaries of the same legal case. Your task is to compare these lists based on their **semantic meaning**-that is, what each item conveys, regardless of wording, format, or phrasing.

You should identify:
1. Items that appear in **both A and B** (i.e., semantically equivalent),
2. Items that appear **only in A**,
3. Items that appear **only in B**.

# Information to Compare
## List A:
{information_A}

## List B:
{information_B}

# Output Format
Structure your response as follows:
**Reasoning:**
Provide your detailed analysis of how the two lists relate to each other. Explain any mappings between items, and how you determined whether they were equivalent or different.

**Final Answer:**
Output a valid JSON object with the following structure:

```json
{{
  "common": [
    {{"A_index": X, "B_index": Y}},
    ...
  ],
  "only_in_A": [X, ...],
  "only_in_B": [Y, ...]
}}
```

Where:
- `A_index` is the index of the item in List A,
- `B_index` is the index of the semantically equivalent item in List B,
- `only_in_A` lists the indices of items in A that do **not** appear in B,
- `only_in_B` lists the indices of items in B that do **not** appear in A.

# Notes
- Both List A and B are numbered using 1-based indexing.
- Match items even if they are paraphrased or formatted differently.
- Treat legal synonyms and abbreviations as equivalent when appropriate.
- Return only valid JSON in the **Final Answer** section.
       \end{Prompt}
  \end{AIbox}
  \vspace{-5pt}
  \caption[]{}
  \label{prompt:checklist_comparison_multi_value}
\end{figure*}

\begin{figure*}[htbp]
  \centering
  \begin{AIbox}[width=\textwidth]{Prompt for Extract Residual Facts from Uncovered Text by the Checklist Items} 
  \footnotesize
   \begin{Prompt}
You are assisting a lawyer in identifying key information from a legal case summary. You will be given a set of text spans extracted from the summary that may contain meaningful legal or factual content.

Your task is to extract distinct atomic facts from the given spans. Each atomic fact should be a single discrete, self-contained, and verifiable piece of information that can stand on its own. Ignore any spans that contain filler phrases, incomplete clauses, or do not convey meaningful information. If multiple spans express the same fact, extract it only once.

# Note: Do not make assumptions or add information that is not present in the spans.

# Text Spans
{text_spans}

# Output Format

Your output should be in the following JSON format-no extra keys, no prose outside of the JSON:

```
{{
  "reasoning": "<brief analysis of which spans contain meaningful factual information and what those facts are>",
  "extracted": [
    {{
      "fact": "<atomic fact 1>",
      "evidence_spans": [<list of 1-based span indices>]
    }},
    {{
      "fact": "<atomic fact 2>",
      "evidence_spans": [<list of 1-based span indices>]
    }}
    // ...
  ]
}}
```

## Definitions of each part
* `reasoning`: A brief analysis of the spans and how you identified any meaningful atomic facts.
* `extracted`: A list of objects, each representing one atomic fact. Every object must have:
  - `fact`: A clear, concise sentence or phrase conveying a distinct, self-contained fact.
  - `evidence_spans`: A list of 1-based indices of the spans that support or directly contain the fact.

## Rules for the JSON schema
{it is the same as the checklist extraction prompt.}
          \end{Prompt}
  \end{AIbox}
  \vspace{-5pt}
  \caption[]{}
  \label{prompt:residual_facts_extraction}
\end{figure*}

\begin{figure*}[htbp]
  \centering
  \begin{AIbox}[width=\textwidth]{Prompt for Rating Writing Style Similarity on Five Aspects} 
  \small
   \begin{Prompt}
You are given two summaries of the same legal case (Summary A and Summary B). Your task is to evaluate how similar they are in terms of structure and writing style across five specific dimensions. You should focus on **similarity** rather than quality-we want to know how alike these summaries are, not which one is better.

# Summaries to Compare
## Summary A:
{summary_A}

## Summary B:
{summary_B}

# Evaluation Dimensions with Specific Similarity Scales

{all_5_aspects_definitions}

# Output Format

Structure your response as follows:

**Analysis:**
Provide a detailed comparison for each dimension, explaining specific similarities and differences you observe between Summary A and Summary B.

**Scores:**
Output a valid JSON object with your similarity ratings:

```json
{{
  "readability_jargon": X,
  "narrative_order": X,
  "sentence_structure": X,
  "formatting_layout": X,
  "citation_style": X
}}
```

Where X is your similarity rating (1-5) for each dimension.

# Important Notes
- Focus on similarity, not quality or factual correctness
- Evaluate style and structure only, ignore content differences
- Consider the summaries as a whole when rating each dimension
- Apply the scale objectively for every dimension, strictly following each definition

          \end{Prompt}
  \end{AIbox}
  \vspace{-5pt}
  \caption[]{}
  \label{prompt:writing_style_rating}
\end{figure*}

\begin{figure*}[htbp]
  \centering
  \begin{AIbox}[width=\textwidth]{Prompt for Legal Summarization} 
  \small
   \begin{Prompt}
You are given multiple documents related to a legal case. Your task is to generate a clear, legally precise, and self-contained summary that would let the reader grasp the case without consulting the source files without being excessively long or overly detailed.

Write the summary as a factual narrative. The checklist below shows what to include. Items marked "(if applicable)" should only be included when relevant. If information isn't in the documents, omit it-do not speculate.

# Legal Case Summary Checklist
{all_26_checklist_item_definitions}

# Case Documents
{case_documents}

# Output Format
Please structure your response as follows:
**Reasoning:** Briefly explain what key elements you focused on in the documents to build your summary.

**Case Summary:** A clear, legally precise narrative of the case, written in paragraph form, without being too long.

# Guidelines
* Write as a narrative in paragraph form using clear language. Use a logical order-chronological if helpful, but flexible if another sequence improves clarity.
* Include enough detail for understanding while remaining concise.
* Use accurate legal terminology but avoid jargon-write for a general audience.
* Stay strictly factual; do not add analysis beyond what appears in the record.

Now read the case documents and generate the summary following the checklist, output format, and guidelines above.
          \end{Prompt}
  \end{AIbox}
  \vspace{-5pt}
  \caption[]{}
  \label{prompt:summarization}
\end{figure*}

\begin{figure*}[htbp]
  \centering
  \begin{AIbox}[width=\textwidth]{Prompt for End-to-End Extracting Checklist Item from Case Document (Part 1/2)} 
  \footnotesize
   \begin{Prompt}
You are assisting a lawyer in extracting key information from legal case documents. You will be given multiple documents related to a legal case. Your task is to {item_description}

# Note: 
- Do not make assumptions or add information that is not presented in the documents.
- When extracting evidence, quote the exact text from the documents.
- Each extracted value must be self-contained and easy to understand; include important context when available.

# Case Documents
{case_documents}

# Output Format
Your output should be in the following JSON format-no extra keys, no prose outside of the JSON:

```
{
  "reasoning": "<brief analysis of the case documents and how you identified the relevant information or determined that none was present>",
  "extracted": [
    {
      "evidence": [
        {
          "text": "<verbatim snippet 1>",
          "source_document": "<document name>",
          "location": "<page number or section>"
        },
        {
          "text": "<verbatim snippet 2 (if multiple snippets are relevant)>",
          "source_document": "<document name>",
          "location": "<page number or section>"
        }
        // ...
      ],
      "value": "<extracted information from the evidence>"
    }
    // ...
  ]
}
```
          \end{Prompt}
  \end{AIbox}
  \vspace{-5pt}
  \caption[]{}
  \label{prompt:end-to-end-extraction-part-1}
\end{figure*}

\begin{figure*}[htbp]
  \centering
  \begin{AIbox}[width=\textwidth]{Prompt for End-to-End Extracting Checklist Item from Case Document (Part 2/2)} 
  \footnotesize
   \begin{Prompt}
## Definitions of each part
- `reasoning`: A brief analysis of the case documents and how you identified the relevant information or determined that none was present.
- `extracted`: A list of one or more objects, each representing a distinct piece of information relevant to the checklist item. Always use a list, even if there is only one item.
- `evidence`: A list of evidence objects, each containing:
  - `text`: Exact text snippet copied from the case documents
  - `source_document`: The title/name of the document where this evidence was found
  - `location`: The page number or section identifier where the evidence appears
- `value`: The extracted information based on the evidence.

## Rules for the JSON schema
1. **extracted** and **evidence** are always lists, even if they hold a single object.
2. Copy the **text** in evidence objects exactly as it appears in the case documents-no rewriting or paraphrasing.
3. Always include **source_document** and **location** for each piece of evidence.
4. If the case documents contain no relevant information, output the **extracted** as an empty list:

```
{
  "reasoning": "<brief analysis>",
  "extracted": []
}
```

5. Extract information from all relevant documents-do not stop after finding information in just one document.
6. Each distinct piece of information should be a separate item in the **extracted** list.
7. If you cannot determine the specific page number or section, you may use descriptive locations like "beginning of document", "middle section", or "near the end".
          \end{Prompt}
  \end{AIbox}
  \vspace{-5pt}
  \caption[]{}
  \label{prompt:end-to-end-extraction-part-2}
\end{figure*}

\begin{figure*}[htbp]
  \centering
  \begin{AIbox}[width=\textwidth]{Prompt for Chunk-by-Chunk Extracting Checklist Items from Case Documents} 
  \footnotesize
   \begin{Prompt}
You are assisting a lawyer in extracting key information from legal case documents. You will be given a document chunk from a legal case. Your task is to {item_description}

# Note: 
{same as the end-to-end prompt}

# Current State
This is the accumulated extraction state from previous chunks:
{current_state}

# Document Information
- Document Name: {document_name}
- Chunk: {chunk_id}/{total_chunks}

# Document Chunk
{document_chunk}

# Output Format
Your output should be in the following JSON format-no extra keys, no prose outside of the JSON:

```
{{
  "reasoning": "<brief analysis of this document chunk and how you identified any new relevant information or determined that none was present>",
  "extracted": [
    {{
      "evidence": [
        {{
          "text": "<verbatim snippet 1>",
          "source_document": "<document name>",
          "location": "Chunk {chunk_id}/{total_chunks}"
        }},
        {{
          "text": "<verbatim snippet 2 (if multiple snippets are relevant)>",
          "source_document": "<document name>",
          "location": "Chunk {chunk_id}/{total_chunks}"
        }}
        // ...
      ],
      "value": "<extracted information from the evidence>"
    }}
    // ...
  ]
}}
```

## Definitions of each part
{same as the end-to-end prompt}

## Rules for the JSON schema
{{same as the end-to-end prompt}}
          \end{Prompt}
  \end{AIbox}
  \vspace{-5pt}
  \caption[]{}
  \label{prompt:chunk-by-chunk-extraction}
\end{figure*}

\begin{figure*}[htbp]
  \centering
  \begin{AIbox}[width=\textwidth]{System Prompt used in \fwnameAgent (Part 1/3)} 
  \footnotesize
   \begin{Prompt}
You are a document extraction specialist. Your task is to extract **all checklist items specified in the snapshot** from the provided documents, citing evidence for every value.
  
  You operate by analyzing the snapshot and selecting **exactly ONE action per turn**. You must **respond with valid JSON only** - no prose, no extra keys.
  
  # Snapshot
  Provided every turn:
  - Task description
  - Checklist definitions (what items to extract; any number of items)
  - Document catalog with coverage statistics (and catalog_state/version)
  - Checklist summary (which keys are filled/empty/Not Applicable)
  - Recent action history
  
  # Goal
  Systematically extract all applicable checklist items with proper evidence.
  
  # Decision Policy
  Choose exactly one action each turn:
  - If the document catalog is **unknown** -> call `list_documents`.
  - If a specific document likely contains a target value, choose ONE:
    * `read_document` - default choice. Read a targeted window (<=10,000 tokens) in a document.
    * `search_document_regex` - use this when the target is clearly patternable (e.g., "Case No.", "Filed:", citations).
  - When you have confirmed text for one or more keys:
    - Use `append_checklist` for adds new entries for some checklist items.
    - Use `update_checklist` to replace the entire extracted list for some checklist items when you have the authoritative/complete set, when correcting earlier entries, or when setting an item to Not Applicable (see "Not Applicable Encoding").
  - Periodically use `get_checklist` to assess remaining gaps.
  - Stop when all keys are filled or set to Not Applicable.
  
  # Systematic Extraction Process
  **After each read_document or search_document_regex action:**
  - Carefully analyze the returned text to identify ALL checklist items that can be extracted.
  - Cross-reference the text against your checklist definitions to avoid missing relevant values.
  - Your next action MUST be append_checklist or update_checklist if you found extractable values in the text just read.
  
  **After each append_checklist or update_checklist action:**
  - Verify whether all extractable values from the preceding text were included.
  - If you notice missed values, immediately append them as the next action before continuing.
          \end{Prompt}
  \end{AIbox}
  \vspace{-5pt}
  \caption[]{}
  \label{prompt:agent-system-prompt-part-1}
\end{figure*}

\begin{figure*}[htbp]
  \centering
  \begin{AIbox}[width=\textwidth]{System Prompt used in \fwnameAgent (Part 2/3)} 
  \footnotesize
   \begin{Prompt}
  # Document Reading Efficiency
  - **NEVER** reread fully visited documents (marked with  Fully Visited).
  - **NEVER** reread token ranges already viewed (shown as "Viewed tokens: X-Y").
  - When reading partially visited documents (marked with Partially Visited), read ONLY unviewed token ranges.
  - Check the "Viewed tokens" list before calling read_document to avoid redundant reads.
  
  # Write Semantics
  - **Any checklist item can have multiple values**; the `extracted` field is always a list.
  - **append_checklist**: add new entries; **Do not** set Not Applicable via `append_checklist`.
  - **update_checklist**: replace the entire `extracted` list; use for single-valued items, complete/authoritative sets, corrections, or to set "Not Applicable".
  
  # Evidence Requirements
  - **Every extracted entry must include evidence** with:
    - `text` (verbatim snippet),
    - `source_document` (document name),
    - `location` (e.g., page, section, docket entry; include token offsets if available).
  
  # Not Applicable Encoding
  - Represent Not Applicable as a **single extracted entry** for that key, set **via `update_checklist`**:
    - `value`: **"Not Applicable"** (exact string; case-sensitive)
    - `evidence`: required (explicit text or a dispositive posture supporting Not Applicable)
  - A key is treated as **Not Applicable** only if its `extracted` list contains **exactly one** entry whose `value` is "Not Applicable".
  - Do **not** mark Not Applicable solely because you failed to find a value; require explicit text or logically dispositive evidence (e.g., dismissal with prejudice -> no settlement/decree; "no class certification sought" -> class action items Not Applicable).
  - If later evidence shows the item **does** have real values, use `update_checklist` to replace the Not Applicable entry with the confirmed entries.
  
  # Stop Criteria
  - Stop only when every checklist key is either:
    * Complete: all relevant values present in the corpus for that key have been extracted, each with evidence.
    * Not Applicable: represented as a single extracted entry with value "Not Applicable" and supporting evidence.
  - Before stopping, verify state with `get_checklist` (in a prior turn if needed) and, if consolidation is required, issue one final `update_checklist` (in a prior turn) to replace any incrementally built keys with their curated final lists. Then return the stop decision.
          \end{Prompt}
  \end{AIbox}
  \vspace{-5pt}
  \caption[]{}
  \label{prompt:agent-system-prompt-part-2}
\end{figure*}

\begin{figure*}[htbp]
  \centering
  \begin{AIbox}[width=\textwidth]{System Prompt used in \fwnameAgent (Part 3/3)} 
  \footnotesize
   \begin{Prompt}
  {{TOOL_DESCRIPTIONS}}
  
  # Response Format
  - On each assistant turn, do exactly **one** of:
    1) **Issue one function call**, or
    2) **Stop** if all applicable checklist items are fully extracted and any non-applicable items are marked.
  - When stopping, return **only** this JSON (no extra text):
  ```json
  {
    "decision": "stop",
    "reason": "<brief justification>"
  }
          \end{Prompt}
  \end{AIbox}
  \vspace{-5pt}
  \caption[]{}
  \label{prompt:agent-system-prompt-part-3}
\end{figure*}

\end{document}